\documentclass{article}

\usepackage{microtype}
\usepackage{graphicx}
\usepackage{subfigure}
\usepackage{booktabs} % for professional tables
\usepackage{array} % for professional tables

\usepackage[utf8]{inputenc} % allow utf-8 input
\usepackage[T1]{fontenc}    % use 8-bit T1 fonts
\usepackage{hyperref}       % hyperlinks
\usepackage{url}            % simple URL typesetting
\usepackage{booktabs}       % professional-quality tables
\usepackage{amsfonts}       % blackboard math symbols
\usepackage{nicefrac}       % compact symbols for 1/2, etc.
\usepackage{microtype}      % microtypography
\usepackage{mathtools}

\usepackage{amsmath}
\usepackage{algorithm}
\usepackage[noend]{algpseudocode}

\usepackage{caption}
\usepackage{graphicx}
\usepackage{amssymb}
\usepackage[inline]{enumitem}
\usepackage{wrapfig}
\usepackage{color}
\usepackage{multirow}
\usepackage{cleveref}
\usepackage{verbatim}
\usepackage{setspace}

\def\ppo{PPO}
\def\randfm{Rand-FM}
\def\ibacsni{IBAC-SNI}
\def\ppg{PPG}

\def\ucbdrac{UCB-DrAC}
\def\plr{PLR}

\def\mixreg{Mixreg}

\def\gae{DAAC} % decoupled actor critic advantage DACA or decoupled advantage actor critic  DAAC (this is too much like A2C which is not what we are doing)
\def\ordergae{IDAAC} % alternative names: StepInvDACA or SIDACA or IDACA or InvDACA or ODACA or OIDACA

\def\valuegae{DVAC}
\def\gaeppo{AAC} % APPO

\newcommand{\eg}{\textit{e.g.}}
\newcommand{\ie}{\textit{i.e.}}

\DeclareMathOperator*{\argmax}{arg\,max}
\DeclareMathOperator*{\argmin}{arg\,min}

\algdef{SE}[SUBALG]{Indent}{EndIndent}{}{\algorithmicend\ }%
\algtext*{Indent}
\algtext*{EndIndent}

\usepackage{hyperref}

\usepackage[accepted]{icml2021}

\icmltitlerunning{Decoupling Value and Policy for Generalization in Reinforcement Learning}

\begin{document}

\twocolumn[
\icmltitle{Decoupling Value and Policy for Generalization in Reinforcement Learning}

\icmlsetsymbol{equal}{*}

\begin{icmlauthorlist}
\icmlauthor{Roberta Raileanu}{nyu}
\icmlauthor{Rob Fergus}{nyu}
\end{icmlauthorlist}

\icmlaffiliation{nyu}{Deptartment of Computer Science, New York University, New York, USA}

\icmlcorrespondingauthor{Roberta Raileanu}{raileanu@cs.nyu.edu}

\icmlkeywords{Reinforcemet Learning, Deep Learning, Generalization, Machine Learning, ICML}

\vskip 0.3in
]

\printAffiliationsAndNotice{}  % leave blank if no need to mention equal contribution

\begin{abstract}
% \vspace{-1mm}
Standard deep reinforcement learning algorithms use a shared representation for the policy and value function, especially when training directly from images. However, we argue that more information is needed to accurately estimate the value function than to learn the optimal policy. Consequently, the use of a shared representation for the policy and value function can lead to overfitting. To alleviate this problem, we propose two approaches which are combined to create IDAAC: Invariant Decoupled Advantage Actor-Critic. First, IDAAC decouples the optimization of the policy and value function, using separate networks to model them. Second, it introduces an auxiliary loss which encourages the representation to be invariant to task-irrelevant properties of the environment. IDAAC shows good generalization to unseen environments, achieving a new state-of-the-art on the Procgen benchmark and outperforming popular methods on DeepMind Control tasks with distractors. Our implementation is available at \url{https://github.com/rraileanu/idaac}.

% Moreover, IDAAC learns representations and predictions which are more robust to cosmetic changes in the observations that do not change the underlying state of the environment. 

\end{abstract}

% \vspace{-7mm}
\section{Introduction}
\label{sec:intro}
% \vspace{-1mm}

Generalization remains one of the main challenges of deep reinforcement learning (RL). Current methods fail to generalize to new scenarios even when trained on semantically similar environments with the same high-level goal but different dynamics, layouts, and visual appearances~\citep{Farebrother2018GeneralizationAR, Packer2018AssessingGI, Zhang2018ADO, cobbe2018quantifying, Gamrian2019TransferLF, cobbe2019leveraging, Song2020ObservationalOI}. This indicates that standard RL agents memorize specific trajectories rather than learning transferable skills. Several strategies have been proposed to alleviate this problem, such as the use of regularization~\citep{Farebrother2018GeneralizationAR, Zhang2018ADO, cobbe2018quantifying, igl2019generalization}, data augmentation~\citep{cobbe2018quantifying, lee2020network, ye2020rotation, kostrikov2020image, laskin2020reinforcement, Raileanu2020AutomaticDA}, or representation learning~\citep{zhang2020invariant, zhang2020learning, Mazoure2020DeepRA, Stooke2020DecouplingRL, Agarwal2021ContrastiveBS}. 

Here we consider the problem of generalizing to unseen instances (or levels) of procedurally generated environments, after training on a relatively small number of such instances. While the high-level goal is the same, the background, dynamics, layouts, as well as the locations, shapes, and colors of various entities, differ across instances.

In this work, we identify a new factor that leads to overfitting in such settings, namely the use of a shared representation for the policy and value function. We point out that accurately estimating the value function requires instance-specific features in addition to the information needed to learn the optimal policy. When training a common network for the policy and value function (as is currently standard practice in pixel-based RL), the need for capturing level-specific features in order to estimate the value function can result in a policy that does not generalize well to new task instances. 
% overfits to the training environments.
% We point that the value function of a state can depend on the environment instantiation even when the optimal policy is independent of the instance (\ie{} there exists an optimal set of features for learning an optimal policy which does not depend on the environment instance). 

\begin{figure*}[ht!]
    \centering
    % \vspace{-3mm}
    \includegraphics[width=1.0\textwidth]{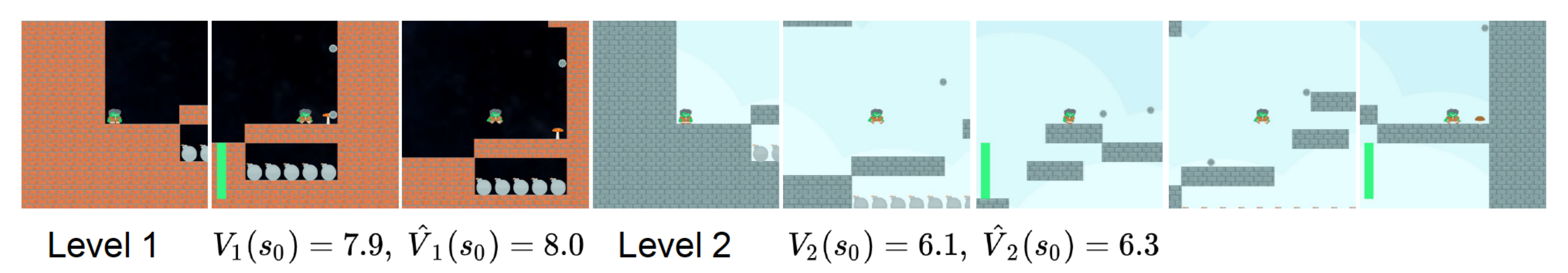}
    % \vspace{-9mm}
    \caption{\textbf{Policy-Value Asymmetry.} Two Ninja levels with initial observations that are \textit{semantically identical but visually different}. Level 1 (first three frames from the left with black background) is much shorter than Level 2 (last five frames with blue background). Both the true values and the estimated values (by a PPO agent trained on 200 levels) of the initial observation are higher for Level 1 than for Level 2 \ie{} $V_1(s_0) > V_2(s_0)$ and $\hat{V}_1(s_0) > \hat{V}_2(s_0)$. Thus to accurately predict the value function, the representations must capture level-specific features (such as the backgrounds), which are irrelevant for learning the optimal policy. Consequently, using a common representation for both the policy and value function can lead to overfitting to spurious correlations and poor generalization to unseen levels.}
    % \vspace{-6mm}
    \label{fig:ninja_levels}
\end{figure*}

% \vspace{-2mm}
\subsection{Policy-Value Representation Asymmetry}
\label{sec:asym}
% \vspace{-1mm}

To illustrate this phenomenon, which we call the \textit{policy-value representation asymmetry}, consider the example in Figure~\ref{fig:ninja_levels} which shows two different levels from the Procgen game Ninja~\citep{cobbe2019leveraging}. The first observations of the two levels are semantically identical and could be represented using the same features (describing the locations of the agent, the bombs, and the platform, while disregarding the background patterns and wall colors). An optimal agent should take the same action in both levels, namely that of moving to the right to further explore the level and move closer to the goal (which is always to its right in this game). However, these two observations have different values. Note that Level 1 is much shorter than Level 2, and the levels look quite different from each other after the initial part. A standard RL agent (\eg{} PPO~\citep{schulman2017proximal}) completes the levels in 24 and 50 steps, respectively. The true value of the initial observation is the expected return (\ie{} sum of discounted rewards) received during the episode. In this game, the agent receives a reward of 10 when it reaches the goal and 0 otherwise. Hence, the true value is higher for Level 1 than for Level 2 since the reward is discounted only for 24 steps rather than 50. In order to accurately estimate the value of an observation (which is part of the objective function for many popular RL methods), the agent must memorize the number of remaining steps in that particular level (which for the initial observation is equivalent to the episode's length). To do this, the agent must use instance-specific features such as the background (which can vary across a level so that each observation within a level has a slightly different background pattern). For the case shown in Figure~\ref{fig:ninja_levels}, the agent can learn to associate a black background with a higher value than a blue background. But this is only a spurious correlation since the background has no causal effect on the state's value, unlike the agent's position relative to items it can interact with. If an agent uses a shared representation for learning the policy and value function, capturing such spurious correlations can result in policies that do not generalize well to new instances, similar to what happens in supervised learning~\citep{Arjovsky2019InvariantRM}.

Furthermore, if the environment is partially observed, an agent should have no way of predicting its expected return in a new level. At the same time, the agent could still select the optimal action if it has learned good state representations (\ie{} that capture the minimal set of features needed to act in the environment in order to solve the task). Thus, in partially observed procedurally generated environments, accurately predicting the value function can require instance-specific features which are not necessary for learning the optimal policy. 

To address the policy-value representation asymmetry, we propose \textbf{I}nvariant \textbf{D}ecoupled \textbf{A}dvantage \textbf{A}ctor-\textbf{C}ritic or \textbf{\ordergae{}} for short, which makes two algorithmic contributions. First, \ordergae{} decouples the policy and value optimization by using two separate networks to learn each of them. The policy network has two heads, one for the policy and one for the generalized advantage function. The value network is needed to compute the advantage, which is used both for the policy-gradient objective and as a target for the advantage predictions. Second, \ordergae{} uses an auxiliary loss which constrains the policy representation to be invariant to the task instance.

To summarize, our work makes the following contributions: \begin{enumerate*}[label=(\roman*)]
\item identifies that using a shared representation for the policy and value function can lead to overfitting in RL;
\item proposes a new approach that uses separate networks for the policy and value function while still learning effective behaviors;
\item introduces an auxiliary loss for encouraging representations to be invariant to the task instance, and
\item demonstrates state-of-the-art generalization on the Procgen benchmark and outperforms popular RL methods on DeepMind Control tasks with distractors. 
\end{enumerate*}

% \vspace{-2mm}
\section{Related Work}
\label{sec:related}
% \vspace{-1mm}

\textbf{Generalization in Deep RL.} A recent body of work has pointed out the problem of overfitting in deep RL~\citep{Rajeswaran2017TowardsGA, Machado2018RevisitingTA, Justesen2018IlluminatingGI, Packer2018AssessingGI, Zhang2018ADO, Zhang2018ASO, Nichol2018GottaLF, cobbe2018quantifying, cobbe2019leveraging, Juliani2019ObstacleTA, Raileanu2020RIDERI, Kuttler2020TheNL, Grigsby2020MeasuringVG}. A promising approach to prevent overfitting is to apply regularization techniques originally developed for supervised learning such as dropout~\citep{Srivastava2014DropoutAS, igl2019generalization}, batch normalization~\citep{Ioffe2015BatchNA, Farebrother2018GeneralizationAR, igl2019generalization}, or data augmentation~\citep{cobbe2018quantifying, ye2020rotation, lee2020network, laskin2020reinforcement, Raileanu2020AutomaticDA, Wang2020ImprovingGI}. Other methods use representation learning techniques to improve generalization in RL~\citep{igl2019generalization, Sonar2020InvariantPO, Stooke2020DecouplingRL}. For example,~\citet{zhang2020invariant, zhang2020learning} and~\citet{Agarwal2021ContrastiveBS} learn state abstractions using various bisimulation metrics, while~\citet{roy2020visual} align the features of two domains using Wasserstein distance. Other approaches for improving generalization in RL consist of reducing the non-stationarity inherent in RL using policy distillation~\citep{Igl2020TheIO}, minimizing surprise~\citep{Chen2020ReinforcementLG}, maximizing the mutual information between the agent's internal representation of successive time steps~\cite{Mazoure2020DeepRA}, or generating an automatic curriculum for replaying different levels based on the agent's learning potential~\cite{Jiang2020PrioritizedLR}. Recently,~\citet{Bengio2020InterferenceAG} study the link between generalization and interference in TD-learning, while~\citet{Bertrn2020InstanceBG} prove that agents trained with off-policy actor-critic methods overfit to their training instances, which is consistent with our empirical results. However, none of these works focus on the asymmetry between the optimal policy and value representation.

\textbf{Decoupling the Policy and Value Function.}
While the current standard practice in deep RL from pixels is to share parameters between the policy and value function in order to learn good representations and reduce computational complexity~\citep{Mnih2016AsynchronousMF, Silver2017MasteringTG, schulman2017proximal}, a few papers have explored the idea of decoupling the two for improving sample efficiency~\citep{BarthMaron2018DistributedDD, Pinto2018AsymmetricAC, Yarats2019ImprovingSE, Andrychowicz2020WhatMI, Cobbe2020PhasicPG}. In contrast with prior work, our paper focuses on generalization to unseen environments and is the first one to point out that using shared features for the policy and value functions can lead to overfitting to spurious correlations. Most similar to our work,~\citet{Cobbe2020PhasicPG} aim to alleviate the interference between policy and value optimization, but there are some key differences between their approach and ours. In particular, our method does not require an auxiliary learning phase for distilling the value function while constraining the policy, it does not use gradients from the value function to update the policy parameters, and it uses two auxiliary losses for training the policy network, one based on the advantage function and one that enforces invariance with respect to the environment instance. Prior work has also explored the idea of predicting the advantage in the context of Q-learning~\citep{Wang2016DuelingNA}, but this setting does not pose the same challenges since it does not learn policies directly.

% \vspace{-2mm}
\section{Background}
\label{sec:background}
% \vspace{-1mm}

We consider a distribution $q(m)$ of Partially Observable Markov Decision Processes (POMDPs) $m \in \mathcal{M}$, with $m$ defined by the tuple $(\mathcal{S}_m, \mathcal{O}_m, \mathcal{A}, \mathcal{T}_m, \Omega_m \mathcal{R}_m, \gamma)$, where $\mathcal{S}_m$ is the state space, $\mathcal{O}_m$ is the observation space, $\mathcal{A}$ is the action space, $\mathcal{T}_m(s'| s, a)$ is the state transition probability distribution, 
$\Omega_m(o' | s, a)$ is the observation probability distribution, $\mathcal{R}_m(s, a)$ is the reward function, and $\gamma$ is the discount factor. During training, we restrict access to a fixed set of POMDPs, $\mathcal{M}_{train} = \{m_1, \ldots, m_n\}$, where $m_i \sim q$, $\forall \, i=\overline{1, n}$. The goal is to find a policy $\pi_{\theta}$ which maximizes the expected  discounted reward over the entire distribution of POMDPs,  $J(\pi_{\theta})= \mathbb{E}_{q, \pi, T_m}\big[\sum_{t=0}^T \gamma^t R_m(s_t, a_t)\big]$. 

In practice, we use the Procgen benchmark which contains 16 procedurally generated games. Each game corresponds to a distribution of POMDPs $q(m)$, and each level of a game corresponds to a POMDP sampled from that game's distribution $m \sim q$. The POMDP $m$ is determined by the seed (\ie{} integer) used to generate the corresponding level. Following the setup from~\citet{cobbe2019leveraging}, agents are trained on a fixed set of $n = 200$ levels (generated using seeds from 1 to 200) and tested on the full distribution of levels (generated using any computer integer seed).

% \vspace{-4mm}
\section{Invariant Decoupled Advantage Actor-Critic}
\label{sec:method}
% \vspace{-1mm}

\begin{figure*}[ht!]
    \centering
    \includegraphics[width=0.49\textwidth]{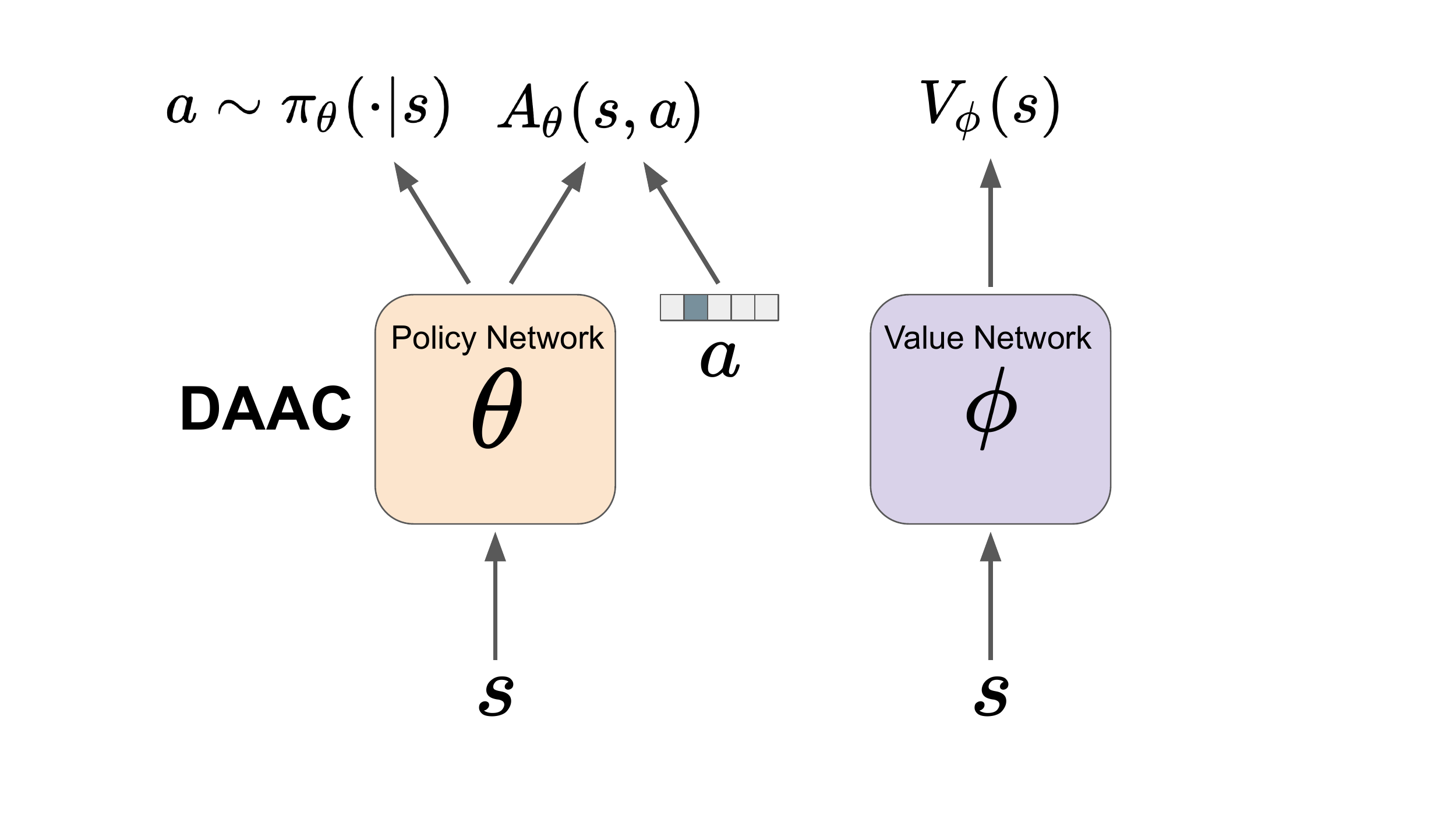}
    \includegraphics[width=0.49\textwidth]{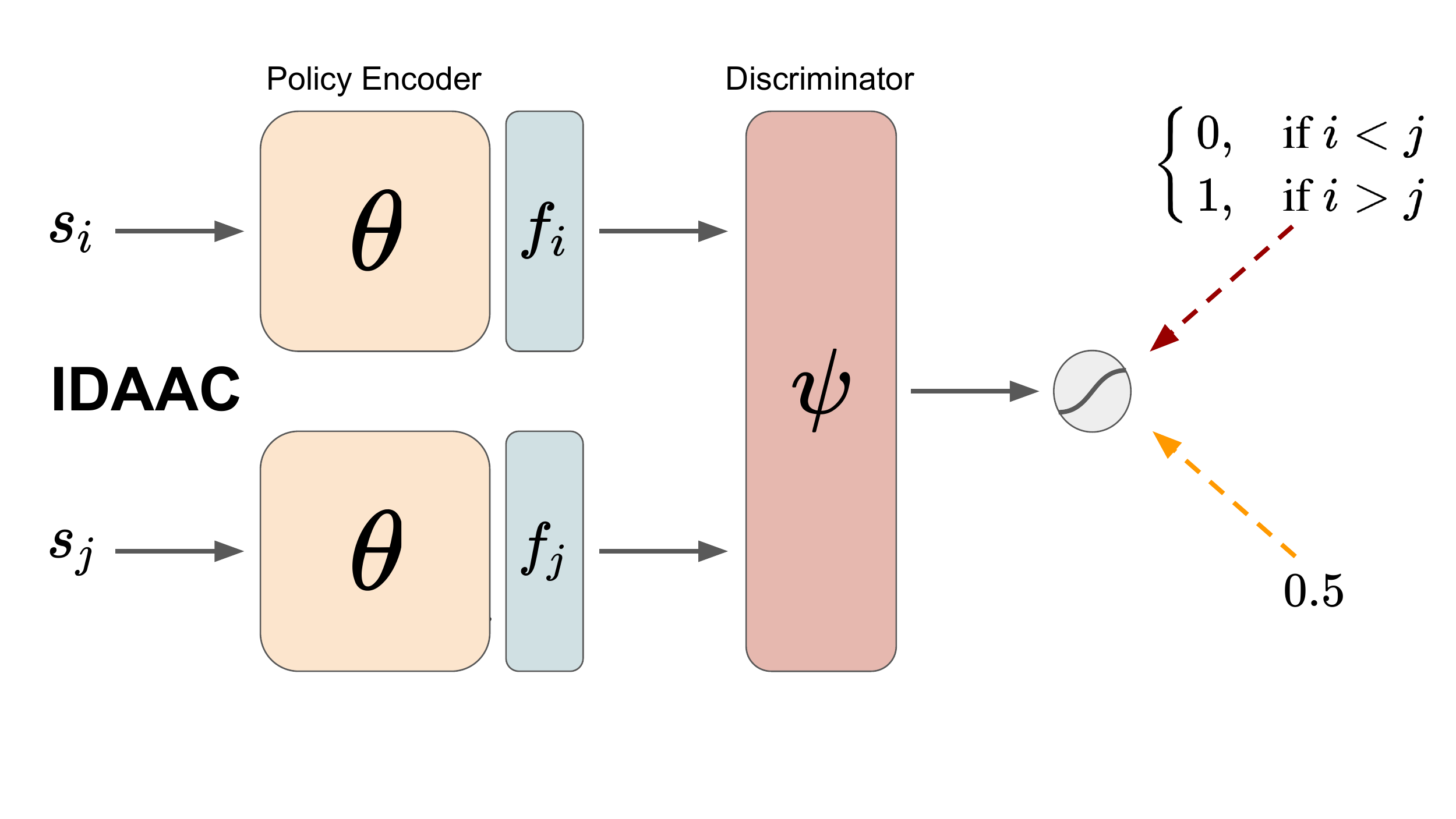}
    % \vspace{-9mm}
    \caption{\textbf{Overview of DAAC (left) and IDAAC (right)}. DAAC uses two separate networks, one for learning the policy and advantage, and one for learning the value. The value estimates are used to compute the advantage targets. IDAAC adds an additional regularizer to the DAAC policy encoder to ensure that it does not contain episode-specific information. The encoder is trained adversarially with a discriminator so that it cannot classify which observation from a given pair $(s_i, s_j)$ was first in a trajectory.}
    % \vspace{-6mm}
    \label{fig:model_diagrams}
\end{figure*}

We start by describing our first contribution, namely the \textbf{D}ecoupled \textbf{A}dvantage \textbf{A}ctor-\textbf{C}ritic (\textbf{DAAC}) algorithm, which uses separate networks for learning the policy and value function (Figure~\ref{fig:model_diagrams} (left), Section~\ref{sec:gae}). Then, we extend this method by adding an auxiliary loss to constrain the policy representation to be invariant to the environment instance, which yields the \textbf{I}nvariant \textbf{D}ecoupled \textbf{A}dvantage \textbf{A}ctor-\textbf{C}ritic (\textbf{IDAAC}) algorithm (Figure~\ref{fig:model_diagrams} (right), Section~\ref{sec:ordergae}). %See Figure~\ref{fig:model_diagrams} for an overview of our methods.

% \subsection{Decoupling the Policy and Value Representations}
% \subsection{Separating the Policy and Value Representations}
% \vspace{-1mm}
\subsection{Decoupling the Policy and Value Function}
\label{sec:decoupling}
% \vspace{-1mm}

To address the problem of overfitting due to the coupling of the policy and value function, we propose a new actor-critic algorithm that uses separate networks for learning the policy and value function, as well as two auxiliary losses. A naive solution to the problem of parameter sharing between the policy and value function would be to simply train two separate networks. However, this approach is insufficient for learning effective behaviors because the policy network relies on gradients from the value function to learn useful features for the training environments. As shown by \citet{Cobbe2020PhasicPG}, using separate networks for optimizing the policy and value function leads to drastically worse training performance than using a shared network, with no sign of progress on many of the tasks (see Figure 8 in their paper). Since such approaches cannot even learn useful behaviors for the training environments, they are no better on the test environments. These results indicate that without gradients from the value to update the policy network, the agent struggles to learn good behaviors. This is consistent with the fact that the gradients from the policy objective are notoriously sparse and high-variance, making training difficult, especially for high-dimensional state spaces (\eg{} when learning from images). In contrast, the value loss can provide denser and less noisy gradients, leading to more efficient training. Given this observation, it is natural to investigate whether other auxiliary losses can provide similarly useful gradients for training the policy network, while also alleviating the problem of overfitting to spurious correlations.

% \vspace{-1mm}
\subsection{Using the Advantage instead of the Value Function}
\label{sec:gae}
% \vspace{-1mm}

As an alternative to the value function, we propose to predict the generalized advantage estimate (GAE) or advantage for short. As illustrated in Section~\ref{sec:adv_val} and Appendix~\ref{app:adv_val}, the advantage is less prone to overfitting to certain types of environment idiosyncrasies. Intuitively, the advantage is a measure of the expected additional return which can be obtained by taking a particular action relative to following the current policy. Because the advantage is a \textit{relative} measure of an action's value while the value is an \textit{absolute} measure of a state's value, the advantage can be expected to vary less with the number of remaining steps in the episode. Thus, the advantage is less likely to overfit to such instance-specific features. To learn generalizable representations, we need to fit a metric invariant to cosmetic changes in the observation which do not modify the underlying state. As shown in Appendix~\ref{app:adv_val}, semantically identical yet visually distinct observations can have very different values but the same advantages (for a given action). This indicates that the advantage might be a good candidate to replace the value as an auxiliary loss for training the policy network.

In order to predict the advantage, we need an estimate of the value function which we obtain by simply training a separate network to output the expected return for a given state. Thus, our method consists of two separate networks, the value network parameterized by $\phi$ which is trained to predict the value function, and the policy network parameterized by $\theta$ which is trained to learn a policy that maximizes the expected return and also to predict the advantage function.
% \rob{maybe add notation here for the different networks}. 

The policy network of \gae{} is trained to maximize the following objective: 
\begin{equation}
\label{eq:daac_loss}
    J_{\mathrm{\gae{}}}(\theta) =  J_{\pi}(\theta) + \alpha_\mathrm{s}\mathrm{S}_{\pi}(\theta) 
 - \alpha_\mathrm{a} L_{\mathrm{A}}(\theta),
\end{equation}
where $J_{\pi}(\theta)$ is the policy gradient objective, $S_{\pi}(\theta)$ is an entropy bonus to encourage exploration, $L_{\mathrm{A}}(\theta)$ is the advantage loss, while $\alpha_\mathrm{s}$ and $\alpha_\mathrm{a}$ are their corresponding weights determining each term's contribution to the total objective.

The policy objective term is the same as the one used by PPO (see Appendix~\ref{app:ppo}):
\begin{equation*}
\label{eq:pg_loss}
    J_{\pi}(\theta) = \hat{\mathbb{E}}_{t} \left[\min\left(r_t(\theta)\hat{A}_t, \; \mathrm{clip}\left(r_t(\theta), 1 - \epsilon, 1 + \epsilon\right)\hat{A}_t\right)\right],
\end{equation*}
where $r_t(\theta) = \frac{\pi_{\theta}(a_t|s_t)}{\pi_{\theta_{old}}(a_t|s_t)}$  and $\hat{A}_t$ is the advantage function at time step $t$.

The advantage function loss term is defined as:
\begin{equation*}
\label{eq:adv_loss}
    L_{\mathrm{A}}(\theta) = \hat{\mathbb{E}}_{t} \left[\left(A_{\theta}(s_t, a_t) - \hat{A_t}\right)^2\right],
\end{equation*}
where $\hat{A}_t$ is the corresponding generalized advantage estimate at time step $t$, $\hat{A}_t = \sum_{k = t}^{T} (\gamma\lambda)^{k-t} \delta_{k}$, 
with $\delta_t = r_t + \gamma V_{\phi}(s_{t+1}) - V_{\phi}(s_t)$ which is computed using the estimates from the value network. 

The value network of \gae{} is trained to minimize the following loss: 
\begin{equation*}
\label{eq:value_loss}
    L_{\mathrm{V}}(\phi) = \hat{\mathbb{E}}_{t} \left[\left(V_{\phi}(s_t) - \hat{V_t}\right)^2\right],
\end{equation*}
where $\hat{V}_t$ is the total discounted reward obtained during the corresponding episode after step $t$,  $\hat{V}_t = \sum_{k = t}^{T}  \gamma^{k-t} r_{k}$.

During training, we alternate between $E_{\pi}$ epochs for training the policy network and $E_V$ epochs for training the value network every $N_\pi$ policy updates. See Algorithm~\ref{alg:daac} from Appendix~\ref{app:algos} for a more detailed description of \gae{}.

As our experiments show, predicting the advantage rather than the value provides useful gradients for the policy network so it can learn effective behaviors on the training environments, thus overcoming the challenges encountered by prior attempts at learning separate policy and value networks~\citep{Cobbe2020PhasicPG}. In addition, it mitigates the problem of overfitting caused by the use of value gradients to update the policy network, thus also achieving better performance on test environments.  % this para could be removed if necessary

% \vspace{-1mm}
\subsection{Learning Instance-Invariant Features}
\label{sec:ordergae}
% \vspace{-1mm}

From a generalization perspective, a good state representation is characterized by one that captures the minimum set of features necessary to learn the optimal policy and ignores instance-specific features which might lead to overfitting. As emphasized in Figure~\ref{fig:ninja_levels}, due to the diversity of procedurally generated environments, the observations may contain information indicative of the number of remaining steps in the corresponding level. Since different levels have different lengths, capturing such information given only a partial view of the environment translates into capturing information specific to that level. Because such features overfit to the idiosyncrasies of the training environments, they can results in suboptimal policies on unseen instances of the same task.

Hence, one way of constraining the learned representations to be agnostic to the environment instance is to discourage them from carrying information about the number of remaining steps in the level. This can be formalized using an adversarial framework so that a discriminator cannot tell which observation from a given pair came first within an episode, based solely on their learned features. Similar ideas have been proposed for learning disentangled representations of videos~\citep{Denton2017UnsupervisedLO}.

Let $E_\theta$ be an encoder that takes as input an observation $s$ and outputs a feature vector $f$. This encoder is the same as the one used by the policy network so it is also parameterized by $\theta$. Let $D$ be a discriminator parameterized by $\psi$ that takes as input two features $f_i$ and $f_j$ (in this order), corresponding to two observations from the same trajectory $s_i$ and $s_j$, and outputs a number between 0 and 1 which represents the probability that observation $s_i$ came before observation $s_j$. The discriminator is trained using a cross-entropy loss that aims to predict which observation was first in the trajectory:
\begin{equation}
\label{eq:discriminator_loss}
\begin{split}
    L_{\mathrm{D}}(\psi) = & - \mathrm{log} \left[ \mathrm{D}_\psi\left(\mathrm{E}_\theta(s_i), \mathrm{E}_\theta(s_j)\right) \right]  \\
                        & - \mathrm{log}\left[1 - \mathrm{D}_\psi\left(\mathrm{E}_\theta(s_i), \mathrm{E}_\theta(s_j)\right)\right].    
\end{split}
\end{equation}
Note that only the discriminator's parameters are updated by minimizing the loss in eq.~\ref{eq:discriminator_loss}, while the encoder's parameters remain fixed during this optimization. 

The other half of the adversarial framework imposes a loss function on the encoder that tries to maximize the uncertainty (\ie{} entropy) of the discriminator regarding which observation was first in the episode:
\begin{equation}
\label{eq:encoder_loss}
\begin{split}
    L_{\mathrm{E}}(\theta) = & - \frac{1}{2} \mathrm{log} \left[ \mathrm{D}_\psi\left(\mathrm{E}_\theta(s_i), \mathrm{E}_\theta(s_j)\right) \right]  \\
                        & - \frac{1}{2} \mathrm{log}\left[1 - \mathrm{D}_\psi\left(\mathrm{E}_\theta(s_i), \mathrm{E}_\theta(s_j)\right)\right].    
\end{split}
\end{equation}
 
Similar to the above, only the encoder's parameters are updated by minimizing the loss in eq.~\ref{eq:encoder_loss}, while the discriminator's parameters remain fixed during this optimization.

Thus, the policy network is encouraged to learn state representations so that the discriminator cannot identify whether a state came before or after another state. In so doing, the learned representations cannot carry information about the number of remaining steps in the environment, yielding features which are less instance-dependent and thus more likely to generalize outside the training distribution. Note that this adversarial loss is only used for training the policy network and not the value network. 

To train the policy network, we maximize the following objective which combines the \gae{} objective from eq.~\ref{eq:daac_loss} with the above adversarial loss, resulting in \ordergae{}'s objective:
\begin{equation}
\label{eq:idaac_loss}
\begin{split}
    J_{\mathrm{\ordergae{}}}(\theta) = J_{\mathrm{\gae{}}}(\theta) - \alpha_i L_{\mathrm{E}}(\theta),
\end{split}
\end{equation}
where $\alpha_i$ is the weight of the adversarial loss relative to the policy objective. Similar to \gae{}, a separate value network is trained. See Algorithm~\ref{alg:idaac} from Appendix~\ref{app:algos} for a more detailed description of \ordergae{}.

% \vspace{-2mm}
\section{Experiments}
\label{sec:experiments}
% \vspace{-1mm}

\begin{table*}[ht!]
    \centering
    % \vspace{-5mm}
    \caption{PPO-Normalized Procgen scores on train and test levels after training on 25M environment steps. Our approaches, \gae{} and \ordergae{}, establish a new state-of-the-art on the test distribution of environments from the Procgen benchmark, while also showing strong training performance. The mean and standard deviation are computed using 10 runs with different seeds.}
    \scriptsize
    \begin{tabular}{c|c|c|c|c|c|c|c|c}
    \toprule
    Score & RAND-FM  & IBAC-SNI  & Mixreg & PLR & UCB-DrAC & PPG & DAAC (Ours) & IDAAC (Ours) \\
    \toprule
    Train & $87.6\pm8.9$ & $103.4\pm8.5$ & $104.2\pm3.1$ & $106.7\pm5.6$ & $118.9\pm8.0$ & $\mathbf{144.5\pm5.7}$  &  $131.0\pm6.1$ & $132.2\pm5.9$ \\
    \midrule
    Test & $78.0\pm9.0$ & $102.9\pm8.6$ & $114.6\pm3.3$ & $128.3\pm5.8$ & $139.7\pm8.3$ & $152.2\pm5.8$ & $162.3\pm6.2$ & $\mathbf{163.7\pm6.1}$  \\
    \bottomrule
    % \vspace{-7mm}
    \end{tabular}
    \label{tab:procgen_agg}
\end{table*}

\begin{figure*}[ht!]
    \centering
    \includegraphics[width=0.24\textwidth]{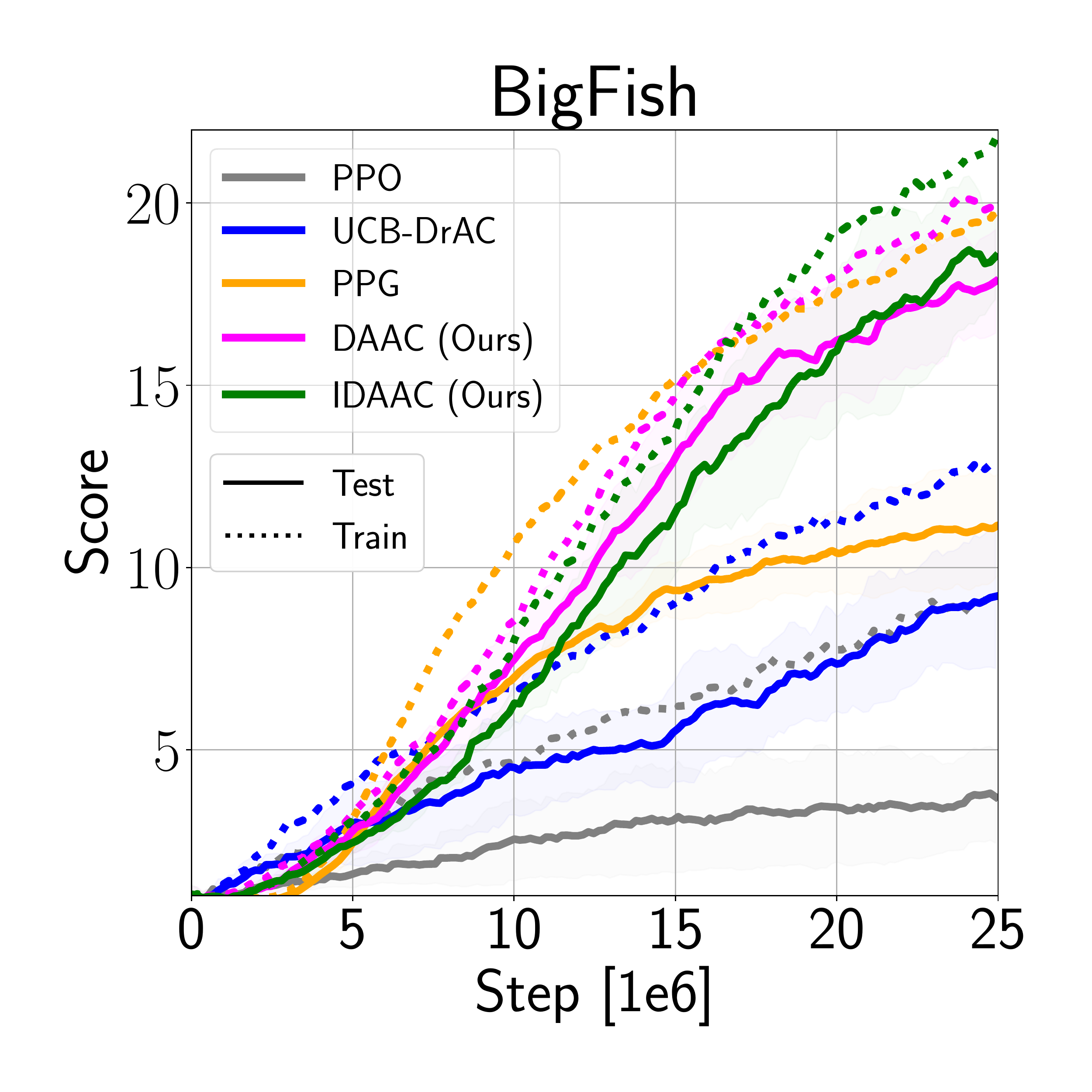}
    \includegraphics[width=0.24\textwidth]{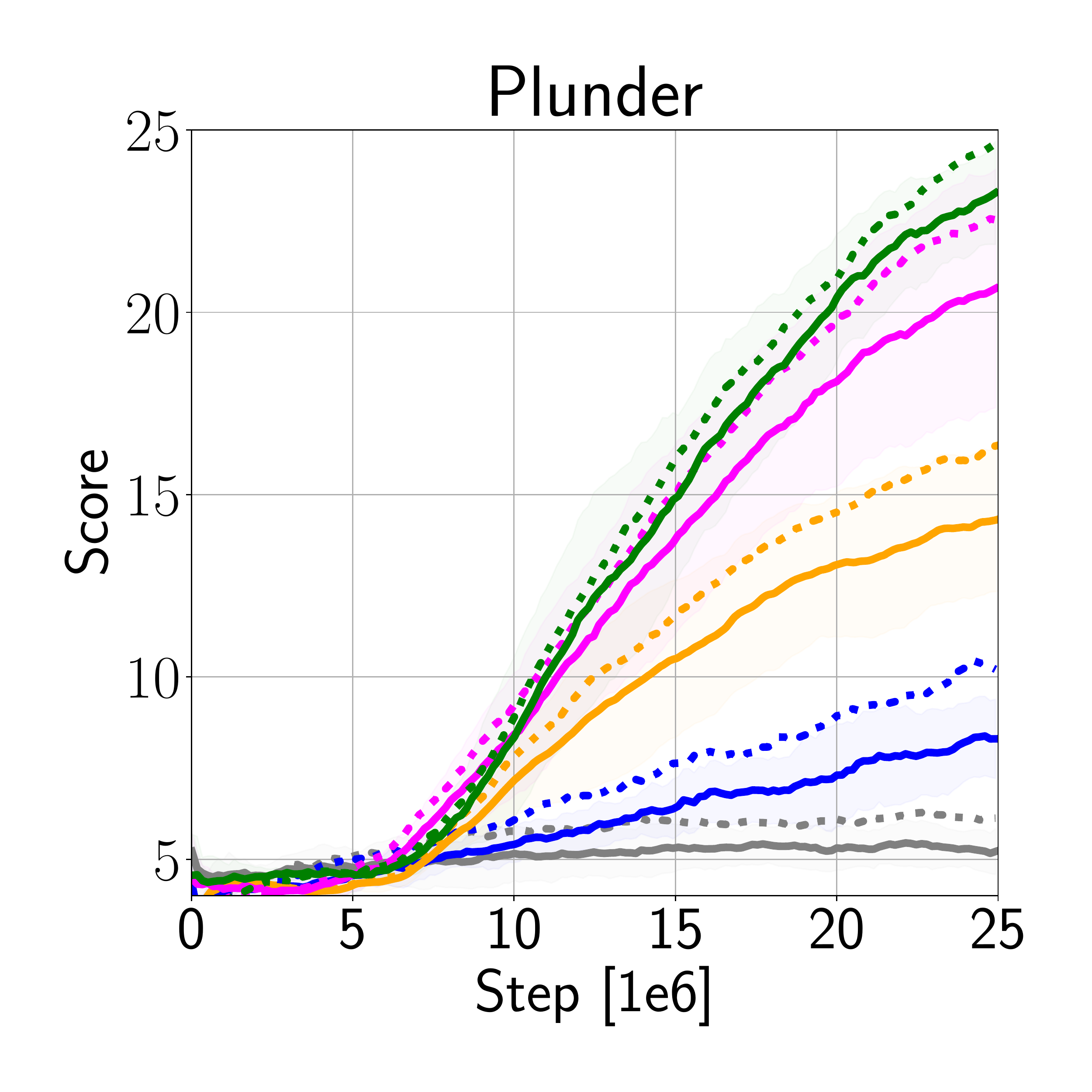}
    \includegraphics[width=0.24\textwidth]{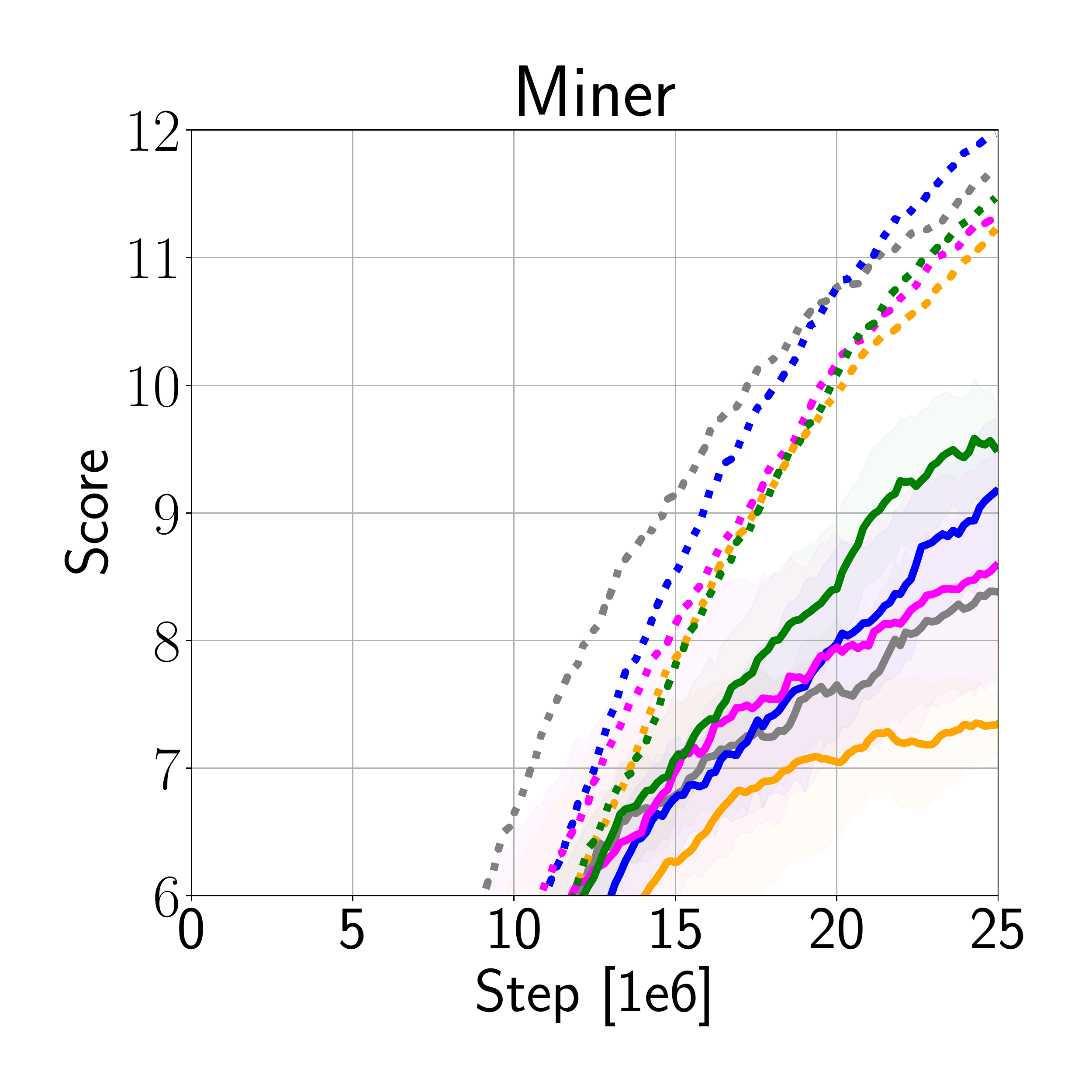}
    \includegraphics[width=0.24\textwidth]{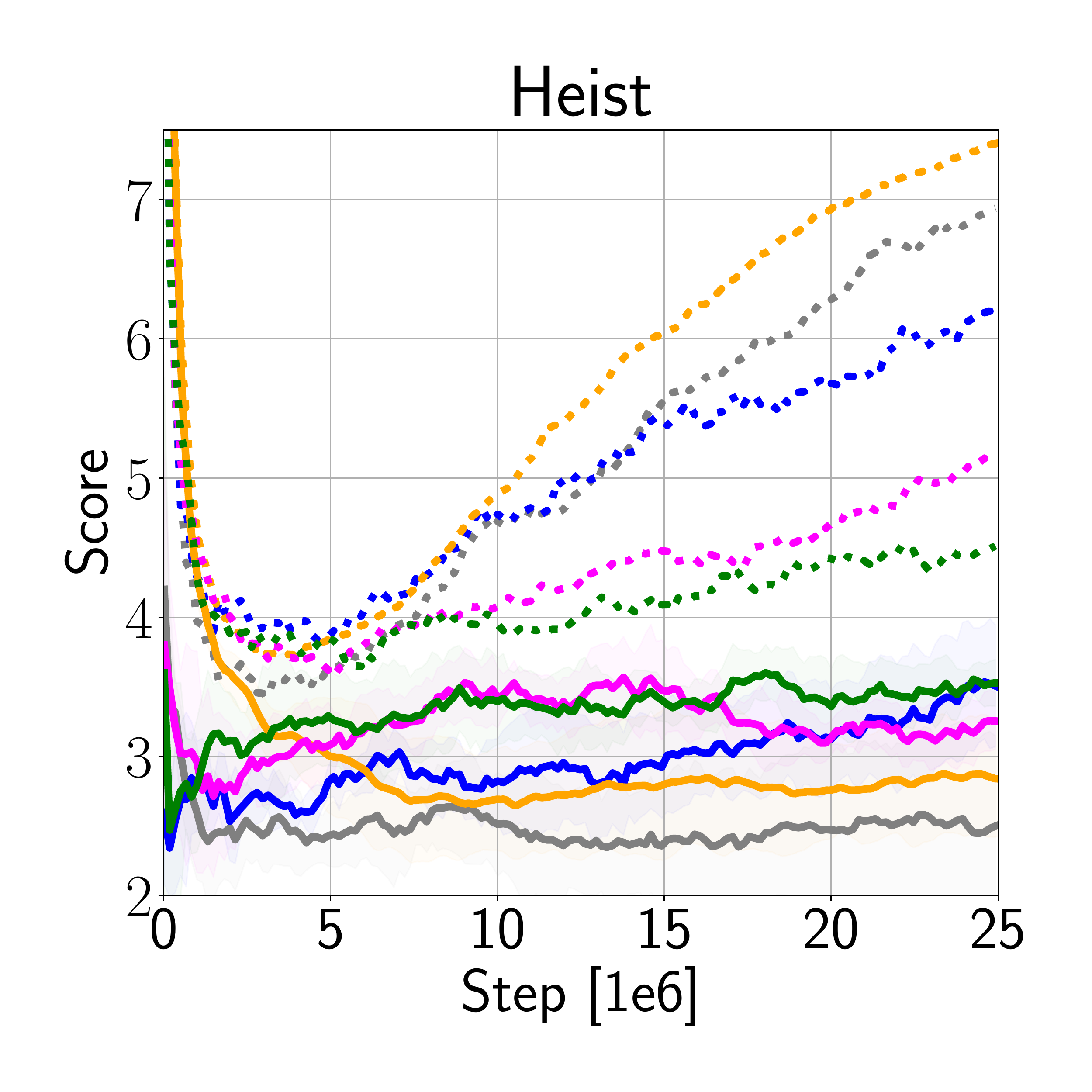}
    \includegraphics[width=0.24\textwidth]{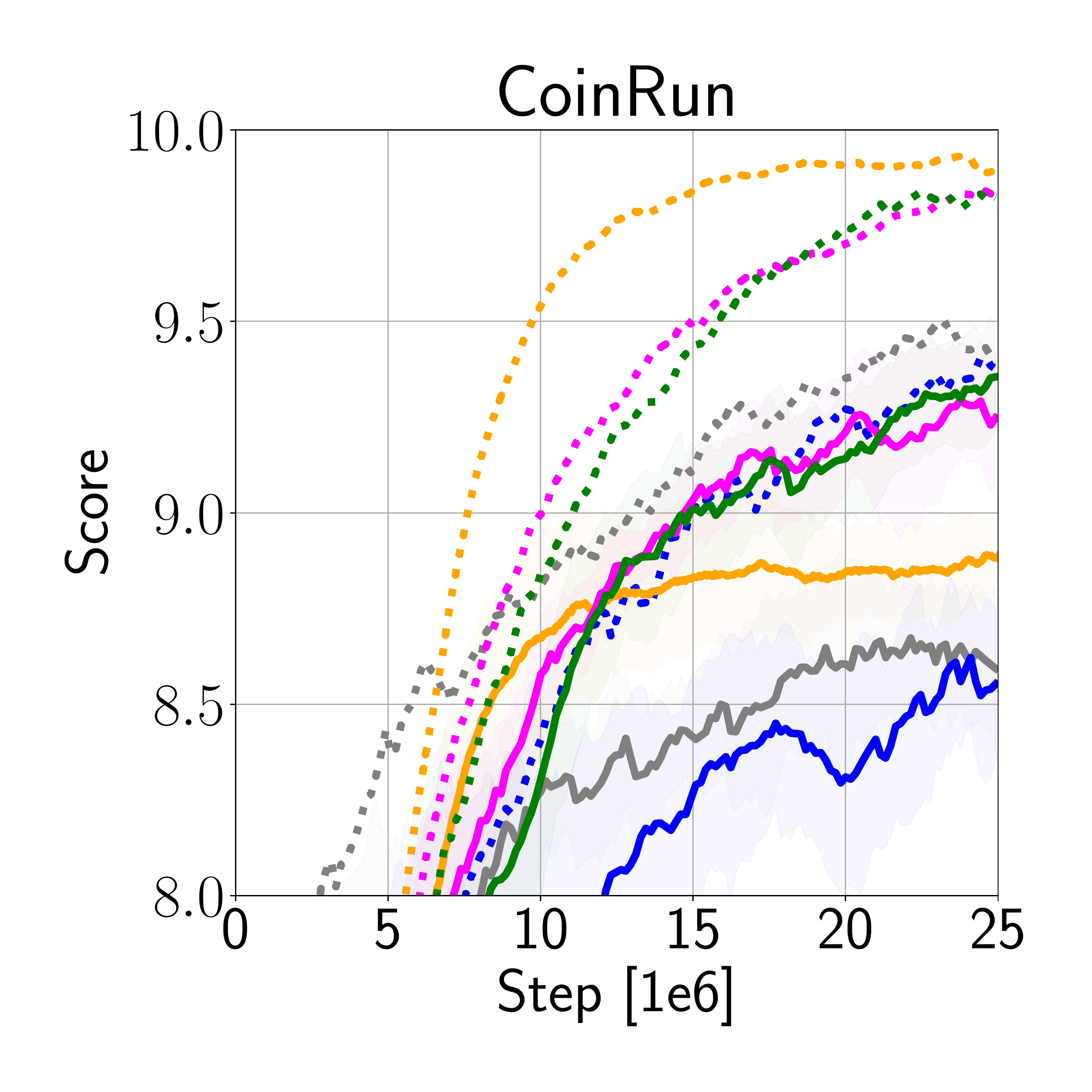}
    \includegraphics[width=0.24\textwidth]{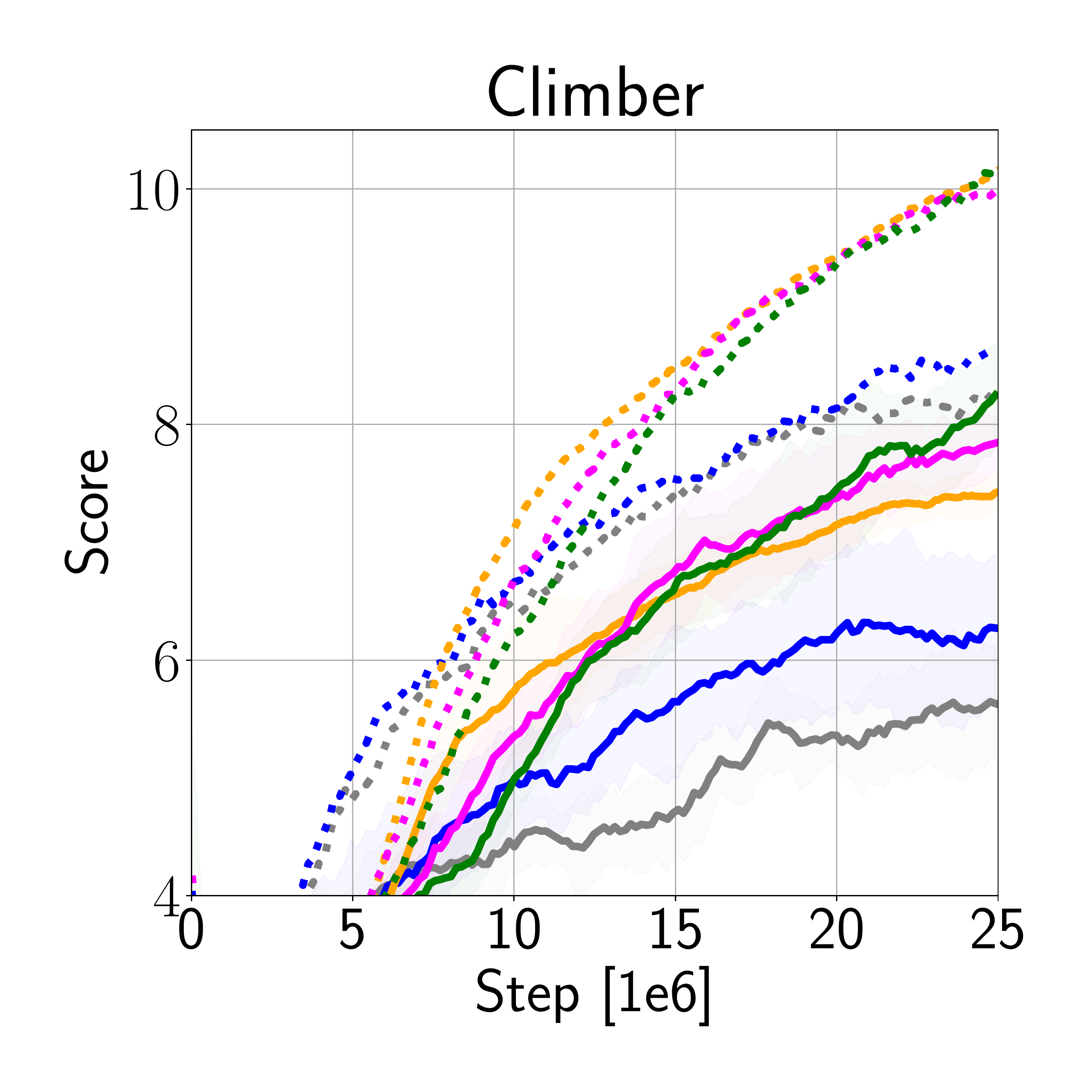}
    \includegraphics[width=0.24\textwidth]{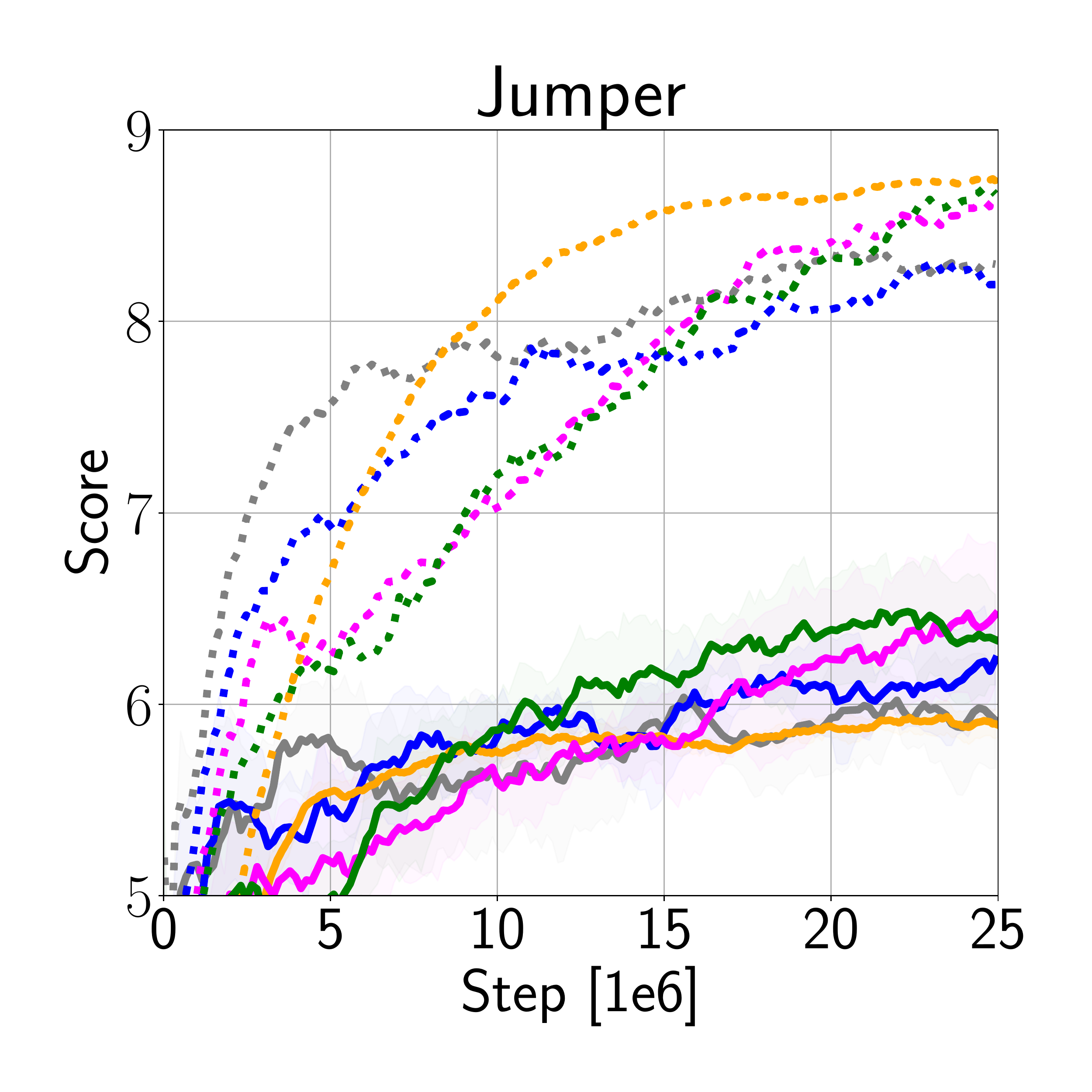}
    \includegraphics[width=0.24\textwidth]{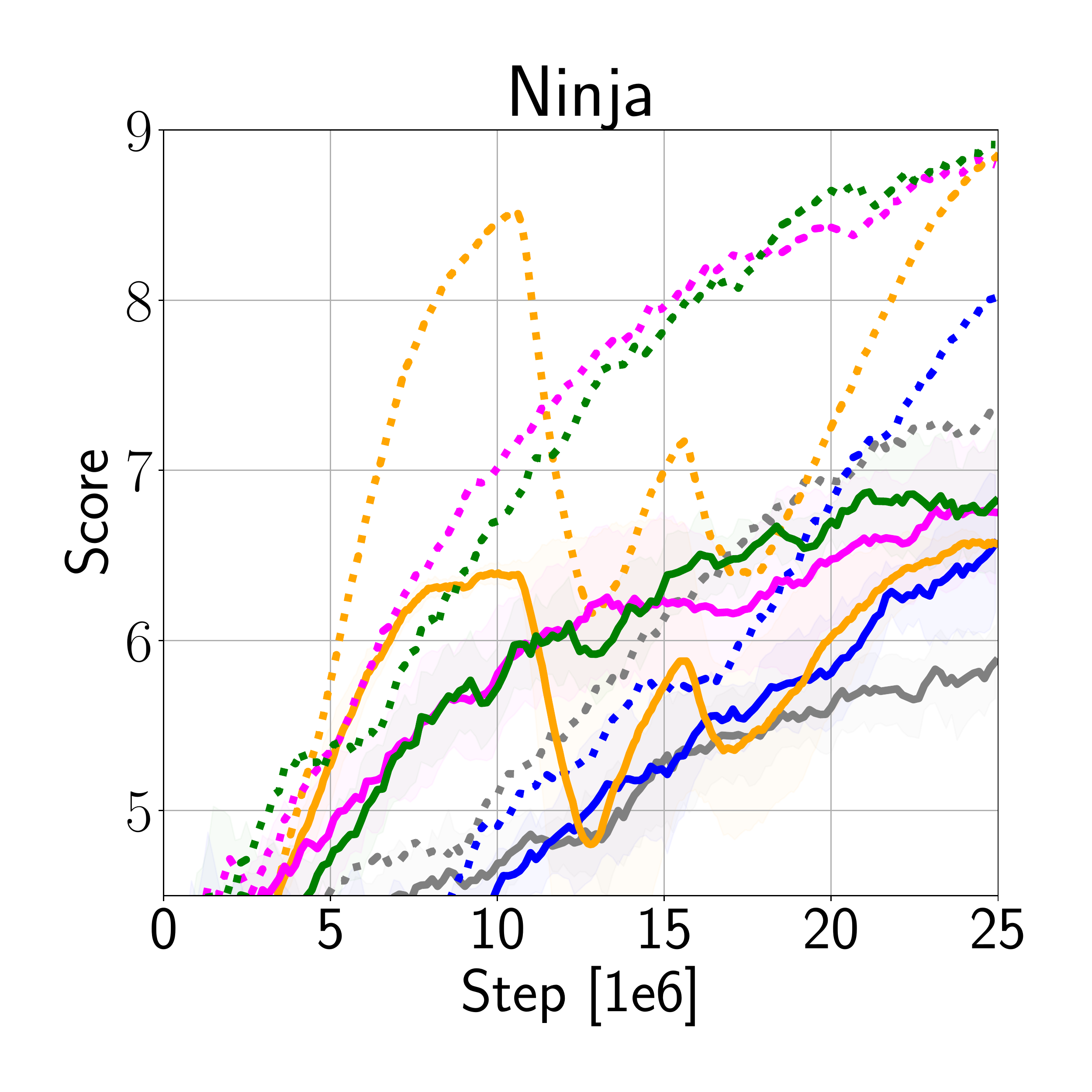}
    % \includegraphics[width=0.24\textwidth]{fig/gae-maze-blog.pdf}
    % \vspace{-6mm}
    \caption{\textbf{Train and Test Performance for \ordergae{}, \gae{}, \ppg{}, \ucbdrac{}, and \ppo{}, on eight diverse Procgen games.} \ordergae{} and \gae{} display state-of-the-art performance on the test levels, beating leading approaches (\ppg{} and \ucbdrac{}), as well a \ppo{} baseline. Furthermore, \ordergae{} and \gae{} exhibit a smaller generalization gap than other methods. The mean and standard deviation are computed over 10 runs with different seeds.}
    % \vspace{-5mm}
    \label{fig:procgen_results}
\end{figure*}

In this section, we evaluate our methods on two distinct environments: (i) three DeepMind Control suite tasks with synthetic and natural background distrators~\citep{Zhang2020LearningIR} and (ii) the full Procgen benchmark~\citep{cobbe2019leveraging} which consists of 16 procedurally generated games. 
Procgen in particular has a number of attributes that make it a good testbed for generalization in RL: \begin{enumerate*}[label=(\roman*)]
\item it has a diverse set of games in a similar spirit with the ALE benchmark \citep{bellemare2013arcade};
\item each of these games has procedurally generated levels which present agents with meaningful generalization challenges;
\item agents have to learn motor control directly from images, and
\item it has a clear protocol for testing generalization, the focus of our investigation.
\end{enumerate*} % this para could also be removed if needed

All Procgen environments use a discrete 15 dimensional action space and produce $64 \times 64 \times 3$ RGB observations. We use Procgen's \textit{easy} setup, so for each game, agents are trained on 200 levels and tested on the full distribution of levels. More details about our experimental setup and hyperparameters can be found in Appendix~\ref{app:hyperparams}. 

% \textbf{Evaluation Metrics.} At the end of training, for each method and each game, we compute the average score over 100 episodes and 10 different seeds. The scores are then normalized using the corresponding PPO score on the same game. We aggregate the normalized scores over all 16 Procgen games and report the resulting mean, median, and standard deviation (Table~\ref{tab:agg_results}). For a per-game breakdown, see Tables~\ref{tab:train_all} and~\ref{tab:test_all} in Appendix~\ref{appendix:procgen_scores}.

%%%%%%%%%%%%%%%%%%%%%%%%%%%%%%%%%%%%%%%%%%%%%%%%%%%%%%%%%%%%%%%%%%%%%%%%
% \vspace{-1mm}
\subsection{Generalization Performance on Procgen}
\label{sec:procgen_results}
% \vspace{-1mm}

We compare  \gae{} and \ordergae{} with seven other RL algorithms: \textbf{\ppo{}}~\citep{schulman2017proximal}, \textbf{\ucbdrac{}}~\citep{Raileanu2020AutomaticDA},  \textbf{\plr{}}~\citep{Jiang2020PrioritizedLR}, \textbf{\mixreg{}}~\citep{Wang2020ImprovingGI}, \textbf{\ibacsni{}}~\citep{igl2019generalization}, \textbf{\randfm{}}~\citep{lee2020network}, and \textbf{\ppg{}}~\citep{Cobbe2020PhasicPG}. 

\ucbdrac{} is the previous state-of-the-art on Procgen and uses data augmentation to learn policy and value functions invariant to various input transformations \plr{} is a newer approach that uses an automatic curriculum based on the learning potential of each level and achieves strong results on Procgen. \randfm{} uses a random convolutional network to regularize the learned representations, \ibacsni{} uses an information bottleneck with selective noise injection, while \mixreg{} uses mixtures of observations to impose linearity constraints between the agent's inputs and outputs. All three were designed to improve generalization in RL and evaluated on Procgen games. Finally, \ppg{} is the only method we are aware of that learns good policies while decoupling the optimization of the policy and value function. However, \ppg{} was designed to improve sample efficiency rather than generalization and the method was evaluated only on Procgen's training distribution of environments. See Section~\ref{sec:related} for a detailed discussion of the differences between \ppg{} and our methods.

\begin{figure*}[ht!]
    \centering
    % \vspace{-3mm}
    \includegraphics[width=0.24\textwidth]{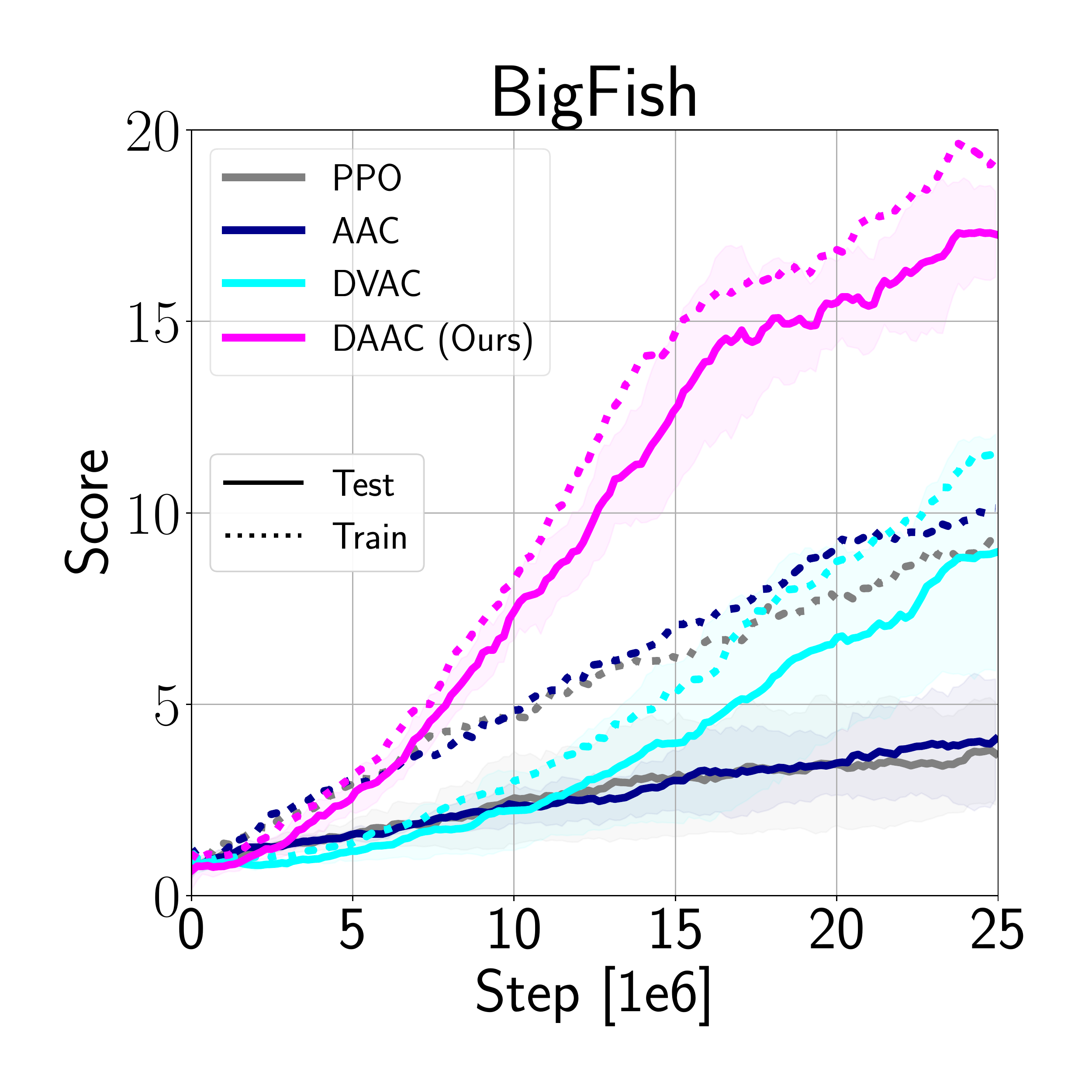}
    \includegraphics[width=0.24\textwidth]{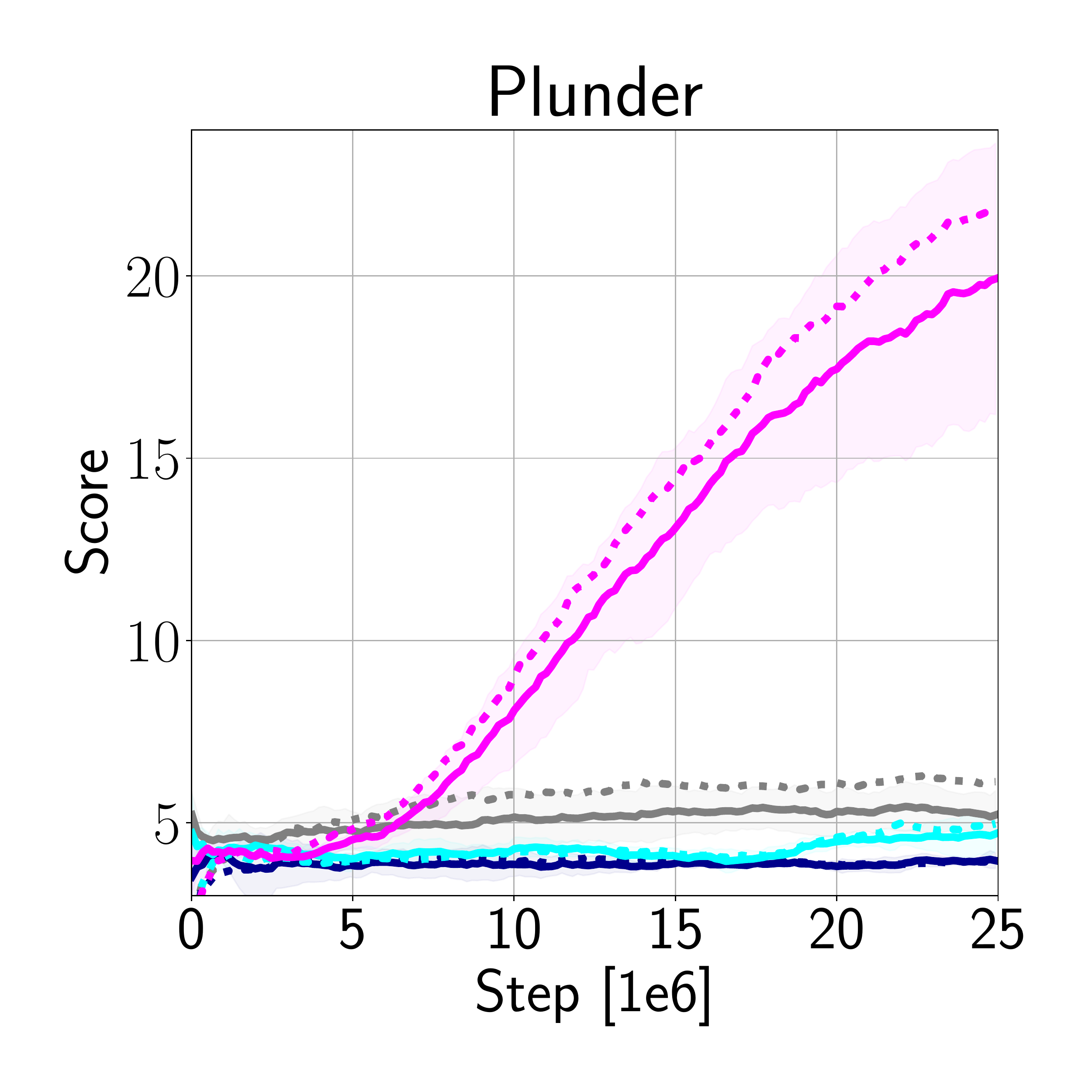}
    \includegraphics[width=0.24\textwidth]{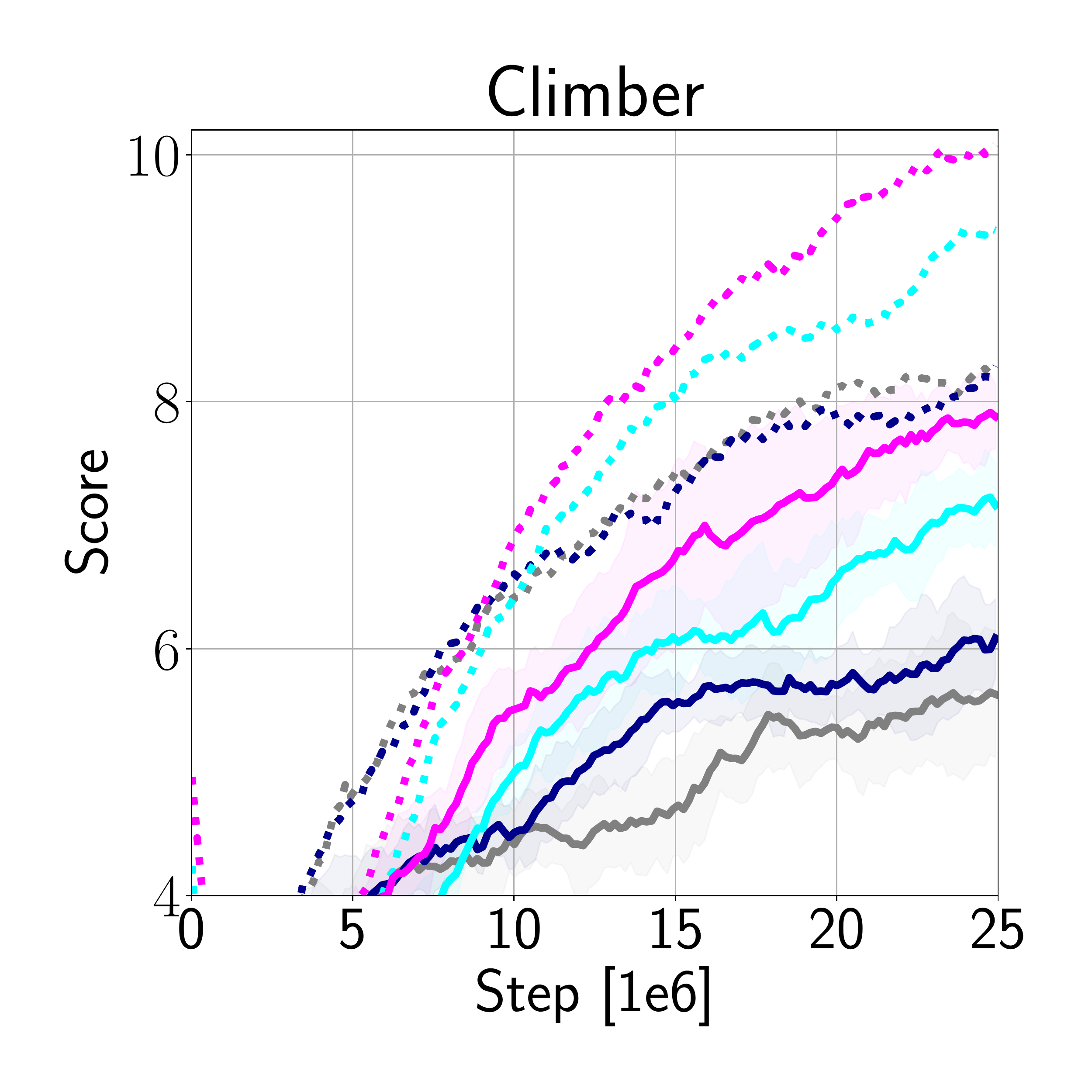}
    \includegraphics[width=0.24\textwidth]{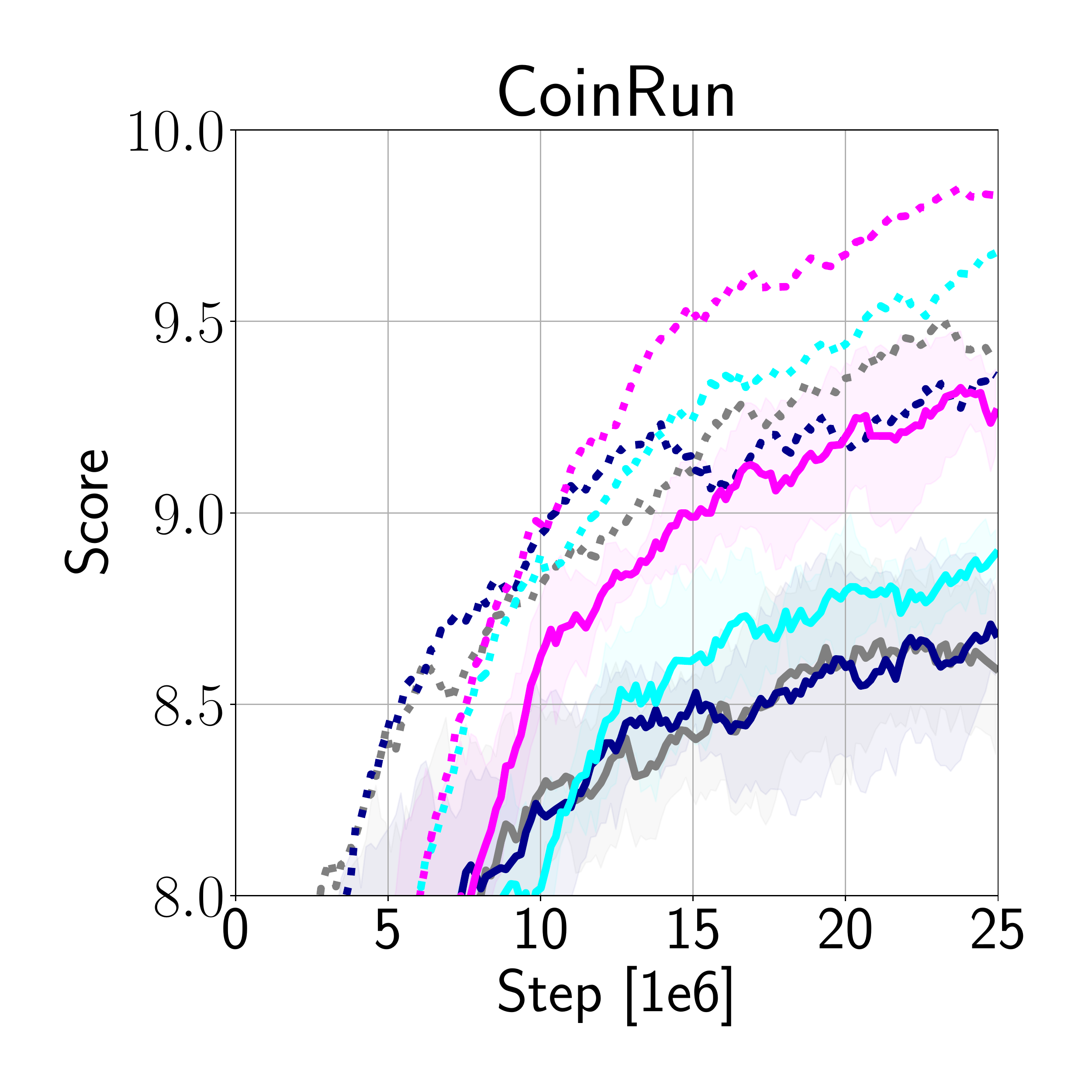}
    % \vspace{-6mm}
    \caption{\textbf{Train and Test Performance for \ppo{}, \gae{}, and two of its ablations, \valuegae{} and \gaeppo{}, on four Procgen games.} \gae{} outperforms all the ablations on both train and test, emphasizing the importance of each component. The mean and standard deviation are computed over 5 runs with different seeds.}
    % \vspace{-3mm}
    \label{fig:procgen_ablations}
\end{figure*}

\begin{figure*}[ht!]
    \centering
    \includegraphics[width=0.24\textwidth]{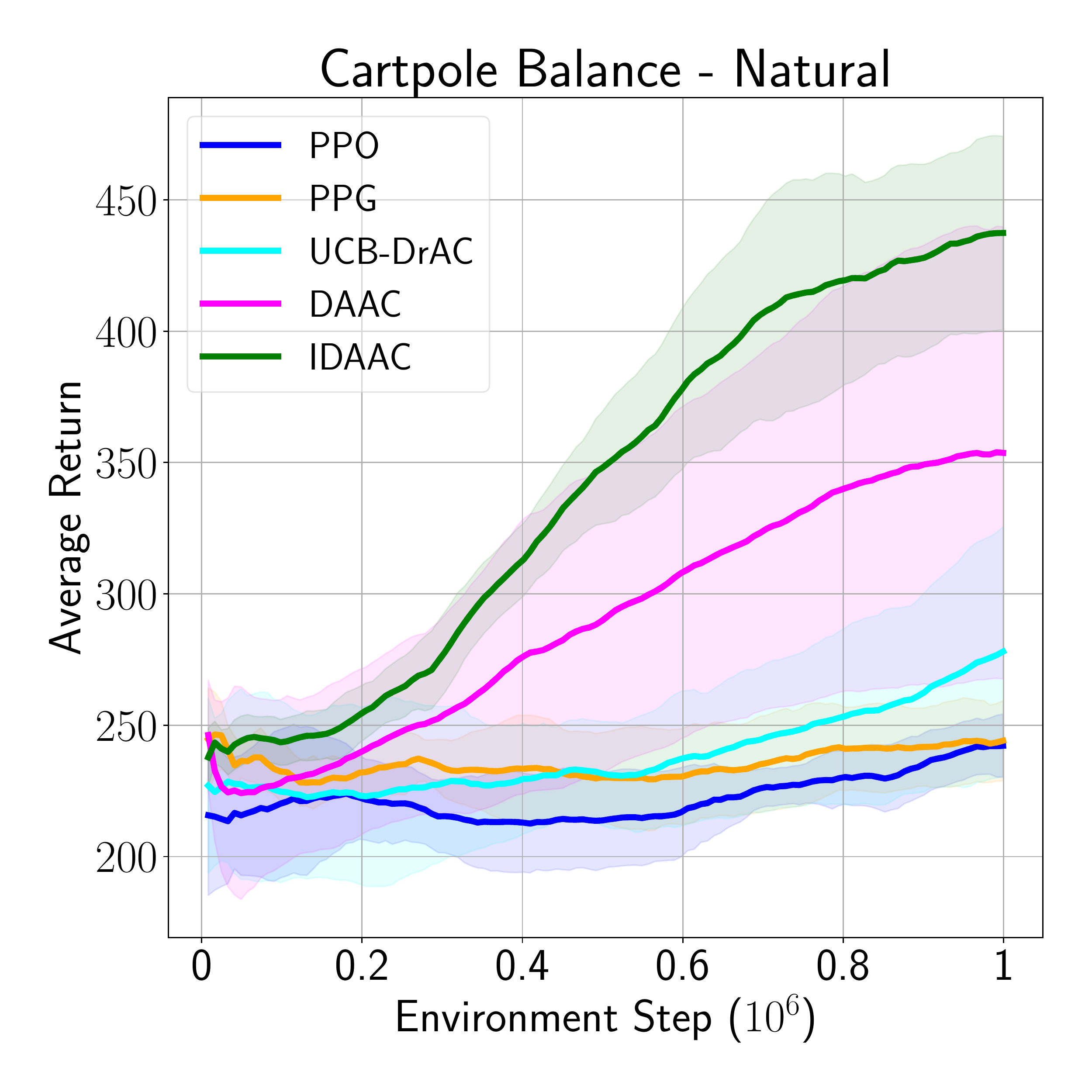}
    \includegraphics[width=0.24\textwidth]{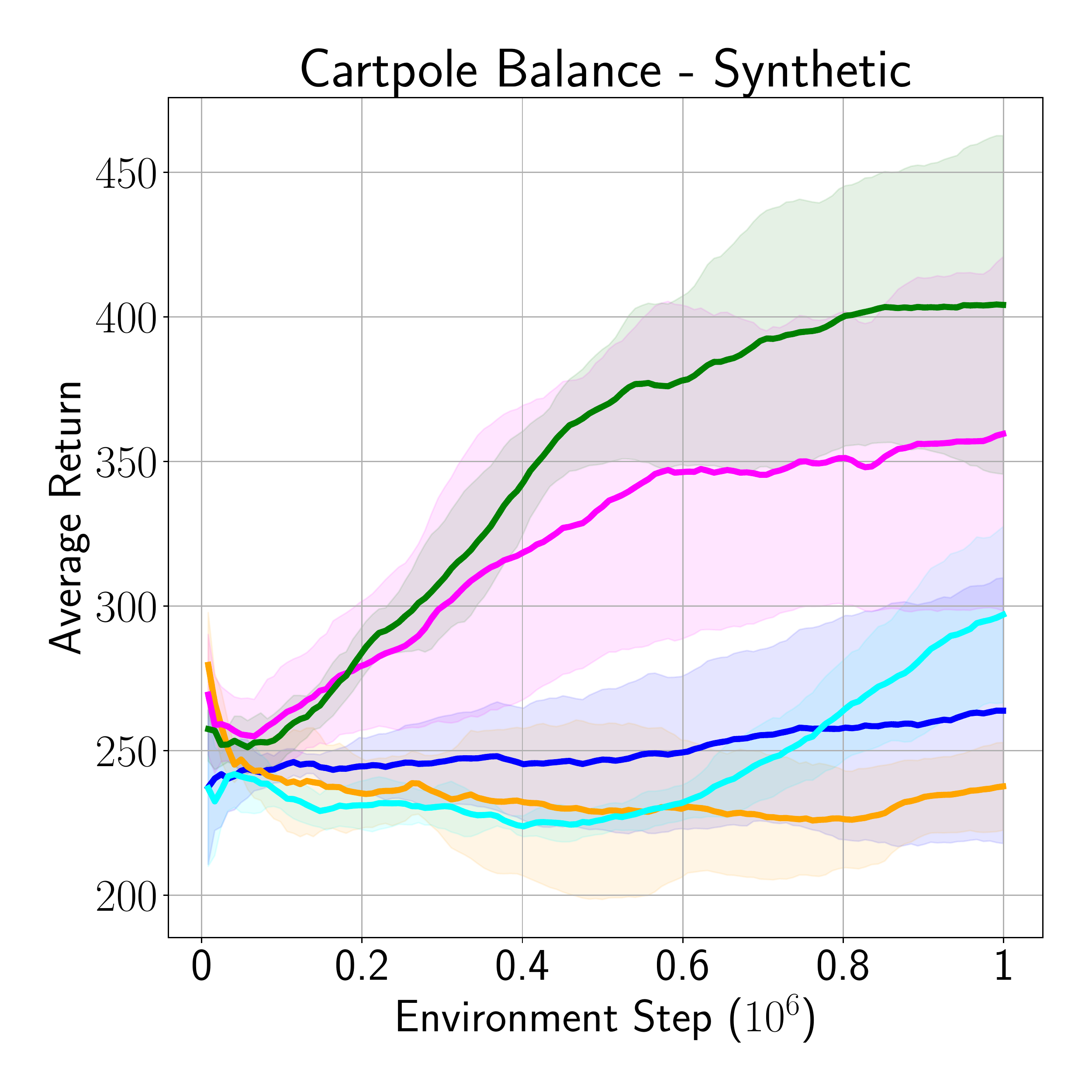}
    \includegraphics[width=0.24\textwidth]{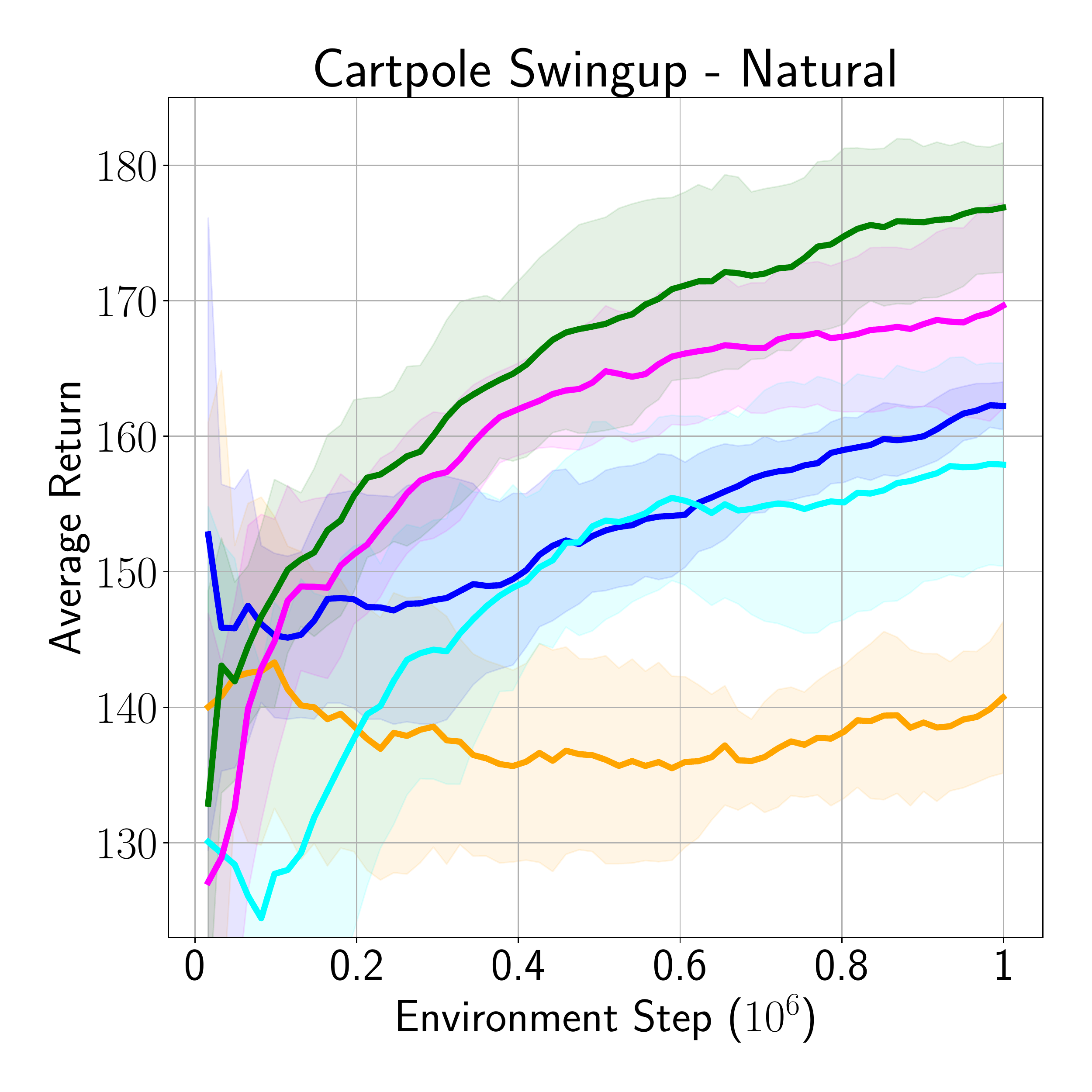}
    \includegraphics[width=0.24\textwidth]{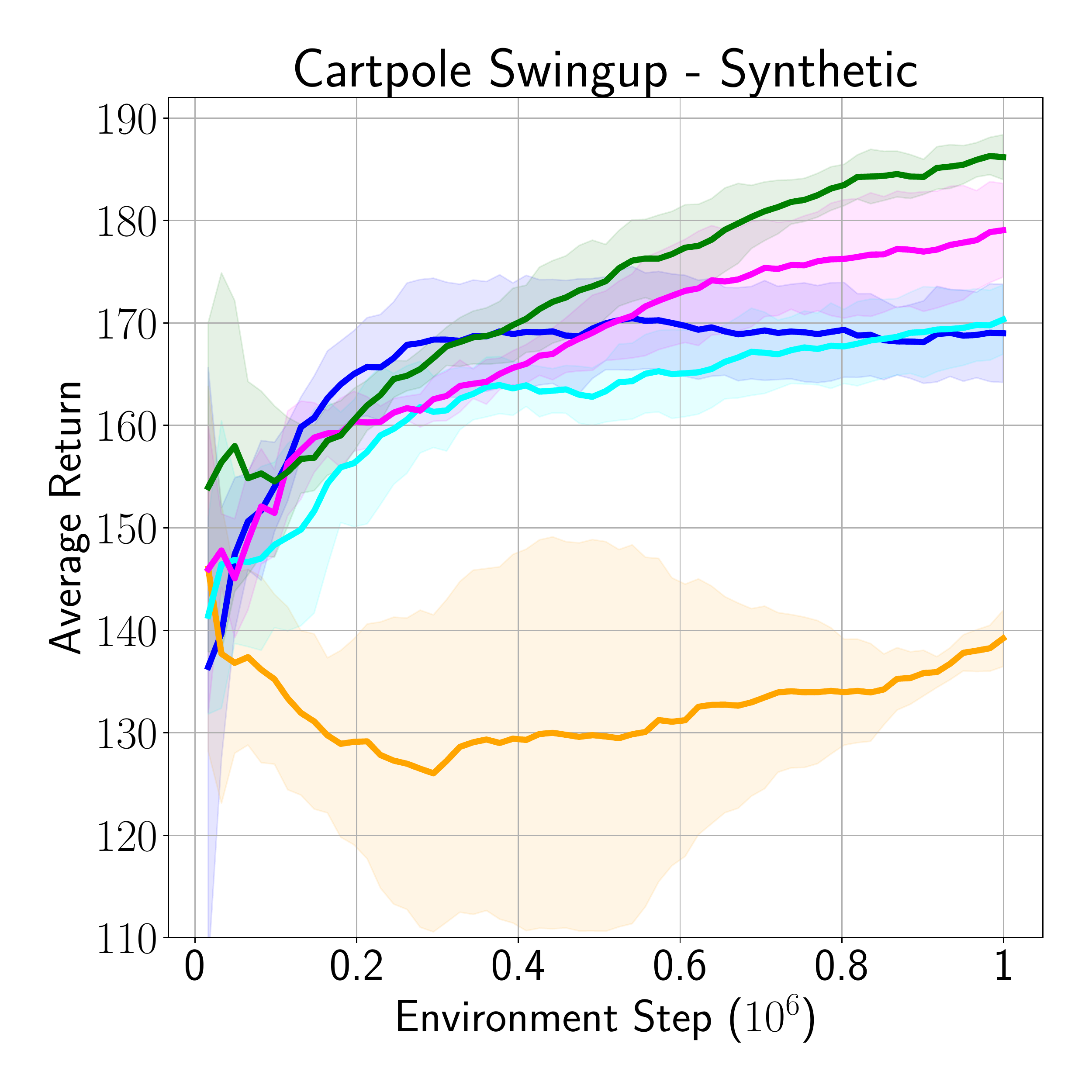}
    % \includegraphics[width=0.24\textwidth]{fig/dmc/ball-natural-ball_in_cup-catch-blog.pdf}
    % \includegraphics[width=0.24\textwidth]{fig/dmc/ball-synthetic-ball_in_cup-catch-blog.pdf}
    % \vspace{-6mm}
    \caption{\textbf{Average return on two DMC tasks, Cartpole Balance and Cartpole Swingup with natural and synthetic video backgrounds.} Our \gae{} and \ordergae{} approaches outperform \ppo{}, \ppg{}, and \ucbdrac{}. The mean and standard deviation are computed over 10 runs with different seeds. }
    % \vspace{-5mm}
    \label{fig:dmc_results1}
\end{figure*}

Table~\ref{tab:procgen_agg} shows the train and test performance of all methods, aggregated across all Procgen games. \gae{} and \ordergae{} outperform all the baselines on both train and test.  Figure~\ref{fig:procgen_results} shows the train and test performance on eight of the Procgen games. We show comparisons with a vanilla RL algorithm \ppo{}, the previous state-of-the-art \ucbdrac{} and our strongest baseline \ppg{}. Both of our approaches, \gae{} and \ordergae{}, show superior results on the test levels, relative to the other methods. In addition, \ordergae{} and \gae{} achieve better or comparable performance on the training levels for most of these games. While \gae{} already shows notable gains over the baselines, \ordergae{} further improves upon it, thus emphasizing the benefits of combining our two contributions. See Appendix~\ref{app:procgen_results} for results on all games.

% \vspace{-1mm}
\subsection{Ablations}
\label{sec:ablations}
% \vspace{-1mm}

We also performed a number of ablations to emphasize the importance of each component used by our method. First, \textbf{D}ecoupled \textbf{V}alue \textbf{A}ctor-\textbf{C}ritic or \textbf{\valuegae{}} is an ablation to \textbf{\gae{}} that learns to predict the value rather than the advantage for training the policy network. This ablation helps disentangle the effect of predicting the advantage function from the effect of using a separate value network and performing multiple updates for the value than for the policy. In principle, this decoupling could result in a more accurate estimate of the value function, which can in turn lead to more effective policy optimization. Second, \textbf{A}dvantage \textbf{A}ctor-\textbf{C}ritic or \textbf{\gaeppo{}} is a modification to PPO that includes an extra head for predicting the advantage function (in a single network). The role of this ablation is to understand the importance of \textit{not} backpropagating gradients from the value into the policy network, even while using gradients from the advantage. 

Figure~\ref{fig:procgen_ablations} shows the train and test performance of \gae{}, \valuegae{}, \gaeppo{}, and \ppo{} on four Procgen games. \gae{} outperforms all the ablations on both train and test environments, emphasizing the importance of each component. In particular, the fact that \gaeppo{}'s generalization ability is worse than that of \gae{} suggests that predicting the advantage function in addition to also predicting the value function (as part of training the policy network) does not solve the problem of overfitting. Hence, using the value loss to update the policy parameters still hurts generalization even when the advantage is also used to train the network. In addition, the fact that \valuegae{} has worse test performance than \gae{} indicates that \gae{}'s gains are not merely due to having access to a more accurate value function or to the reduced interference between optimizing the policy and value. 

These results are consistent with our claim that using gradients from the value to update the policy can lead to representations that overfit to spurious correlations in the training environments. In contrast, using gradients from the advantage to train the policy network leads to agents that generalize better to new environments. 
% In fact, as our results suggest, it is better to not use a value loss when updating the policy parameters in order to maximally reduce overfitting.

\begin{figure*}[ht!]
    \centering
    % \vspace{-3mm}
    \includegraphics[width=0.16\textwidth]{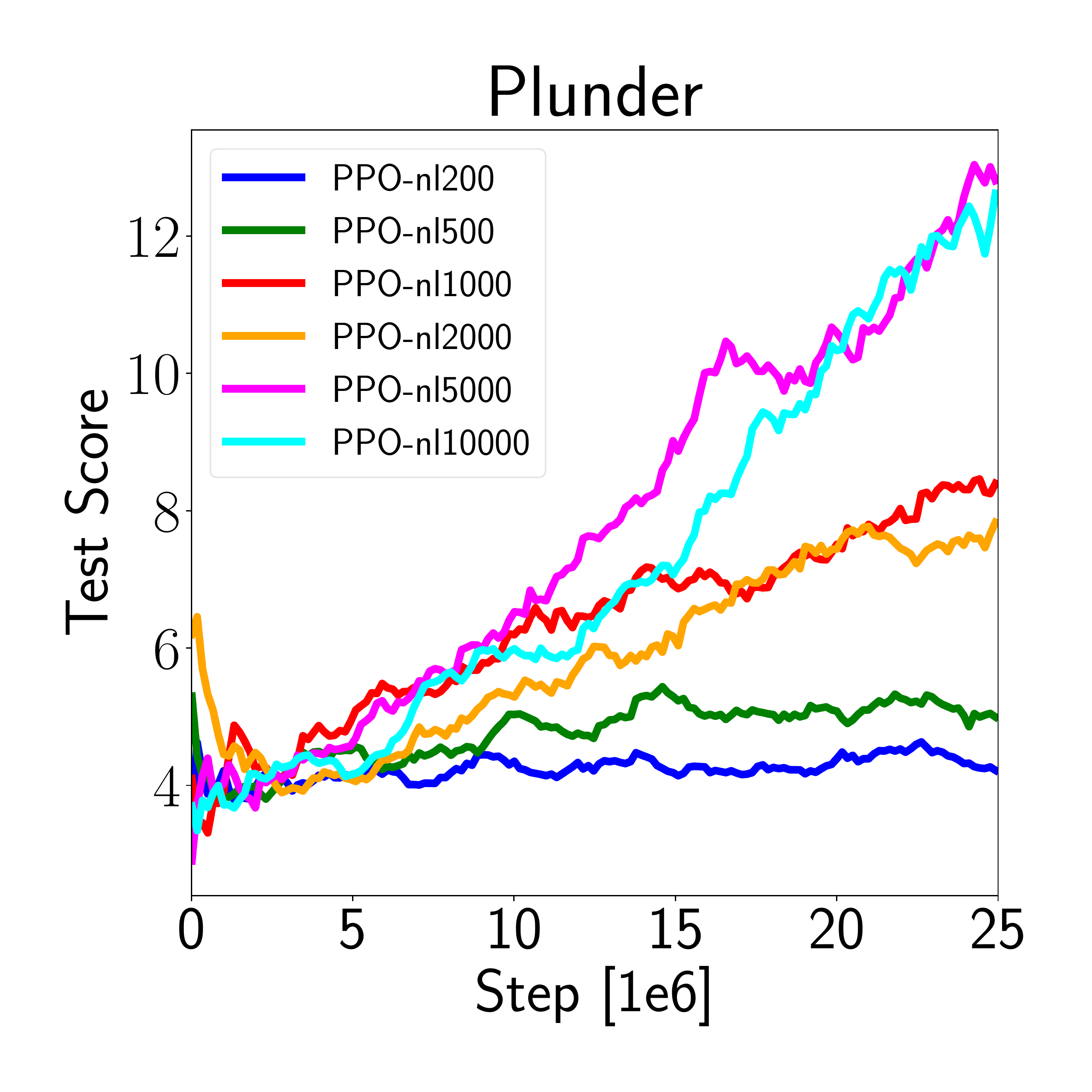}
    \includegraphics[width=0.16\textwidth]{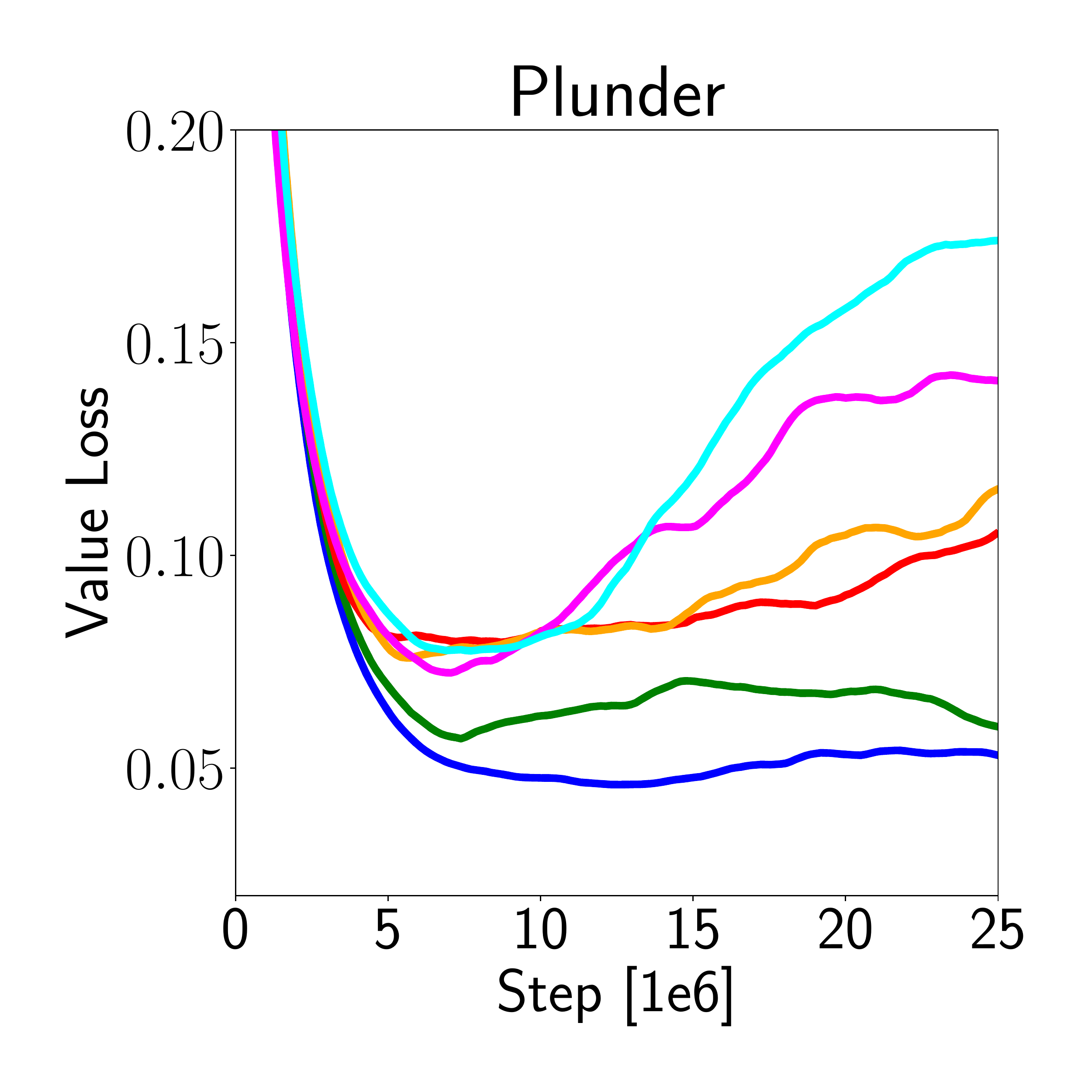}
    \includegraphics[width=0.15\textwidth]{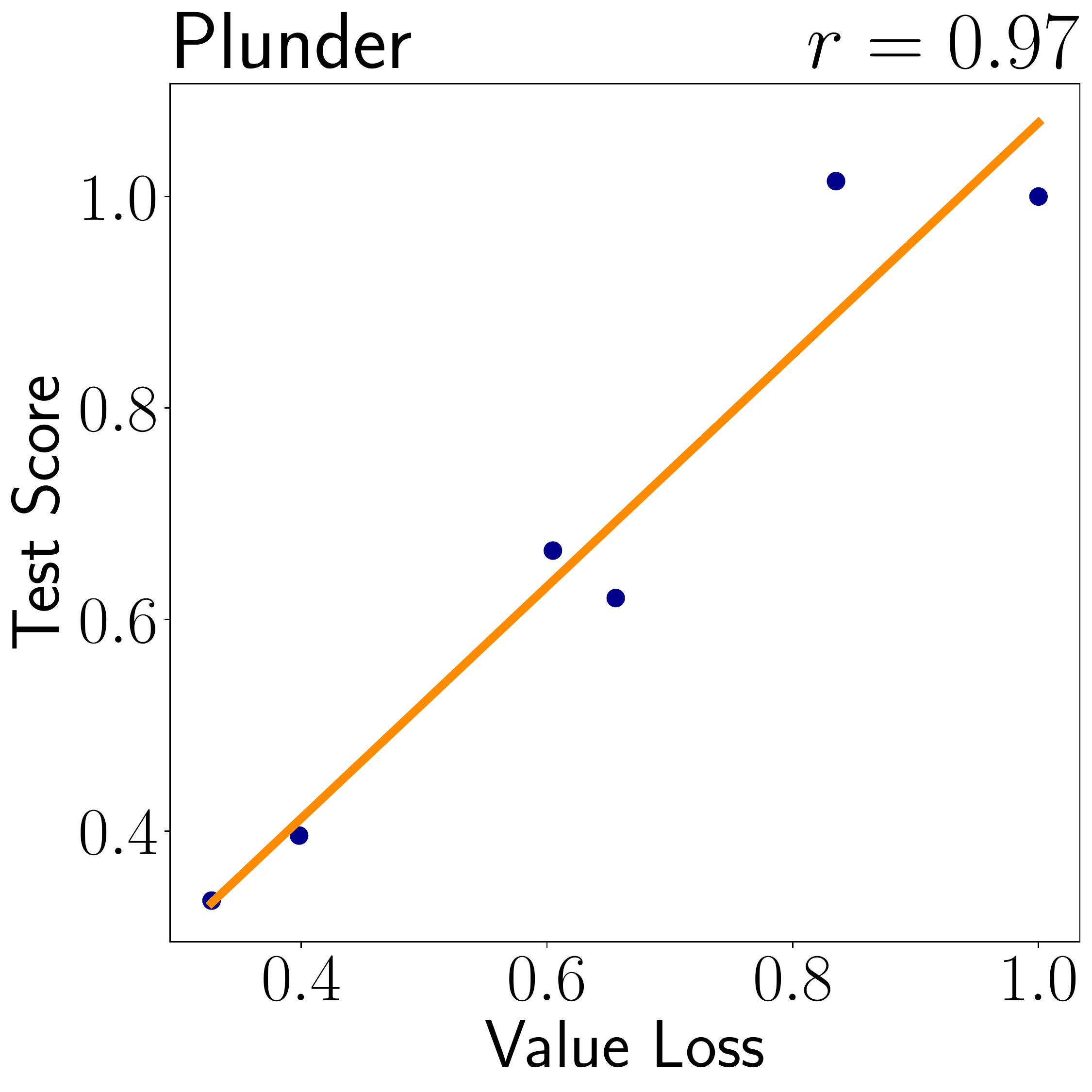}
    \includegraphics[width=0.16\textwidth]{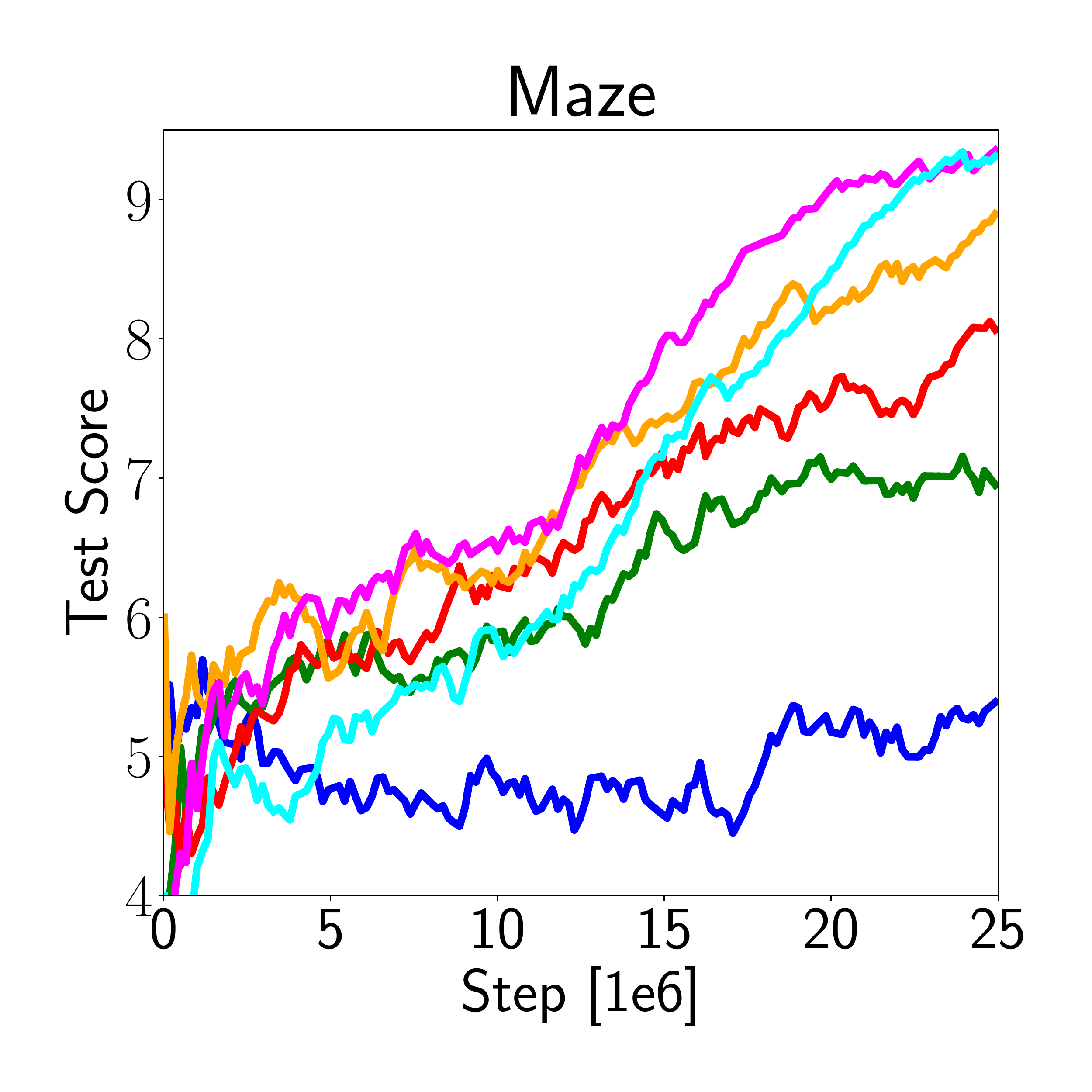}
    \includegraphics[width=0.16\textwidth]{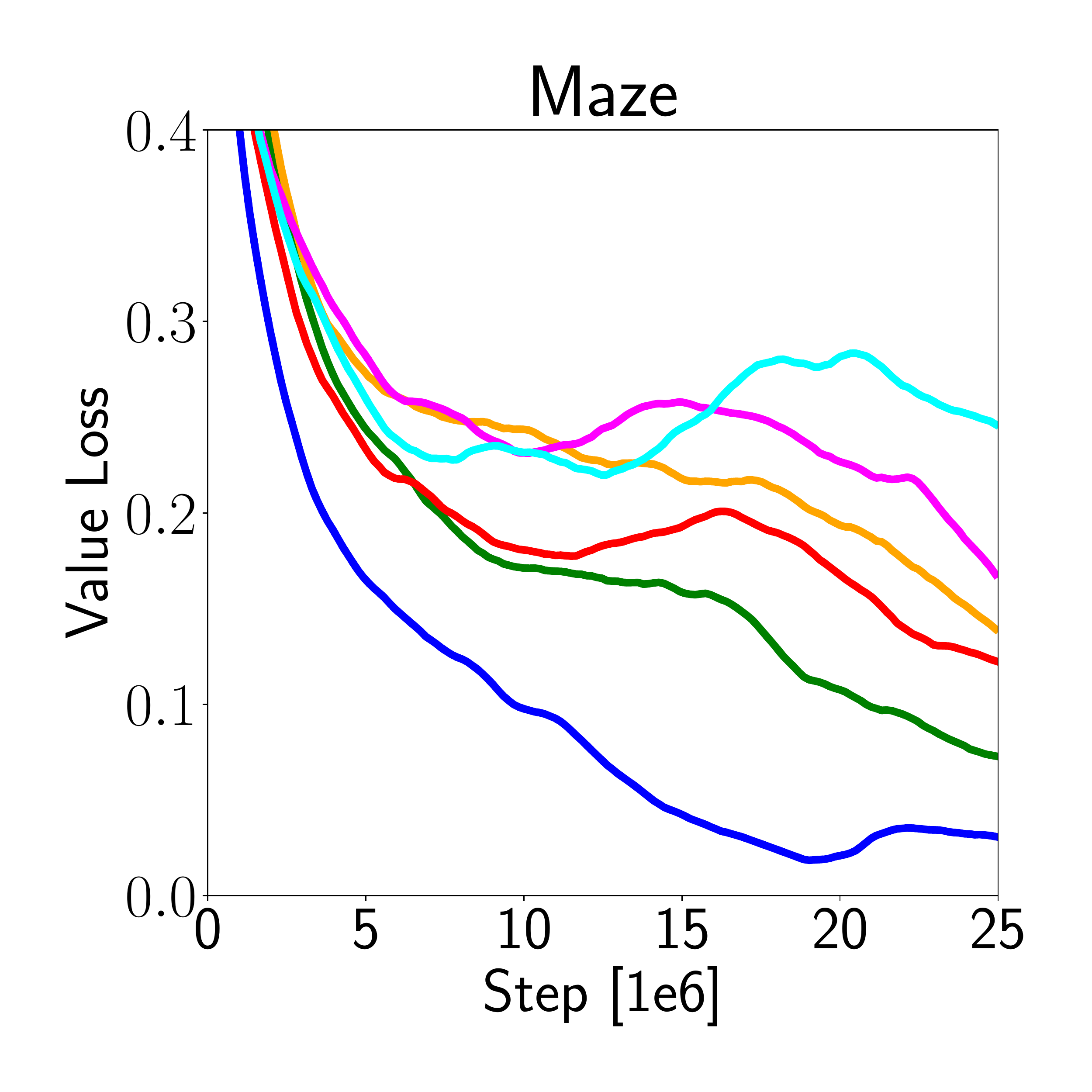}
    \includegraphics[width=0.15\textwidth]{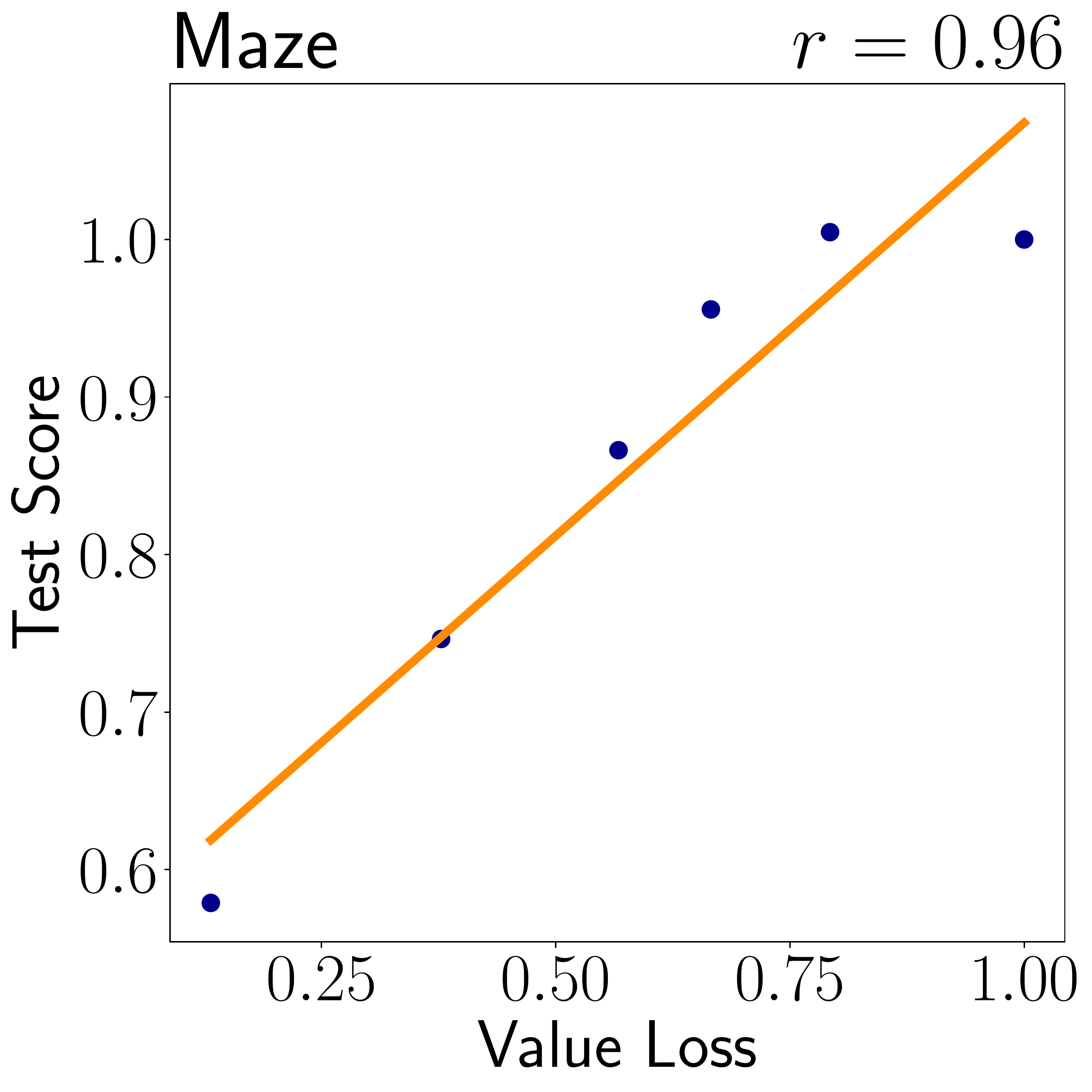}
    % \vspace{-5mm}
    \caption{\textbf{PPO agents trained on varying numbers of levels: 200, 500, 1000, 2000, 5000, and 10000 for Plunder (left) and Maze (right).} For each game, from left to right we show the test score, value loss, and correlation between value loss and test score after 25M training steps. Both the test score and the value loss increase with the number of training levels used. These results indicate there is a \textit{positive correlation between value loss and generalization} when using a shared network for the policy and value function.} 
    % \textit{models with a higher value loss generalize better}.}
    % \vspace{-3mm}
    \label{fig:corr_value_test}
\end{figure*}

%%%%%%%%%%%%%%%%%%%%%%%%%%%%%%%%%%%%%%%%%%%%%%%%%%%%%%%%%%%%%%%%%%%%%%%%
% \vspace{-1mm}
\subsection{DeepMind Control with Distractors}
\label{sec:procgen_results}
% \vspace{-1mm}

In this section, we evaluate our methods on the DeepMind Control Suite from pixels (DMC,  ~\citet{Tassa2018DeepMindCS}). We use three tasks, namely Cartpole Balance, Cartpole Swingup, and Ball In Cup. For each task, we study two settings with different types of backgrounds, namely \textit{synthetic} distractors and \textit{natural} videos from the Kinetics dataset~\citep{Kay2017TheKH}, as introduced in~\citet{Zhang2020LearningIR}. Note that in the synthetic and natural settings, the background is sampled from a list of videos at the beginning of each episode, which creates spurious correlations between backgrounds and rewards. As shown in Figure~\ref{fig:dmc_results1}, \gae{} and \ordergae{} significantly outperform \ppo{}, \ucbdrac{}, and \ppg{} on all these environments. See Appendix~\ref{app:dmc} for more details about the experimental setup and results on other DMC tasks.

%%%%%%%%%%%%%%%%%%%%%%%%%%%%%%%%%%%%%%%%%%%%%%%%%%%%%%%%%%%%%%%%%%%%%%%%
% \subsection{Policy-Value Representation Asymmetry}
% \label{sec:analysis}

% \vspace{-1mm}
\subsection{Value Loss and Generalization}
\label{sec:corr}
% \vspace{-1mm}

When using actor-critic policy-gradient algorithms, a more accurate estimate of the value function leads to lower variance gradients and thus better policy optimization on a given environment~\citep{Sutton1999PolicyGM}. While an accurate value function improves sample efficiency and training performance, it can also lead to overfitting when the policy and value share the same representation. To validate this claim, we looked at the correlation between value loss and test performance. More specifically, we trained 6 PPO agents on varying numbers of Procgen levels, namely 200, 500, 1000, 2000, 5000, and 10000. As expected, models trained on a larger number of levels generalize better to unseen levels as illustrated in Figure~\ref{fig:corr_value_test}. However, \textit{agents trained on more levels have a higher value loss at the end of training than agents trained on fewer levels}, so the \textit{value loss is positively correlated with generalization ability}. This observation is consistent with our claim that using a shared network for the policy and value function can lead to overfitting. Our hypothesis was that, when using a common network for the policy and value function, accurately predicting the values implies that the learned representation relies on spurious correlation, which would likely lead to poor generalization at test time. Similarly, an agent with good generalization suggests that its representation relies on the features needed to learn an optimal policy for the entire family of environments, which are insufficient for accurately predicting the value function (as explained in Section~\ref{sec:asym}). See Appendix~\ref{app:corr} for the relationship between value loss, test score, and the number of training levels for all Procgen games. In Appendix~\ref{app:val_var}, you can see that the variance of predicted values for the initial observations decreases with the number of training levels, which further supports our claim.

\begin{table*}[ht!]
%   \vspace{-2mm}
  \caption{\textbf{The trade-off between generalization and value accuracy illustrated on a single Ninja level.} A PPO agent trained on 200 levels (blue) has high value accuracy but low generalization performance, and its value predictions have a near linear dependency on the episode step. This linear relationship further supports the claim that the agent memorizes level-specific features which are needed to predict the value function given only partial observations. In contrast, a PPO agent trained on 10k levels (red) with good generalization but low value accuracy does not display this linear trend. When sharing parameters for the policy and value function, there is a trade-off between fitting the value and learning general policies. By decoupling the policy and value, our model \gae{} can achieve both high value accuracy and good generalization performance. To train the policy, \gae{} uses gradients from predicting the advantage (green), which does not display the linear trend, thus is less prone to overfitting. \gae{}'s value estimate (orange) still shows a linear trend  but, in contrast to \ppo{}, this does not negatively affect the policy since \gae{} uses separate networks to learn the policy and value.}
  \small
  \begin{tabular}{>{}m{0.61in} >{}m{1.0in}| >{}m{0.61in} >{}m{1.0in}| >{}m{0.61in} >{}m{0.9in} >{}m{0.9in}}
    \toprule \multicolumn{2}{c}{PPO-200} & \multicolumn{2}{c}{PPO-10k} & \multicolumn{3}{c}{DAAC-200}  \\
    % \toprule PPO-200 & PPO-200 & PPO-10k & PPO-10k & DAAC-200  & DAAC-200  & DAAC-200  \\
    \midrule
    \begin{tabular}{@{}c@{}} Test Score: 5.9  \\ Value Loss: 0.2 \end{tabular} & \hspace{-2.1mm}
    \includegraphics[width=0.16\textwidth]{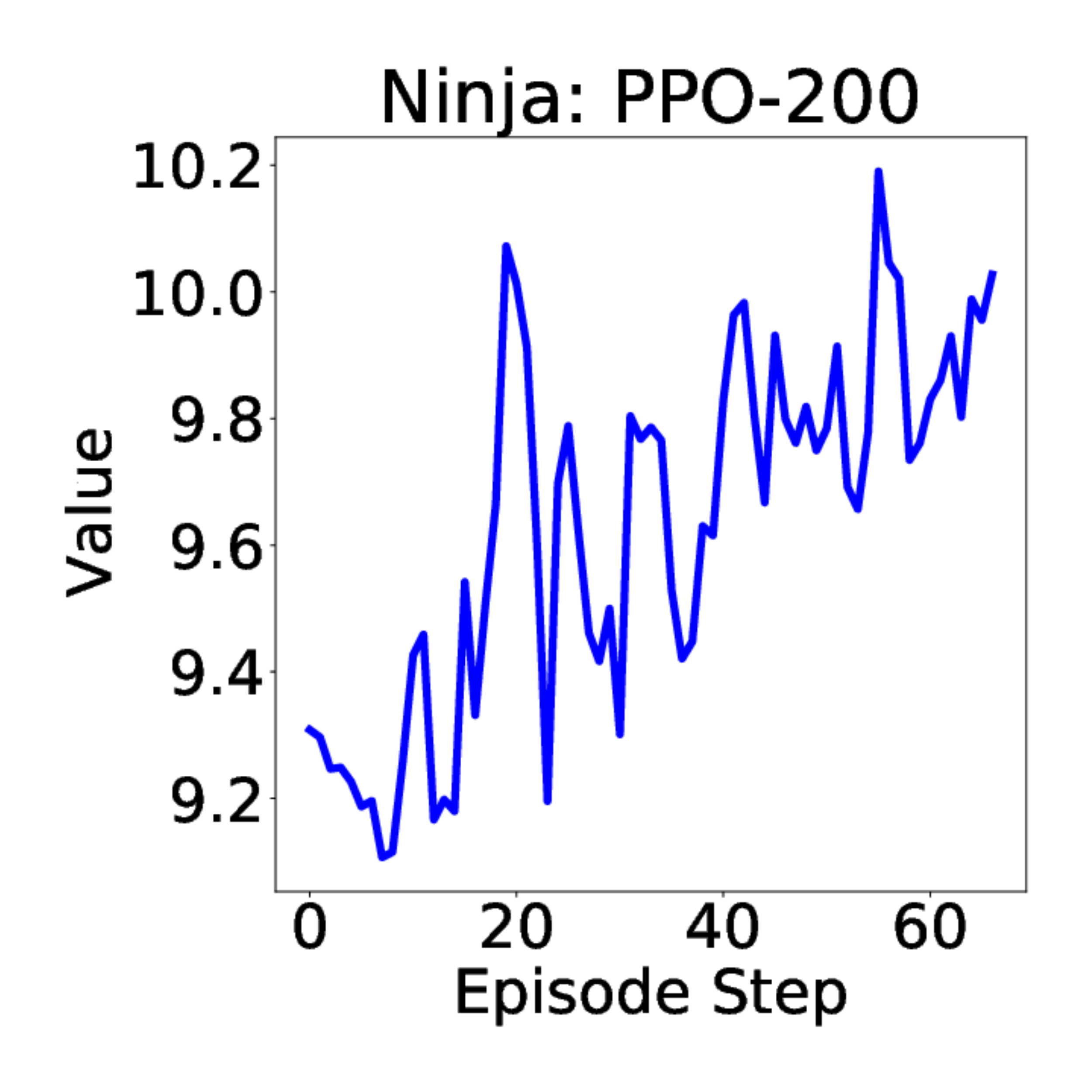} & 
    
    \begin{tabular}{@{}c@{}} Test Score: 8.8  \\ Value Loss: 0.3 \end{tabular} & \hspace{-2.1mm}
     \includegraphics[width=0.16\textwidth]{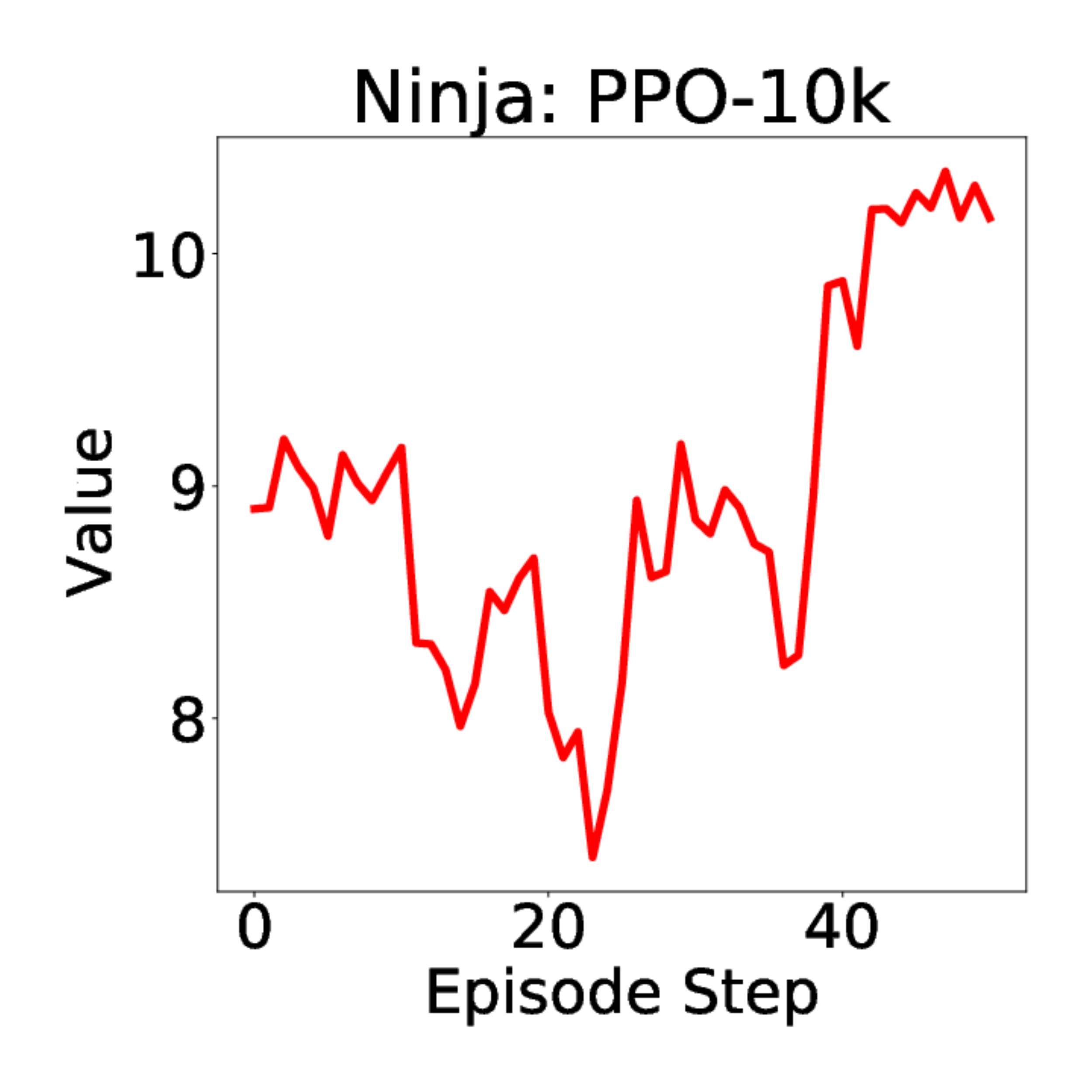} & 
    
    \begin{tabular}{@{}c@{}}Test Score: 7.3  \\ Value Loss: 0.2 \end{tabular} & \hspace{-2.1mm}
     \includegraphics[width=0.16\textwidth]{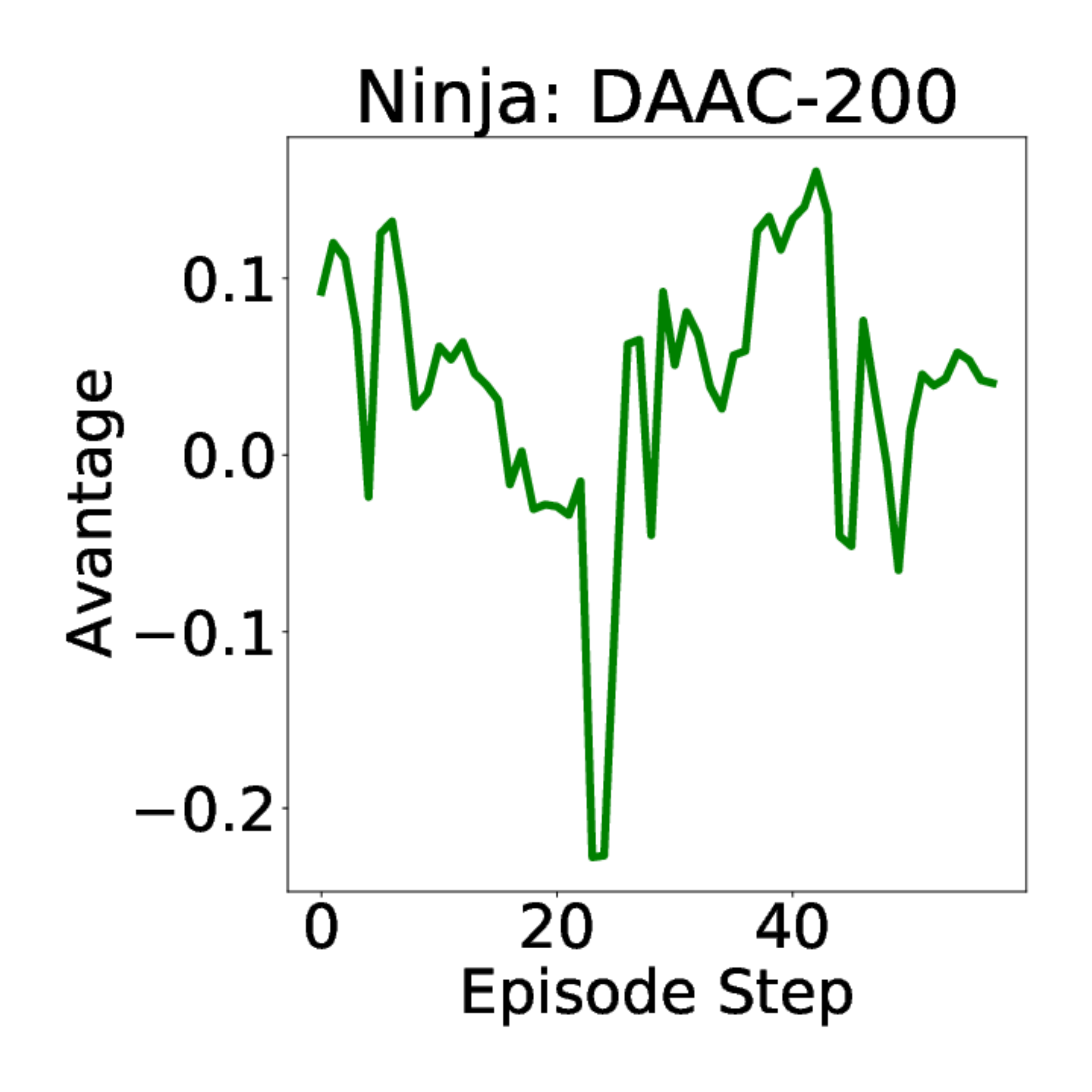} & \hspace{-2.0mm}
      \includegraphics[width=0.16\textwidth]{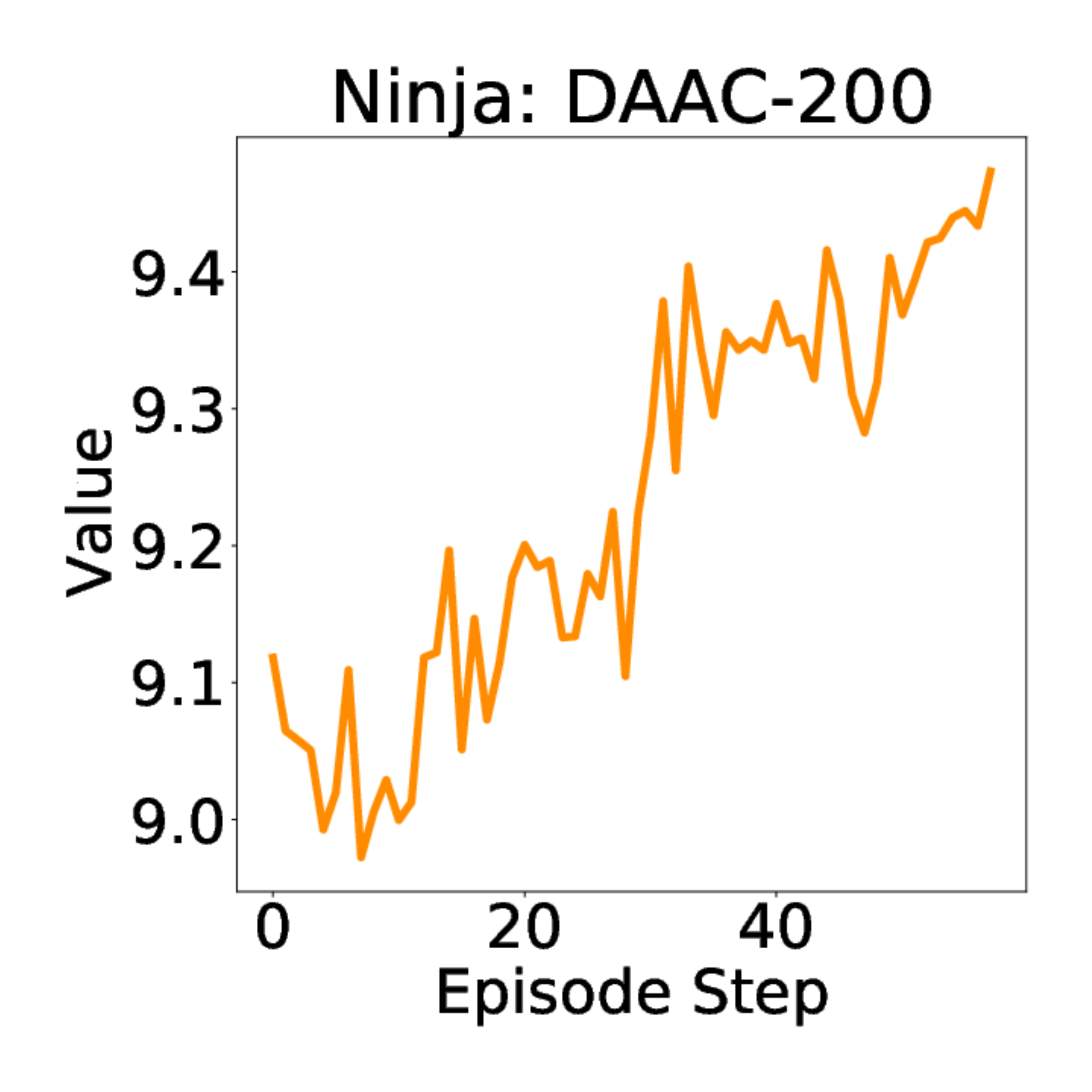} \\
    %   \hline
      \bottomrule
  \end{tabular}
  \label{tab:step_value_gae}
%   \vspace{-6.5mm}
  \vspace{-1mm}
\end{table*}

% \vspace{-1mm}
\subsection{Advantage vs. Value During an Episode}
\label{sec:adv_val}

In Section~\ref{sec:intro}, we claim that in procedurally generated environments with partial observability, in order to accurately estimate the value function, the agent needs to memorize the number of remaining steps in the level. As Figure~\ref{fig:ninja_levels} shows, a standard RL agent predicts very different values for the initial states of two levels even if the observations are semantically identical. This suggests that the agent must have memorized the length of each level since the partial observation at the beginning of a level does not contain enough information to accurately predict the expected return.

We further investigate this issue by exploring the predicted value's dependency on the episode step. Instead of just comparing the initial states as in Figure~\ref{fig:ninja_levels}, we plot the value predicted by the agent over the course of an entire trajectory in one of the Ninja levels (see Table~\ref{tab:step_value_gae}). Since the agent is rewarded only if it reaches the goal at the end of the episode, the true value increases linearly with the episode step over the course of the agent's trajectory. As seen in Table~\ref{tab:step_value_gae}, the estimated value of a PPO agent trained on 200 levels also has a quasi-linear dependence on the episode step, suggesting that the agent must know how many remaining steps are in the game at any point during this episode. However, the episode step cannot be inferred solely from partial observations since the training levels contain observations with the same semantics but different values, as illustrated in Figure~\ref{fig:ninja_levels}. In order to accurately predict the values of such observations, the agent must learn representations that capture level-specific features (such as the backgrounds) which would allow it to differentiate between semantically similar observations with different values. Since PPO uses a common representation for the policy and value function, this can lead to policies that overfit to the particularities of the training environments. Note that a PPO agent trained on 10k levels does not show the same linear trend between the value and episode step. This implies that there is a trade-off between generalization and value accuracy for models that use a shared network to learn the policy and value.

By decoupling the policy and value, our model \gae{} can achieve both high value accuracy and good generalization performance. Note that the advantage estimated by \gae{} (trained on 200 levels) shows no clear dependence on the environment step, thus being less prone to overfitting. Nevertheless, \gae{}'s value estimate  still shows a linear trend  but, in contrast to \ppo{}, this does not negatively affect the policy since \gae{} uses separate networks for the policy and value. This analysis indicates that using advantages rather than values to update the policy network leads to better generalization performance while also being able to accurately predict the value function. See Appendix~\ref{app:adv_val} for similar results on other Procgen games, as well as comparisons with \ppg{} which displays a similar trend as \ppo{} trained on 200 levels. 

\section{Discussion}
% \vspace{-7mm}

In this work, we identified a new problem with standard deep reinforcement learning algorithms which causes overfitting, namely the asymmetry between the policy and value representation. To alleviate this problem, we propose \ordergae{}, which decouples the optimization of the policy and value function while still learning effective behaviors. IDAAC also introduces an auxiliary loss which constrains the policy representation to be invariant with respect to the environment instance. \ordergae{} achieves a new state-of-the-art on the Procgen benchmark and outperforms strong RL algorithms on DeepMind Control tasks with distractors. In contrast to other popular methods, our approach can both achieve good generalization while also learning accurate value estimates. Moreover, \ordergae{} learns representations and predictions which are more robust to cosmetic changes in the observations that do not change the underlying state of the environment (see Appendix~\ref{app:robustness}). . 

% \textbf{Limitations.}
One limitation of our work is the focus on learning representations which are invariant to the number of remaining steps in the episode. While this inductive bias will not be helpful for all problems, the settings where we can expect most gains are those with partial observability, a set of goal states, and episode length variations (\eg{} navigation of different layouts). A promising avenue for future work is to investigate other auxiliary losses in order to efficiently learn more general behaviors. One desirable property of such auxiliary losses is to capture the minimal set of features needed to act in the environment. While our experiments show that predicting the advantage function improves generalization, we currently lack a firm theoretical argument for this. The advantage could act as a regularizer, being less prone to memorizing the remaining episode length, or it could be better correlated with the underlying state of the environment rather than its visual appearance. Investigating these hypotheses could further improve our understanding of what leads to better representations and what we are still missing. Finally, the solution we propose here is only a first step towards solving the policy-value representation asymmetry and we hope many other ideas will be explored in future work.

% % Acknowledgements should only appear in the accepted version.
% \vspace{-4mm}
\section*{Acknowledgements}
% \vspace{-1mm}
We would like to thank our ICML reviewers, as well as Vitaly Kurin, Denis Yarats, Mahi Shafiullah, Ilya Kostrikov, and Max Goldstein for their valuable feedback on this work. Roberta was supported by the DARPA Machine Commonsense program.

\clearpage
\bibliography{main}
\bibliographystyle{icml2021}

\appendix
\appendix
\onecolumn

%%%%%%%%%%%%%%%%%%%%%%%%%%
\clearpage
\section{PPO}
\label{app:ppo}
\textbf{Proximal Policy Optimization} (PPO) \cite{schulman2017proximal} is an actor-critic algorithm that learns a policy $\pi_{\theta}$ and a value function $V_{\theta}$ with the goal of finding an optimal policy for a given MDP. PPO alternates between sampling data through interaction with the environment and optimizing an objective function using stochastic gradient ascent. At each iteration, PPO maximizes the following objective: 
\begin{equation}
\label{eq:ppo_loss}
    J_{\mathrm{PPO}} = J_{\pi} - \alpha_1 J_{V} + \alpha_2 S_{\pi_\theta},
\end{equation}
where $\alpha_1$, $\alpha_2$ are weights for the different loss terms, $S_{\pi_\theta}$ is the entropy bonus for aiding exploration, $J_{V}$ is the value function loss defined as
\begin{equation*}
    J_{V} = \left(V_{\theta}(s) - V_t^{\mathrm{target}}\right)^2.
\end{equation*}

The policy objective term $J_{\pi}$ is based on the policy gradient objective which can be estimated using importance sampling in off-policy settings (\ie{} when the policy used for collecting data is different from the policy we want to optimize):
\begin{equation}
\label{eq:pg_loss}
    J_{PG}(\theta) = \sum_{a \in \mathcal{A}} \pi_{\theta}(a | s) \hat{A}_{\theta_{\mathrm{old}}}(s, a) = 
    \mathbb{E}_{a\sim \pi_{\theta_{\mathrm{old}}}} 
        \left[ \frac{\pi_{\theta} (a | s)}{\pi_{\theta_{\mathrm{old}}}(a | s)} \hat{A}_{\theta_{\mathrm{old}}}(s, a) \right], 
\end{equation}
where $\hat{A}(\cdot)$ is an estimate of the advantage function, $\theta_{old}$ are the policy parameters before the update, $\pi_{\theta_{old}}$ is the behavior policy used to collect trajectories (\ie{} that generates the training distribution of states and actions), and $\pi_{\theta}$ is the policy we want to optimize (\ie{} that generates the true distribution of states and actions).  

This objective can also be written as
\begin{equation}
    J_{PG}(\theta) = \mathbb{E}_{a\sim \pi_{\theta_{\mathrm{old}}}} 
        \left[ r(\theta) \hat{A}_{\theta_{\mathrm{old}}}(s, a) \right], 
\end{equation}
where
\begin{equation*}
\label{eq:imp_weight}
    r_\theta = \frac{\pi_{\theta}(a | s)}{\pi_{\theta_{\mathrm{old}}}(a | s)}
\end{equation*}
is the importance weight for estimating the advantage function. 

PPO is inspired by TRPO~\citep{Schulman2015TrustRP}, which constrains the update so that the policy does not change too much in one step. This significantly improves training stability and leads to better results than vanilla policy gradient algorithms. TRPO achieves this by minimizing the KL divergence between the old (\ie{} before an update) and the new (\ie{} after an update) policy. PPO implements the constraint in a simpler way by using a clipped surrogate objective instead of the more complicated TRPO objective. More specifically, PPO imposes the constraint by forcing $r(\theta)$ to stay within a small interval around 1, precisely $[1 - \epsilon, 1 + \epsilon]$, where $\epsilon$ is a hyperparameter. The policy objective term from equation~\eqref{eq:ppo_loss} becomes
\begin{equation*}
    J_{\pi} = \mathbb{E}_{\pi}\left[\min\left(r_\theta\hat{A}, \; \mathrm{clip}\left(r_\theta, 1 - \epsilon, 1 + \epsilon\right)\hat{A}\right)\right],
\end{equation*}
where $\hat{A} =  \hat{A}_{\theta_{\mathrm{old}}}(s, a)$ for brevity.
The function clip($r(\theta), 1 - \epsilon, 1 + \epsilon$) clips the ratio to be no more than $1 + \epsilon$ and no less than $1 - \epsilon$. The objective function of PPO takes the minimum one between the original value and the clipped version so that agents are discouraged from increasing the policy update to extremes for better rewards.

%%%%%%%%%%%%%%%%%%%%%%%%%%
\clearpage
\section{DAAC and IDAAC}
\label{app:algos}

As described in the paper, \gae{} and \ordergae{} alternate between optimizing the policy network and optimizing the value network. The value estimates are used to compute the advantage targets which are needed by both the policy gradient objective and the auxiliary loss based on predicting the advantage function. Since we use separate networks for learning the policy and value function, we can now use different numbers of epochs for updating the two networks. We use $E_\pi$ epochs for every policy update and $E_V$ epochs for every value update. Similar to~\citet{Cobbe2020PhasicPG}, we find that the value network allows for larger amounts of sample reuse than the policy network. This decoupling of the policy and value optimization allows us to also control the frequency of value updates relative to policy updates. Updating the value function less often can help with training stability since can result in lower variance gradients for the policy and advantage losses. We update the value network every $N_\pi$ updates of the policy network. See algorithms~\ref{alg:daac} and~\ref{alg:idaac} for the pseudocodes of \gae{} and \ordergae{}, respectively.

\begin{algorithm}
    \caption{\textbf{DAAC}: Decoupled Advantage Actor-Critic}
    \label{alg:daac}
    \begin{spacing}{1.0}
	\begin{algorithmic}[1]
	    \State \textbf{Hyperparameters:} Total number of updates N, replay buffer size T, number of epochs per policy update $E_\pi$, number of epochs per value update $E_V$, frequency of value updates $N_\pi$, weight for the advantage loss $\alpha_a$, initial policy parameters $\theta$, initial value parameters $\phi$.
        \For{$n=1,\ldots,N$}
            \State Collect $\mathcal{D} = \{(s_t, a_t, r_t, s_{t+1})\}_{t=1}^T$ using $\pi(\theta)$.
            \State Compute the value and advantage targets $\hat{V}_t$ and $\hat{A}_t$ for all states $s_t$
            \For{$i=1,\ldots,E_\pi$}
                \State $ L_{\mathrm{A}}(\theta) = \hat{\mathbb{E}}_{t} \left[\left(A_{\theta}(s_t, a_t) - \hat{A_t}\right)^2\right]$\Comment {Compute the Advantage Loss}
                \State $J_{\mathrm{\gae{}}}(\theta) =  J_{\pi}(\theta) +         \alpha_\mathrm{s}\mathrm{S}_{\pi}(\theta) - \alpha_\mathrm{a} L_{\mathrm{A}}(\theta)$ \Comment {Compute the Policy Loss}
            	\State $\theta\leftarrow\argmax_{\theta} J_{\mathrm{\gae{}}}$  \Comment {Update the Policy Network}
            \EndFor
            \If{$n \; \% \; N_\pi \; = \; 0$}
                 \For{$j=1,\ldots,E_V$}
                    \State  $L_{\mathrm{V}}(\phi) = \hat{\mathbb{E}}_{t} \left[\left(V_{\phi}(s_t) - \hat{V_t}\right)^2\right]$  \Comment {Compute the Value Loss}    
                	\State $\phi\leftarrow\argmin_{\phi} L_{\mathrm{V}}$  \Comment {Update the Value Network}
                \EndFor
            \EndIf
        \EndFor
	\end{algorithmic} 
    \end{spacing} 
\end{algorithm}

\begin{algorithm}
    \caption{\textbf{IDAAC}: Invariant Decoupled Advantage Actor-Critic}
    \label{alg:idaac}	
        \begin{spacing}{1.0}
    	\begin{algorithmic}[1]
	    \State \textbf{Hyperparameters:} Total number of updates N, replay buffer size T, number of epochs per policy update $E_\pi$, number of epochs per value update $E_V$, frequency of value updates $N_\pi$, weight for the invariance loss $\alpha_i$, initial policy parameters $\theta$, initial value parameters $\phi$.
        \For {$n=1,\ldots,N$}
            \State Collect $\mathcal{D} = \{(s_t, a_t, r_t, s_{t+1})\}_{t=1}^T$ using $\pi(\theta)$.
            \State Compute the value and advantage targets $\hat{V}_t$ and $\hat{A}_t$ for all states $s_t$
            \For {$i=1,\ldots,E_\pi$}
                \State $ L_{\mathrm{A}}(\theta) = \hat{\mathbb{E}}_{t} \left[\left(A_{\theta}(s_t, a_t) - \hat{A_t}\right)^2\right]$\Comment {Compute the Advantage Loss}
                \State $L_{\mathrm{E}}(\theta) = - \frac{1}{2} \mathrm{log} \left[ \mathrm{D}_\psi\left(\mathrm{E}_\theta(s_i), \mathrm{E}_\theta(s_j)\right) \right] -
                        \frac{1}{2} \mathrm{log}\left[1 - \mathrm{D}_\psi\left(\mathrm{E}_\theta(s_i), \mathrm{E}_\theta(s_j)\right)\right]$ \Comment {Compute the Encoder Loss}
                \State $J_{\mathrm{\ordergae{}}}(\theta, \phi, \psi) = J_{\pi}(\theta) +         \alpha_\mathrm{s}\mathrm{S}_{\pi}(\theta) - \alpha_\mathrm{a} L_{\mathrm{A}}(\theta) - \alpha_i L_{\mathrm{E}}(\theta)$ \Comment {Compute the Policy Loss}
                \State $L_{\mathrm{D}}(\psi) = - \mathrm{log} \left[ \mathrm{D}_\psi\left(\mathrm{E}_\theta(s_i), \mathrm{E}_\theta(s_j)\right) \right] -
                        \mathrm{log}\left[1 - \mathrm{D}_\psi\left(\mathrm{E}_\theta(s_i), \mathrm{E}_\theta(s_j)\right)\right]$ \Comment {Compute the Discriminator Loss}
            	\State $\theta\leftarrow\argmax_{\theta} J_{\mathrm{\ordergae{}}}$  \Comment {Update the Policy Network}
            	\State $\psi\leftarrow\argmin_{\psi} L_{\mathrm{D}}$  \Comment {Update the Discriminator}
            \EndFor
            
             \If {$n \; \% \; N_\pi \; = \; 0$}
                \For {$j=1,\ldots,E_V$}
                    \State  $L_{\mathrm{V}}(\phi) = \hat{\mathbb{E}}_{t} \left[\left(V_{\phi}(s_t) - \hat{V_t}\right)^2\right]$ \Comment {Compute the Value Loss}    
                	\State $\phi\leftarrow\argmin_{\phi} L_{\mathrm{V}}$ \Comment {Update the Value Network}
                \EndFor
            \EndIf
        \EndFor
	\end{algorithmic} 
    \end{spacing}
\end{algorithm}

%%%%%%%%%%%%%%%%%%%%%%%%%%
\clearpage
\section{Hyperparameters}
\label{app:hyperparams}

We use~\citet{pytorchrl}'s implementation of PPO~\citep{schulman2017proximal}, on top of which all our methods are build. The agent is parameterized by the ResNet architecture from~\citet{espeholt2018impala} which was used to obtain the best results in~\citet{cobbe2019leveraging}. Unless otherwise noted, we use the best hyperparameters found in~\citet{cobbe2019leveraging} for the easy mode of Procgen (\ie{} same experimental setup as the one used here) as found in Table~\ref{tab:hps}:
\begin{table}[t!]
    \centering
    \small
    \caption{List of hyperparameters used to obtain the results in this paper.}
    \begin{tabular}{cc}
    \toprule
    Hyperparameter & Value \\ [0.5ex]
    \midrule
     $\gamma$ & 0.999 \\ [0.5ex]
     $\lambda$ & 0.95 \\ [0.5ex]
      \# timesteps per rollout & 256 \\ [0.5ex]
      \# epochs per rollout & 3 \\ [0.5ex]
      \# minibatches per epoch & 8 \\ [0.5ex]
      entropy bonus & 0.01 \\ [0.5ex]
      clip range & 0.2 \\ [0.5ex]
      reward normalization & yes \\ [0.5ex]
      learning rate & 5e-4 \\ [0.5ex]
      \# workers & 1 \\ [0.5ex]
      \# environments per worker & 64 \\ [0.5ex]
      \# total timesteps & 25M \\ [0.5ex]
      optimizer & Adam \\ [0.5ex]
      LSTM & no \\ [0.5ex]
      frame stack & no \\ [0.5ex]
    \bottomrule
    \end{tabular}
    % \hline
    \label{tab:hps}
\end{table}

We ran a hyperparameter search over the number of epochs used during each update of the policy network $E_\pi \in [1, 3, 6]$ the number epochs used during each update of the value network $E_V \in [1, 5, 9]$, the number of value updates after which we perform a policy update $N_\pi \in [1, 8, 32]$, the weight for the advantage loss $\alpha_a \in [0.05, 0.1, 0.15, 0.2, 0.25, 0.3]$, and the weight for the instance-invariant (adversarial) loss $\alpha_i \in [0.001, 0.005, 0.01, 0.05, 0.1, 0.2]$. We found $E_\pi = 1, E_V = 9, N_\pi = 1, \alpha_a = 0.25, \alpha_i = 0.001$ to be the best hyperparameters overall, so we used these values to obtain the results reported in the paper. For all our experiments we use the Adam~\citep{Kingma2015AdamAM} optimizer. 

% \todo{add HP search for our method and the best HPs found and used for the results in the paper} 
% \todo{add HPs used for all the baselines, PPG, Mixreg, PLR, ablations etc.} 

For all baselines, we use the same experimental setup for training and testing as the one used for our methods. Hence, we train them for 25M frames on the easy mode of each Procgen game, using (the same) 200 levels for training and the entire distribution of levels for testing. 

For \mixreg{}, \plr{}, \ucbdrac{}, and \ppg{}, we used the best hyperparameters reported by the authors, since all these methods use Procgen for evaluation and they performed extensive hyperparameter sweeps. 

For \randfm{} \cite{lee2020network} we use the recommended hyperparameters in the authors' released implementation, which were the best values for CoinRun~\citep{cobbe2018quantifying}, one of the Procgen games used for evaluation in~\citep{lee2020network}. 

For \ibacsni{} \cite{igl2019generalization} we also use the authors' open sourced implementation. We use the parameters corresponding to IBAC-SNI $\lambda=.5$. We use weight regularization with $l_2 = .0001$, data augmentation turned on, and a value of $\beta = .0001$ which turns on the variational information bottleneck, and selective noise injection turned on. This corresponds to the best version of this approach, as found by the authors after evaluating it on CoinRun~\citep{cobbe2018quantifying}.

%%%%%%%%%%%%%%%%%%%%%%%%%%
\clearpage
\section{Procgen Results}
\label{app:procgen_results}

\begin{figure*}[ht!]
    \centering
    \includegraphics[width=\textwidth]{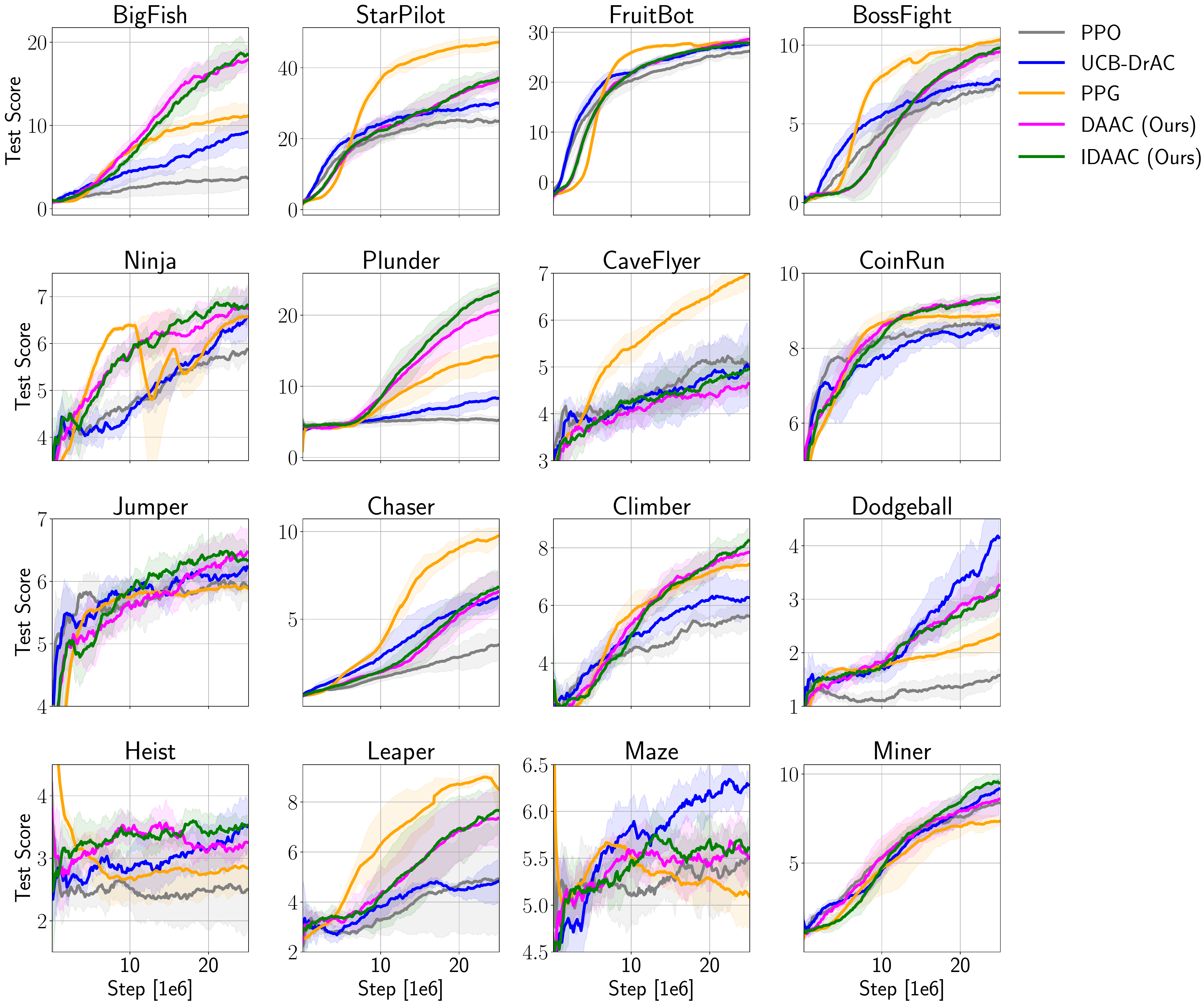}
    \caption{\textbf{Test Performance} of \ordergae{}, \gae{}, \ppg{}, \ucbdrac{}, and \ppo{} on all Procgen games. \ordergae{} outperforms the other methods on most games and is significantly better than \ppo{}. The mean and standard deviation are computed over 10 runs with different seeds.}
    \label{fig:results_test}
\end{figure*}

\begin{figure*}[ht!]
    \centering
    \includegraphics[width=\textwidth]{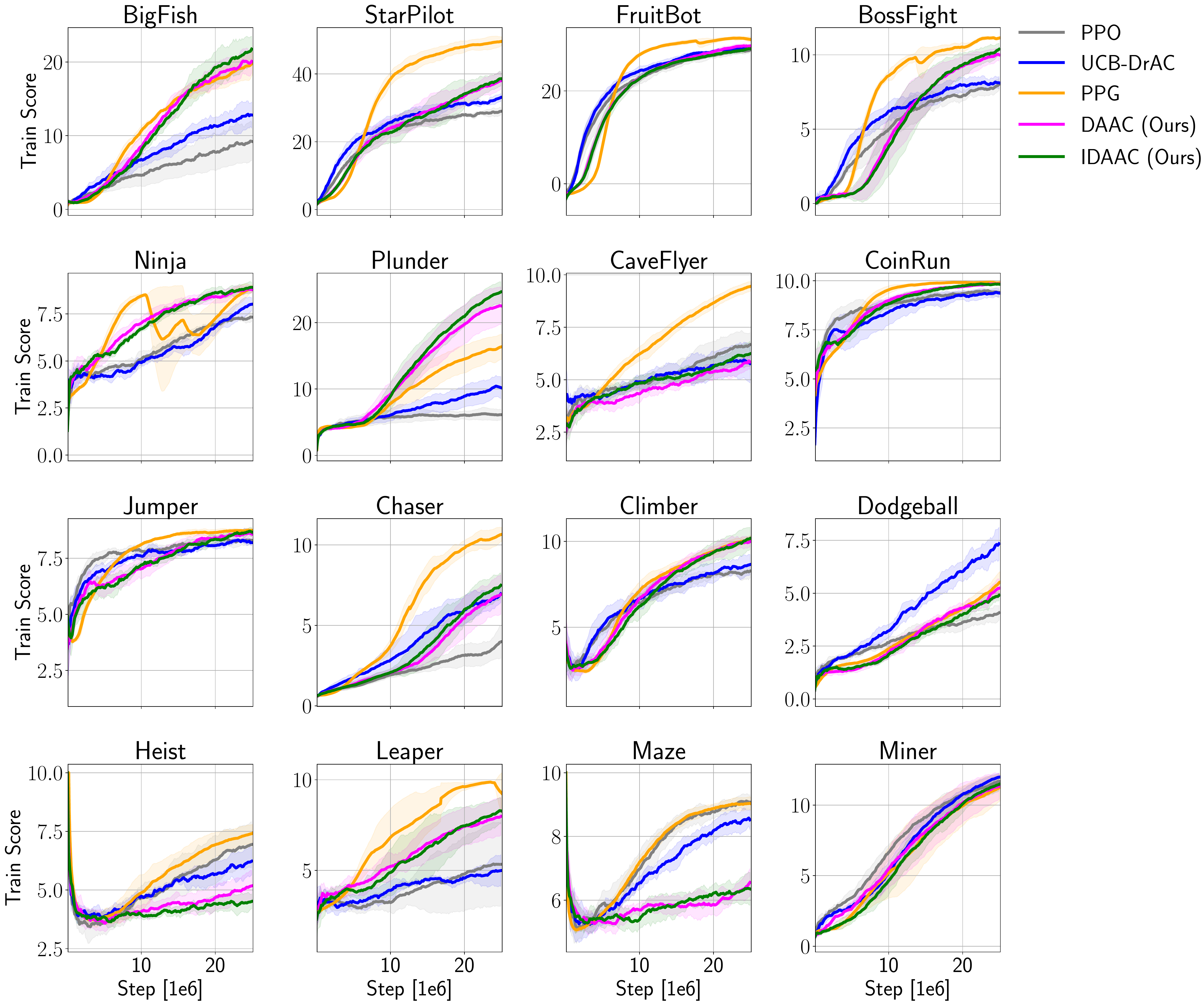}
    \caption{\textbf{Train Performance} of \ordergae{}, \gae{}, \ppg{}, \ucbdrac{}, and \ppo{} on all Procgen games. \ordergae{} outperforms the other methods on most games and is significantly better than \ppo{}. The mean and standard deviation are computed over 10 runs with different seeds.}
    \label{fig:results_train}
\end{figure*}

Figures~\ref{fig:results_test} and~\ref{fig:results_train} show the test and train performance for \ordergae{}, \gae{}, \ppg{}, \ucbdrac{}, and \ppo{} on all Procgen games. Our methods, \ordergae{} and \gae{} demonstrate superior test performance, outperforming all other baselines on the majority of the games, while being comparable on most of the remaining ones. 

% There are a few games such as Starpilot, Chaser, and CaveFlyer, where \ppg{} achieves better results than \ordergae{} and \gae{}. However, note that \ppg{} uses a batch size which is four times larger than the one used by our methods. When using the same batch size as our methods, \ppg{}'s performance on those games becomes comparable with that of \ordergae{}. 

\begin{figure*}[ht!]
    \centering
    \includegraphics[width=\textwidth]{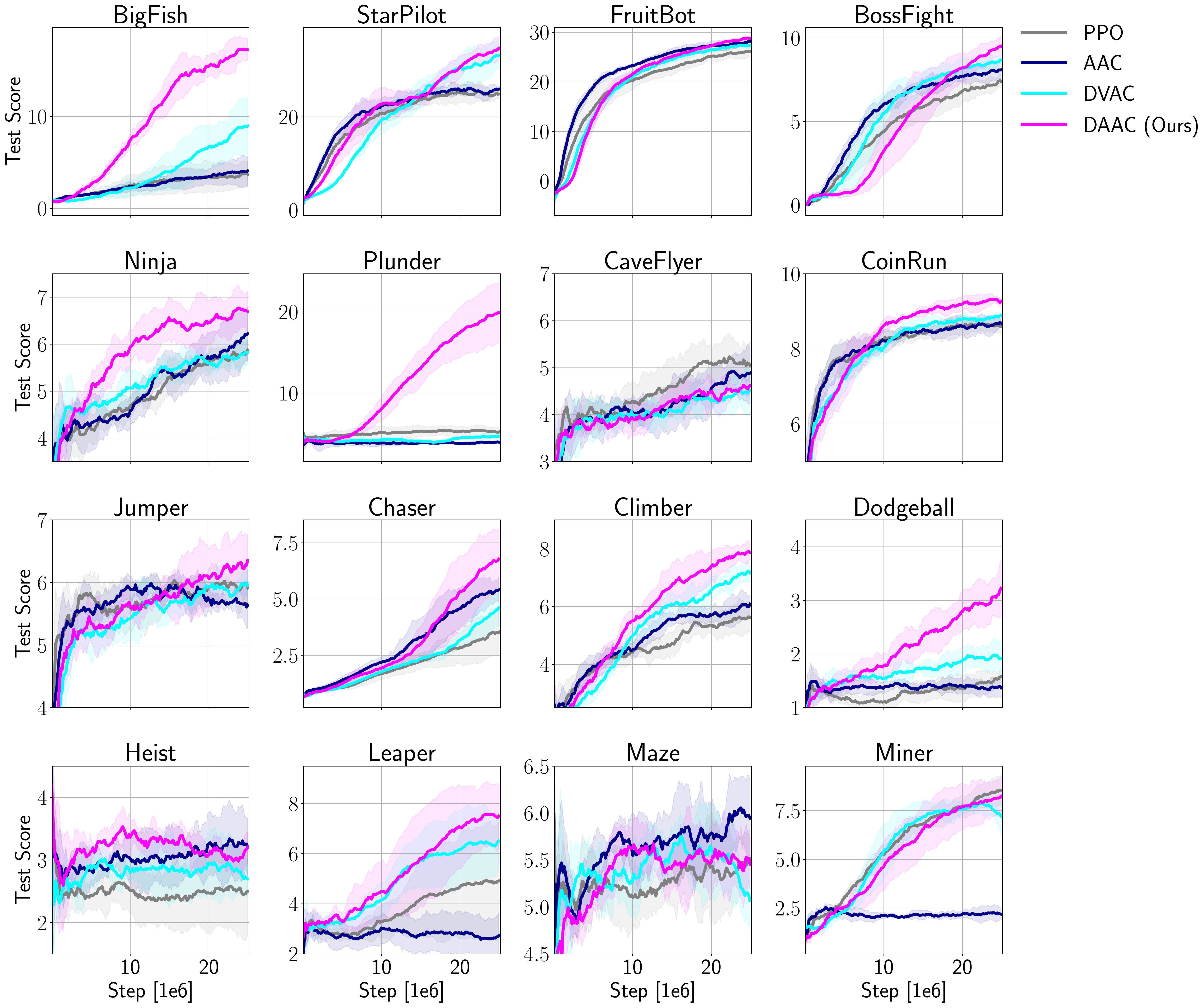}
    \caption{\textbf{Test Performance} of \gae{}, \valuegae{}, \gaeppo{}, and \ppo{} on all Procgen games. \valuegae{} is an ablation of \gae{} that replaces the advantage head of the policy network with a value head. \gaeppo{} is similar to PPO but has an additional advantage head as part of the policy network. \gae{} outperforms all these ablations, emphasizing the importance of all of its components. The mean and standard deviation are computed over 5 runs with different seeds.}
    \label{fig:ablations_test}
\end{figure*}

\begin{figure*}[ht!]
    \centering
    \includegraphics[width=\textwidth]{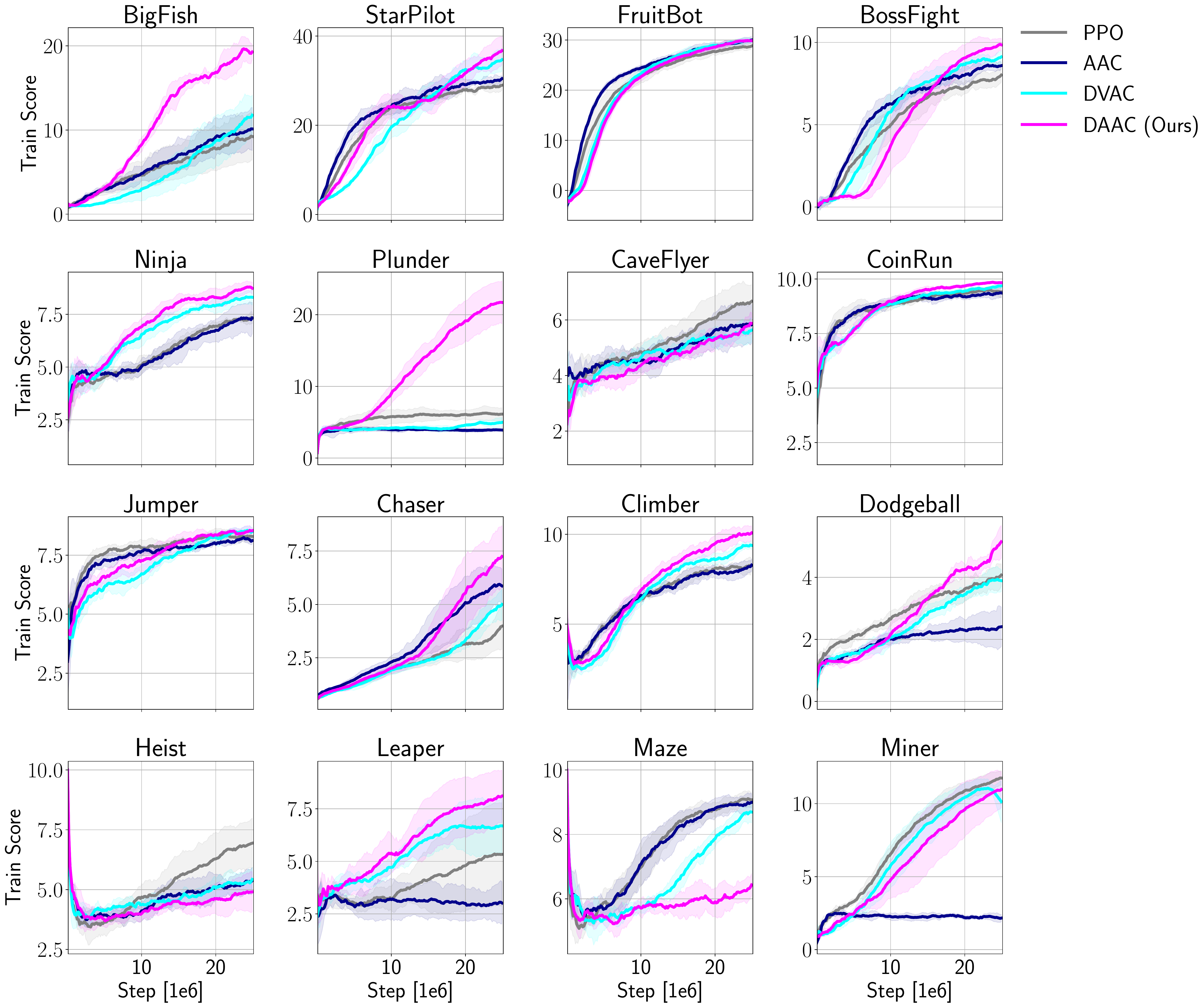}
    \caption{\textbf{Train Performance}  of \gae{}, \valuegae{}, \gaeppo{}, and \ppo{} on all Procgen games. \valuegae{} is an ablation of \gae{} that replaces the advantage head of the policy network with a value head. \gaeppo{} is similar to PPO but has an additional advantage head as part of the policy network. \gae{} outperforms all these ablations, emphasizing the importance of all of its components. The mean and standard deviation are computed over 5 runs with different seeds.}
    \label{fig:ablations_train}
\end{figure*}

Figures~\ref{fig:ablations_test} and~\ref{fig:ablations_train} show the test and train performance for \ppo, \gae{}, and two ablations, \valuegae{} and \gaeppo{}, on all Procgen games. On both train and test environments, our method is substantially better than all the ablations for most of the games and comparable on the remaining ones (\ie{} CaveFlyer and Maze). 

%%%%%%%%%%%%%%%%%%%%%%%%%%
\clearpage
% \section{Breakdown of Procgen Scores}
% \label{app:procgen_scores}

Tables~\ref{tab:test_all} and~\ref{tab:train_all} show the final mean and standard deviation of the test and train scores obtained by our methods and the baselines on each of the 16 Procgen games, after 25M training steps. For brevity, we only include the strongest baselines in these tables. See~\citet{Raileanu2020AutomaticDA} for the scores obtained by \randfm{} and \ibacsni{}. 

\begin{table*}[h!]
    \small
    \caption{Procgen scores on test levels after training on 25M environment steps. The mean and standard deviation are computed using 10 runs with different seeds.}
    \begin{tabular}{l|ccccc|cc}
    \toprule
    Game & PPO & \mixreg{} & \plr{}	& \ucbdrac{}  & \ppg{} & \gae{} (Ours) & \ordergae{} (Ours) \\
    \toprule
    BigFish & $3.7\pm1.3 $  & $7.1\pm1.6$ & $10.9\pm2.8$  &	$9.2\pm2.0$ &  $11.2\pm1.4$ &	$17.8 \pm1.4 $ &	$\pmb{18.5\pm1.2}$  \\
    StarPilot& $24.9\pm1.0 $  & $32.4\pm1.5$ & $27.9\pm4.4$ &	$30.0\pm1.3$ & $\pmb{47.2\pm1.6}$ &	$36.4\pm2.8 $ &	$37.0\pm2.3$  \\
    FruitBot& $26.2\pm1.2 $  & $27.3\pm0.8$ & $28.0\pm1.4$  &$27.6\pm0.4$ & $27.8\pm0.6$ & 	$\pmb{28.6\pm0.6}$ & 	$27.9\pm0.5$
  \\
    BossFight& $7.4\pm0.4$  & $8.2\pm0.7$ & $8.9\pm0.4$  &$7.8\pm0.6 $ & $\pmb{10.3\pm0.2}$ & 	$9.6\pm0.5 $ & 	$9.8\pm0.6$
  \\
    Ninja&      $5.9\pm0.2$  & $6.8\pm0.5$ & $\pmb{7.2\pm0.4}$ &	$6.6\pm0.4 $ &  $6.6\pm0.1$	 & $6.8\pm0.4$	 & $6.8\pm0.4$
  \\
    Plunder&    $5.2\pm0.6$  & $5.9\pm0.5$ &  $8.7\pm2.2$ &	$8.3\pm1.1$ &  $14.3\pm2.0$ & 	$20.7\pm3.3 $ & 	$\pmb{23.3\pm1.4}$
  \\
    CaveFlyer& $5.1\pm0.4$  & $6.1\pm0.6$ & $6.3\pm0.5$  &		$5.0\pm0.8$  & $\pmb{7.0\pm0.4}$	 & $4.6\pm0.2 $ & 	$5.0\pm0.6$
  \\
    CoinRun&	$8.6\pm0.2 $  & $8.6\pm0.3$ &  $8.8\pm0.5$  &$8.6\pm0.2$ & $8.9\pm0.1$ & 	$9.2\pm0.2 $ & 	$\pmb{9.4\pm0.1}$
  \\
    Jumper&	  $5.9\pm0.2 $  & $6.0\pm0.3$ &  $5.8\pm0.5$ &	$6.2\pm0.3$&  $5.9\pm0.1$	 & $\pmb{6.5\pm0.4}$ & 	$6.3\pm0.2$
 \\
    Chaser&	  $3.5\pm0.9 $  & $5.8\pm1.1$ & $6.9\pm1.2$  &	$6.3\pm0.6$&  $\pmb{9.8\pm0.5}$	 & $6.6\pm1.2 $ & 	$6.8\pm1.0$
  \\
        Climber&	 $5.6\pm0.5 $  & $6.9\pm0.7$ & $6.3\pm0.8$  & $6.3\pm0.6 $  & $2.8\pm0.4$ & 	$7.8\pm0.2 $	 & $\pmb{8.3\pm0.4}$
 \\
    Dodgeball& $1.6\pm0.1 $  & $1.7\pm0.4$ &  $1.8\pm0.5$  &	$\pmb{4.2\pm0.9}$& $2.3\pm0.3$ & 	$3.3\pm0.5 $ & 	$3.2\pm0.3$
 \\
    Heist&	    $2.5\pm0.6 $  & $2.6\pm0.4$ & $2.9\pm0.5$  &	$3.5\pm0.4$ &  $2.8\pm0.4$	 & $3.3\pm0.2 $ & 	$\pmb{3.5\pm0.2}$
  \\
    Leaper&	    $4.9\pm2.2 $  & $5.3\pm1.1$ &  $6.8\pm1.2$  &	$4.8\pm0.9$ & $\pmb{8.5\pm1.0}$	 & $7.3\pm1.1 $ & 	$7.7\pm1.0$
 \\
    Maze&	     $5.5\pm0.3$  & $5.2\pm0.5$ &  $5.5\pm0.8$ &	$\pmb{6.3\pm0.1}$ & $5.1\pm0.3$ & 	$5.5\pm0.2 $	 & $5.6\pm0.3$
 \\
    Miner&	 $8.4\pm0.7 $  &  $9.4\pm0.4$ &  $9.6\pm0.6$ &		$9.2\pm0.6$ &  $7.4\pm0.2$ & 	$8.6\pm0.9 $ & 	$\pmb{9.5\pm0.4}$
  \\
    \bottomrule
    \end{tabular}
    \label{tab:test_all}
\end{table*}

\begin{table*}[h!]
    \small
    \caption{Procgen scores on train levels after training on 25M environment steps. The mean and standard deviation are computed using 10 runs with different seeds.}
    \begin{tabular}{l|ccccc|cc}
    \toprule
    Game & PPO & \mixreg{} & \plr{}	& \ucbdrac{}  & \ppg{} & \gae{} (Ours) & \ordergae{} (Ours) \\
    \toprule
    BigFish & $9.2\pm2.7 $  & $15.0\pm1.3$ & $7.8\pm1.0$ &	$12.8\pm1.8$ & $19.9\pm1.7$ & 	$20.1\pm1.6 $ & 	$\pmb{21.8\pm1.8}$
 \\
    StarPilot& $29.0\pm1.1 $  & $28.7\pm1.1$ & $2.6\pm0.3$ &		$33.1\pm1.3$ & $\pmb{49.6\pm2.1}$ & 	$38.0\pm2.6 $ & 	$38.6\pm2.2$
  \\
    FruitBot& $28.8\pm0.6 $  & $29.9\pm0.5$ &  $15.9\pm1.3$ &	$29.3\pm0.5$ &  $\pmb{31.1\pm0.5}$ & 	$29.7\pm0.4 $ & 	$29.1\pm0.7$
  \\
    BossFight& $8.0\pm0.4 $  & $7.9\pm0.8$ & $8.7\pm0.7$ &	$8.1\pm0.4$ & $\pmb{11.1\pm0.1}$ & 	$10.0\pm0.4 $ & 	$10.4\pm0.4$
 \\
    Ninja&      $7.3\pm0.3 $  & $8.2\pm0.4$ & $5.4\pm0.5$ &	$8.0\pm0.4$ &  $8.9\pm0.2$ & 	$8.8\pm0.2 $ & 	$\pmb{8.9\pm0.3}$
  \\
    Plunder&    $6.1\pm0.8 $  & $6.2\pm0.3$ &  $4.1\pm1.3$ &	$10.2\pm1.76$  & $16.4\pm1.9$ & 	$22.5\pm2.8$  & 	$\pmb{24.6\pm1.6}$
  \\
    CaveFlyer& $6.7\pm0.6 $  & $6.2\pm0.7$ & $6.4\pm0.1$ &		$5.8\pm0.9$  & $\pmb{9.5\pm0.2}$ & 	$5.8\pm0.4$ & 	$6.2\pm0.6$
 \\
    CoinRun&	$9.4\pm0.3 $  & $9.5\pm0.2$ & $5.4\pm0.4$ &		$9.4\pm0.2$ & $\pmb{9.9\pm0.0}$ & 	$9.8\pm0.0 $ & 	$9.8\pm0.1$
 \\
    Jumper&	  $8.3\pm0.2 $  & $8.5\pm0.4$ &  $3.6\pm0.5$ &		$8.2\pm0.1$  & $8.7\pm0.1$ & $8.6\pm0.3$ & $\pmb{8.7\pm0.2}$
  \\
    Chaser&	  $4.1\pm0.3 $  & $3.4\pm0.9$ & $6.3\pm0.7$ &		$7.0\pm0.6$  & $\pmb{10.7\pm0.4}$	 & $6.9\pm1.2 $ & 	$7.5\pm0.8$
  \\
    Climber&	 $6.9\pm1.0 $  & $7.5\pm0.8$ &  $6.2\pm0.8$ &		$8.6\pm0.6$ & $10.2\pm0.2$ & $10.0\pm0.3$ & 	$\pmb{10.2\pm0.7}$
 \\
    Dodgeball& $5.3\pm2.3 $  & $9.1\pm0.5$ &  $2.0\pm1.1$ &		$\pmb{7.3\pm0.8}$ &  $5.5\pm0.5$ & 	$5.2\pm0.4 $ & 	$4.9\pm0.3$
  \\
    Heist&	    $7.1\pm0.5$  & $4.4\pm0.3$ & $1.2\pm0.4$ &		$6.2\pm0.6$ & $\pmb{7.4\pm0.4}$ & 	$5.2\pm0.7 $	 & $4.5\pm0.3$
  \\
    Leaper&	    $5.5\pm0.4$  & $3.2\pm1.2$ & $6.4\pm0.4$  &		$5.0\pm0.9$ & $\pmb{9.3\pm1.1}$ & 	$8.0\pm1.1 $ & 	$8.3\pm0.7$
  \\
    Maze&	     $\pmb{9.1\pm0.2 }$  & $8.7\pm0.7$ & $4.1\pm0.5$ &		$8.5\pm0.3$ &  $9.0\pm0.2$ & 	$6.6\pm0.4 $ & 	$6.4\pm0.5$
  \\
    Miner&	 $11.7\pm0.5 $  & $8.9\pm0.9$ & $9.7\pm0.4$ &	$\pmb{12.0\pm0.3}$ &  $11.3\pm1.0$ & 	$11.3\pm0.9$ & 	$11.5\pm0.5$
 \\
    \bottomrule
    \end{tabular}
    \label{tab:train_all}
\end{table*}

%%%%%%%%%%%%%%%%%%%%%%%%%%
\clearpage
\section{DeepMind Control Suite Experiments}
\label{app:dmc} 

\begin{figure*}[ht!]
    \centering
    \includegraphics[width=0.32\textwidth]{fig/dmc/cartpole-balance-natural-cartpole-balance-camera.pdf}
    \includegraphics[width=0.32\textwidth]{fig/dmc/cartpole-swingup-natural-cartpole-swingup-camera.pdf}
    \includegraphics[width=0.32\textwidth]{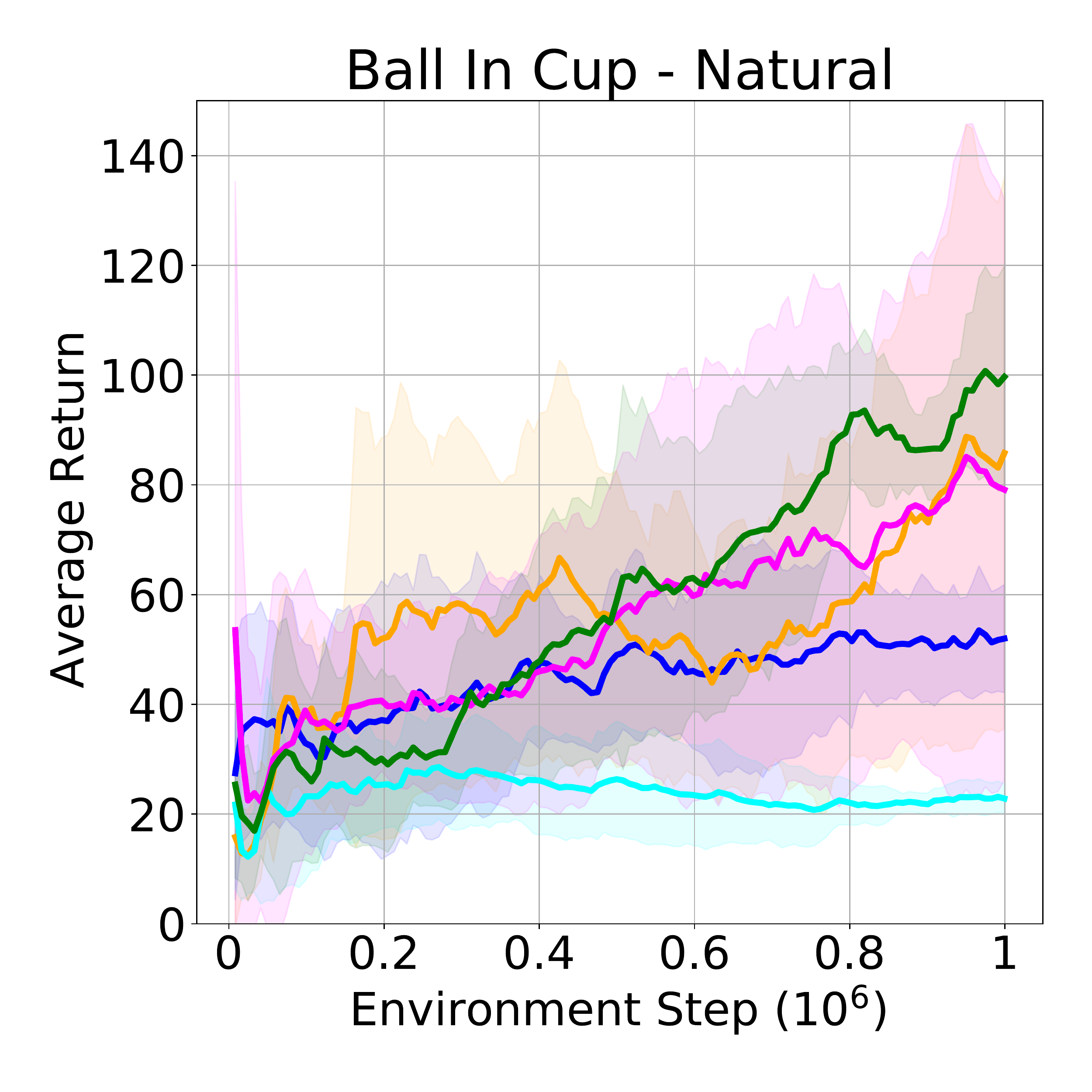}
    \includegraphics[width=0.32\textwidth]{fig/dmc/cartpole-balance-synthetic-cartpole-balance-camera.pdf}
    \includegraphics[width=0.32\textwidth]{fig/dmc/cartpole-swingup-synthetic-cartpole-swingup-camera.pdf}
    \includegraphics[width=0.32\textwidth]{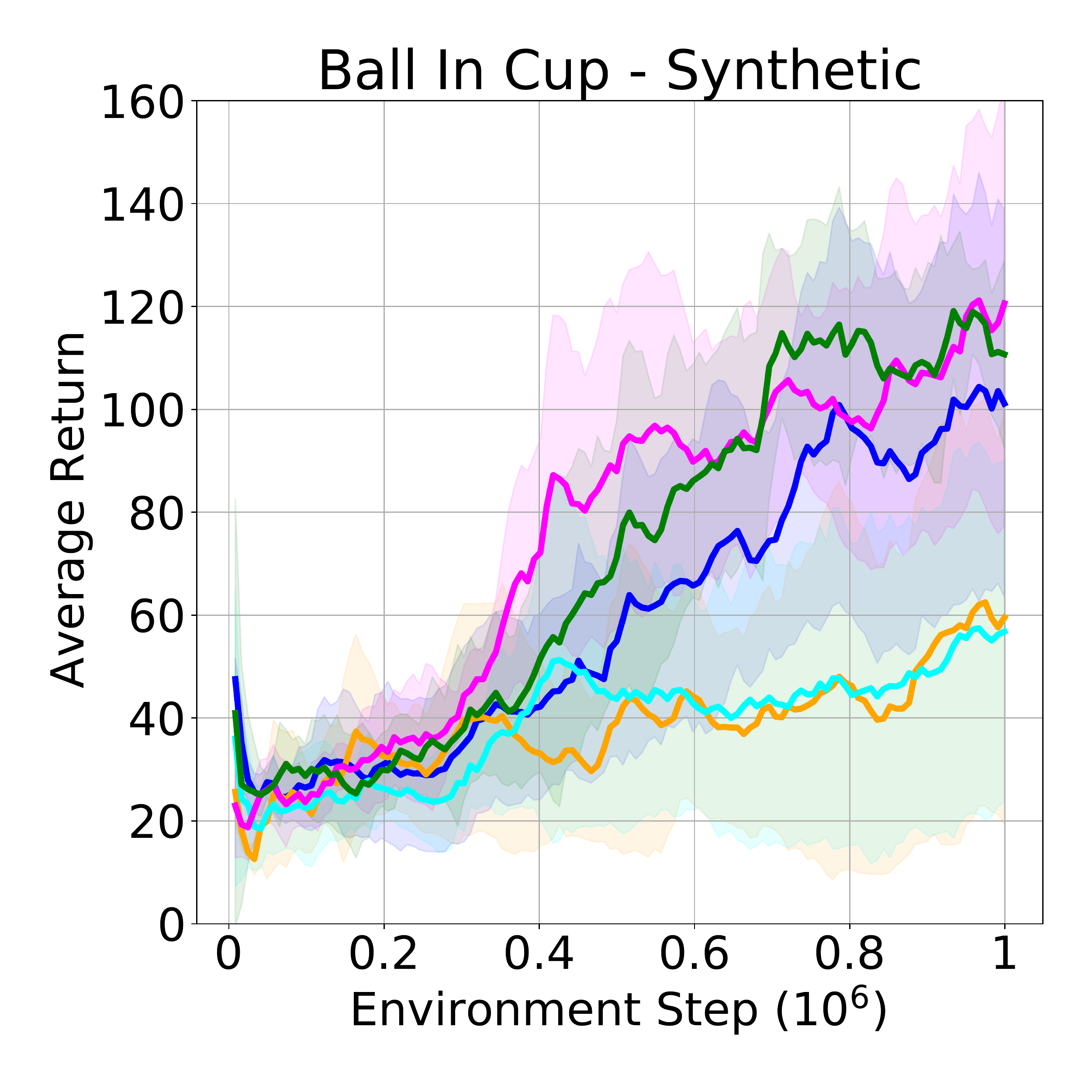}
    \caption{\textbf{Average return on three DMC tasks, Cartpole Balance (left), Cartpole Swingup (center), and Ball In Cup (right), with natural (top) and synthetic (bottom) video backgrounds.} The mean and standard deviation are computed over 10 runs with different seeds. \gae{} and \ordergae{} outperform \ppo{}, \ppg{}, and \ucbdrac{}.}
    \label{fig:dmc_results_all}
\end{figure*}

For our DMC experiments, we followed the protocol proposed in~\citet{Zhang2020LearningIR} to modify the tasks so that they contain natural and synthetic distractors in the background. For each DMC task, we split the generated environments (each with a different background video) into training ($80\%$) and testing ($20\%$). The results shown here correspond to the average return on the test environments, over the course of training. Note that this setting is slightly different from the one used in~\citet{Raileanu2020AutomaticDA} which shows results on all the generated environments, just like~\citet{Zhang2020LearningIR}. As Figure~\ref{fig:dmc_results_all} shows, our methods outperform the baselines on these continuous control tasks.

In line with standard practice for this benchmark, we use 8 action repeats for Cartpole Swingup and 4 for Cartpole Balance and Ball In Cup. We also use 3 stacked frames as observations. To find the best hyperparameters, we ran a grid search over the learning rate in $[0.0001, 0.0003, 0.0007, 0.001]$, the number of minibatches in $[32, 8, 16, 64]$, the entropy coefficient in $[0.0, 0.01, 0.001, 0.0001]$, and the number of PPO epochs per update in $[3, 5, 10, 20]$. We found 10 ppo epochs, 0.0 entropy coefficient, 0.0003 learning rate, and 32 minibatches to work best across these environments. We use $\gamma = 0.99$, $\lambda = 0.95$ for the generalized advantage estimates, 2048 steps, 1 process, value loss coefficient 0.5, and linear rate decay over 1 million environment steps. Following this grid search, we used the best values found for all the methods. Any other hyperaparameters not mentioned here were set to the same values as the ones used for Procgen as described above. For \ucbdrac{}, we used the best hyperparameters found by the corresponding authors. For \ppg{}, we ran the same hyperparameter search as the one performed in the original paper for Procgen and found $N_\pi = 32$, $E_\pi = 1$, $E_V = 1$, $E_{aux} = 6$, and $\beta_{clone} = 1$ to be the best. Similarly, for \gae{} and \ordergae{}, we ran the same hyperparameter search as for Procgen and found that $E_V = 9$, $N_\pi = 32$, $\alpha_a = 0.1$, and $\alpha_i = 0.1$ worked best across all environments. 

% For Cartpole Balance, we used 0.0001 learning rate and 16 minibatches while for the others we used 0.0003 learning rate and 32 minibatches, respectively.
%We used $\alpha_a = 0.1$ for Cartpole Balance, $\alpha_a = 0.75$ for Cartpole Swingup, and $\alpha_a = 0.3$ for Ball In Cup. 

%%%%%%%%%%%%%%%%%%%%%%%%%%
\clearpage
\section{Value Loss and Generalization}
\label{app:corr}

\begin{figure*}[ht!]
    \centering
    \includegraphics[width=0.95\textwidth]{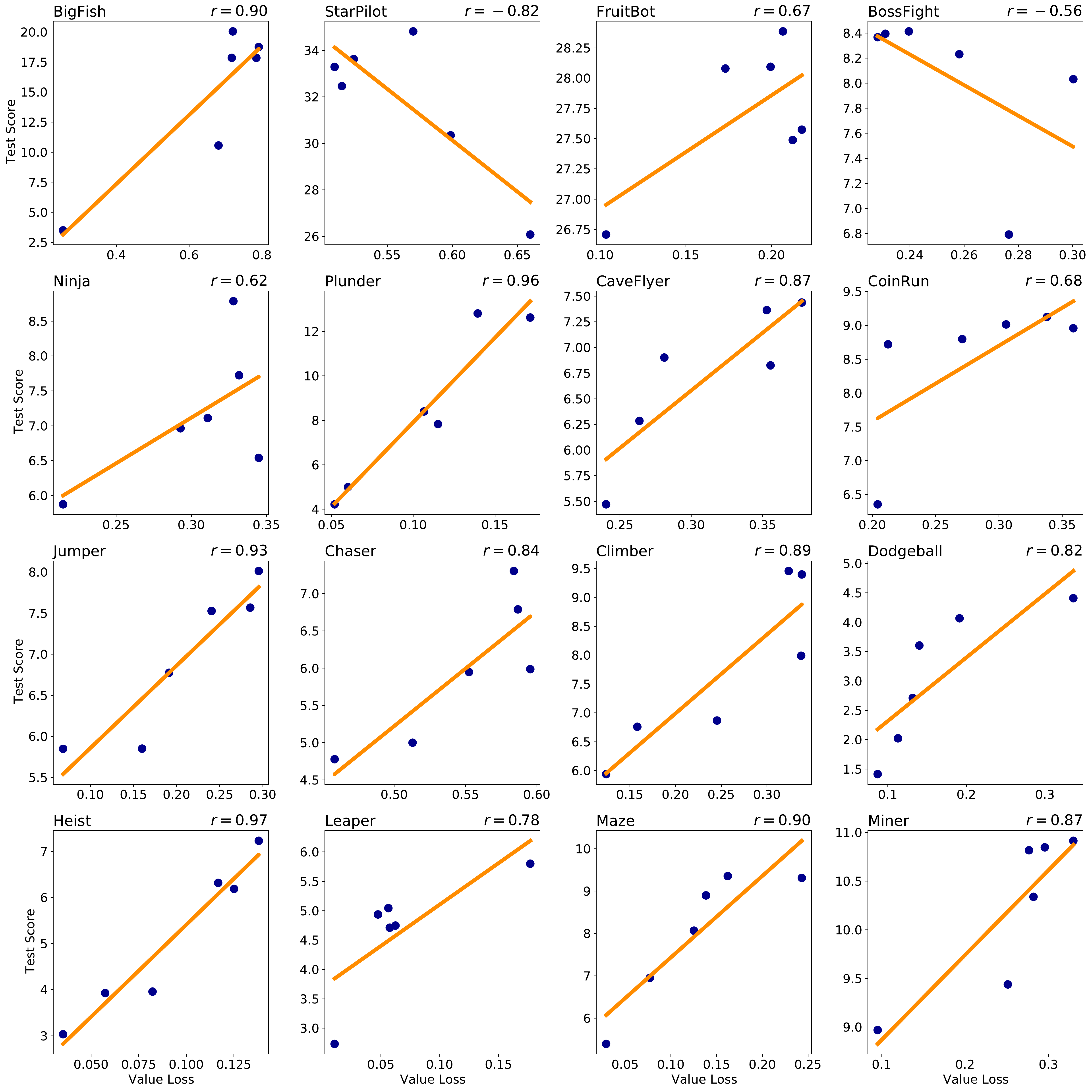}
    \caption{\textbf{Correlation of the value loss and test score for PPO agents trained on varying numbers of levels: 200, 500, 1000, 2000, 5000, and 10000.} Surprisingly, the value loss is positively correlated with the test score for most games, suggesting that \textit{models with larger value loss generalize better}.}
    \label{fig:corr_value_test_all}
\end{figure*}

\begin{figure*}[ht!]
    \centering
    \includegraphics[width=\textwidth]{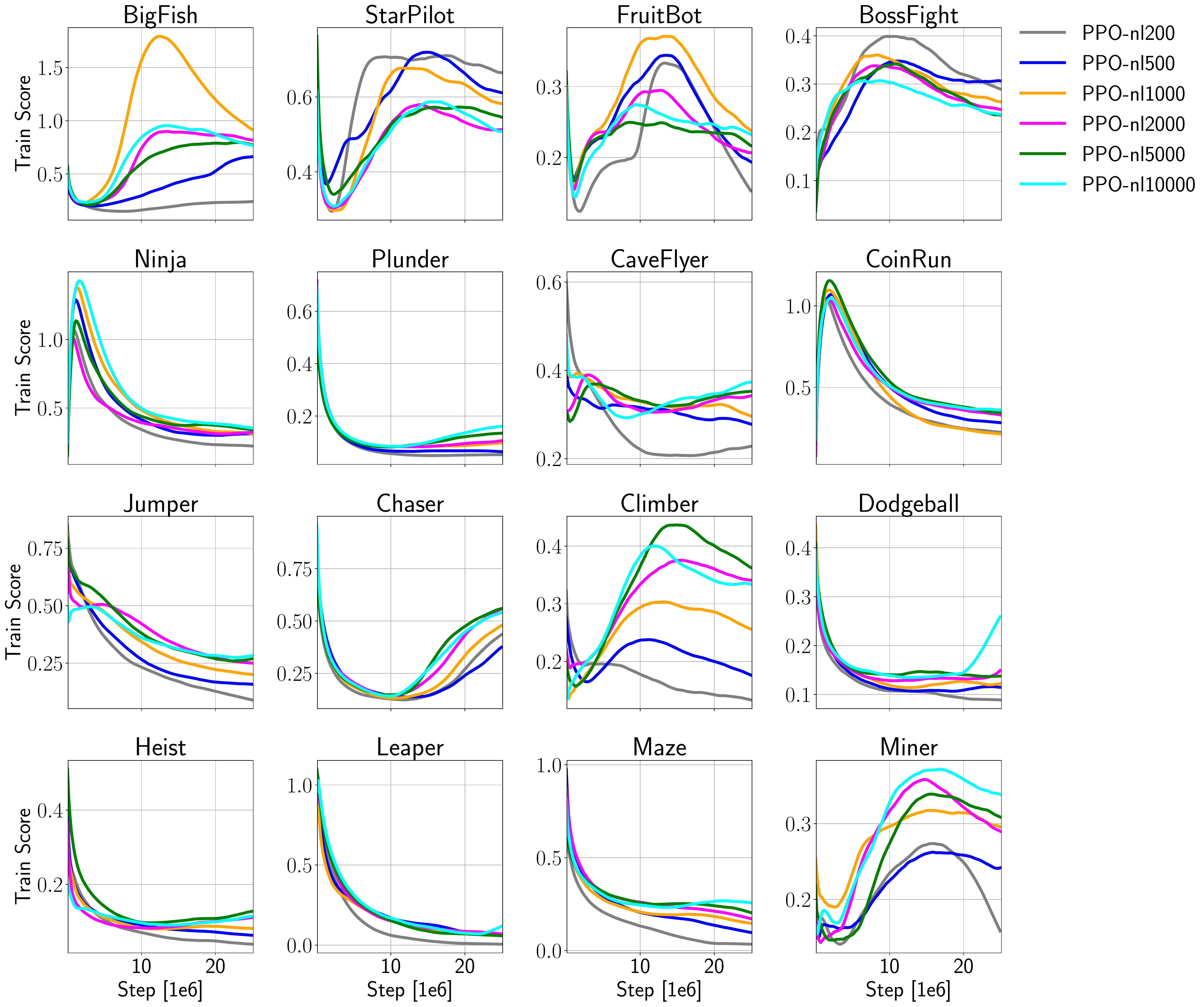}
    \caption{\textbf{Test score for PPO agents trained on varying numbers of levels: 200, 500, 1000, 2000, 5000, and 10000.} For most games, the test score increases with the number of training levels, suggesting that models trained on more levels generalize better to unseen levels, as expected.}
    \label{fig:test_score_all}
\end{figure*}

\begin{figure*}[ht!]
    \centering
    \includegraphics[width=\textwidth]{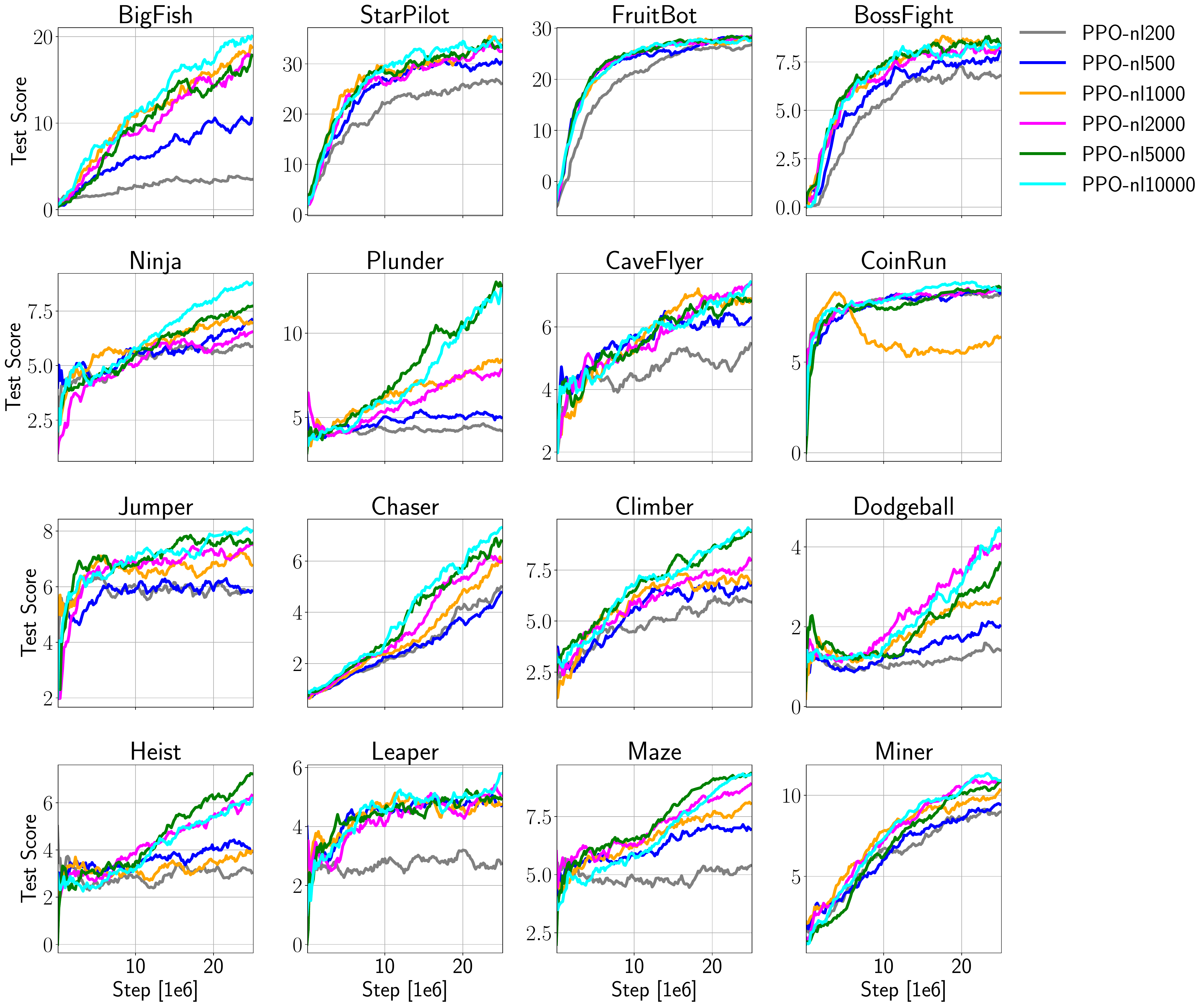}
    \caption{\textbf{Value loss for PPO agents trained on varying numbers of levels: 200, 500, 1000, 2000, 5000, and 10000.} For most games, the value loss increases with the number of training levels used, suggesting that models trained on more levels (and thus with better generalization) have higher value loss.}
    \label{fig:value_loss_all}
\end{figure*}

In this section, we look at the relationship between the value loss, test score, and number of training levels for all Procgen games (see Figures~\ref{fig:corr_value_test_all},~\ref{fig:value_loss_all}, and~\ref{fig:test_score_all}). As discussed in the paper, the value loss is \textit{positively} correlated with the test score and number of training levels. This result goes against our intuition from training RL algorithms on single environments where in general, the value loss is \textit{inversely} correlated with the agent's performance and sample efficiency. This observation further supports our claim that having a more accurate value function can lead to representations that overfit to the training environments. When using a shared network for the policy and value, this results in policies that do not generalize well to new environments. By decoupling the representations of the policy and value function, our methods \gae{} and \ordergae{} can achieve the best of both worlds by \begin{enumerate*}[label=(\roman*)]
\item learning accurate value functions, while also
\item learning representations and policies that better generalize to unseen environments. 
\end{enumerate*}

%%%%%%%%%%%%%%%%%%%%%%%%%%
\clearpage
% \section{Advantage vs. Value Tracking}
% \subsection{Advantage vs. Value for Generalization}
\section{Advantage vs. Value During an Episode}
\label{app:adv_val}

\begin{figure*}[ht!]
    \centering
    \includegraphics[width=1.0\textwidth]{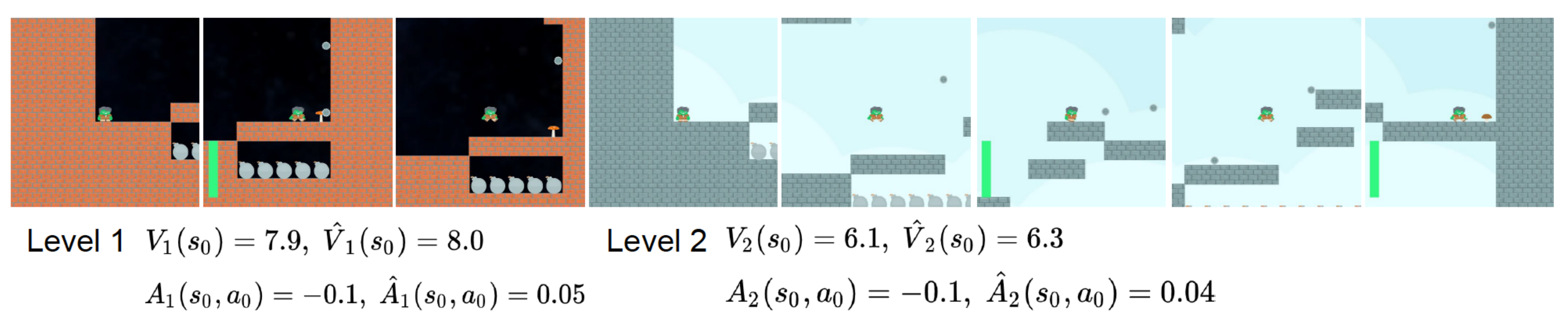}
    \caption{\textbf{Policy-Value Asymmetry.} Two Ninja levels with initial observations that are \textit{semantically identical but visually different}. Level 1 (first three frames from the left with black background) is much shorter than Level 2 (last six frames with blue background). Both the true and the estimated values (by a PPO agent trained on 200 levels) of the initial observation are higher for Level 1 than for Level 2 \ie{} $V_1(s_0) > V_2(s_0)$ and $\hat{V}_1(s_0) > \hat{V}_2(s_0)$. Thus, to accurately predict the value function, the representations must capture level-specific features (such as the backgrounds), which are irrelevant for finding the optimal policy. Consequently, using a common representation for both the policy and value function can lead to overfitting to spurious correlations and poor generalization to unseen levels. In contrast to the value, the true advantage of the initial states and noop action has the same values for the two levels, and the advantages predicted by the agent also have very similar values.}
    \label{fig:ninja_levels_value_adv}
\end{figure*}

Figure~\ref{fig:ninja_levels_value_adv} shows two different Ninja levels and their corresponding true and predicted values and advantages (by a PPO agent trained on 200 levels). The true values and advantages are computed assuming an optimal policy. For illustration purposes, we show the advantage for the noop action, but similar conclusions apply to the other actions. As the figure shows, the true value function is different for the two levels, while the advantage is (approximately) the same. This holds true for the agent's estimates as well, suggesting that using the value loss to update the policy parameters can lead to more overfitting than using the advantage loss (since it exhibits less dependence on the idiosyncrasies of a level such as its length or difficulty).  

\begin{figure*}[ht!]
    \centering
    \includegraphics[width=0.19\textwidth]{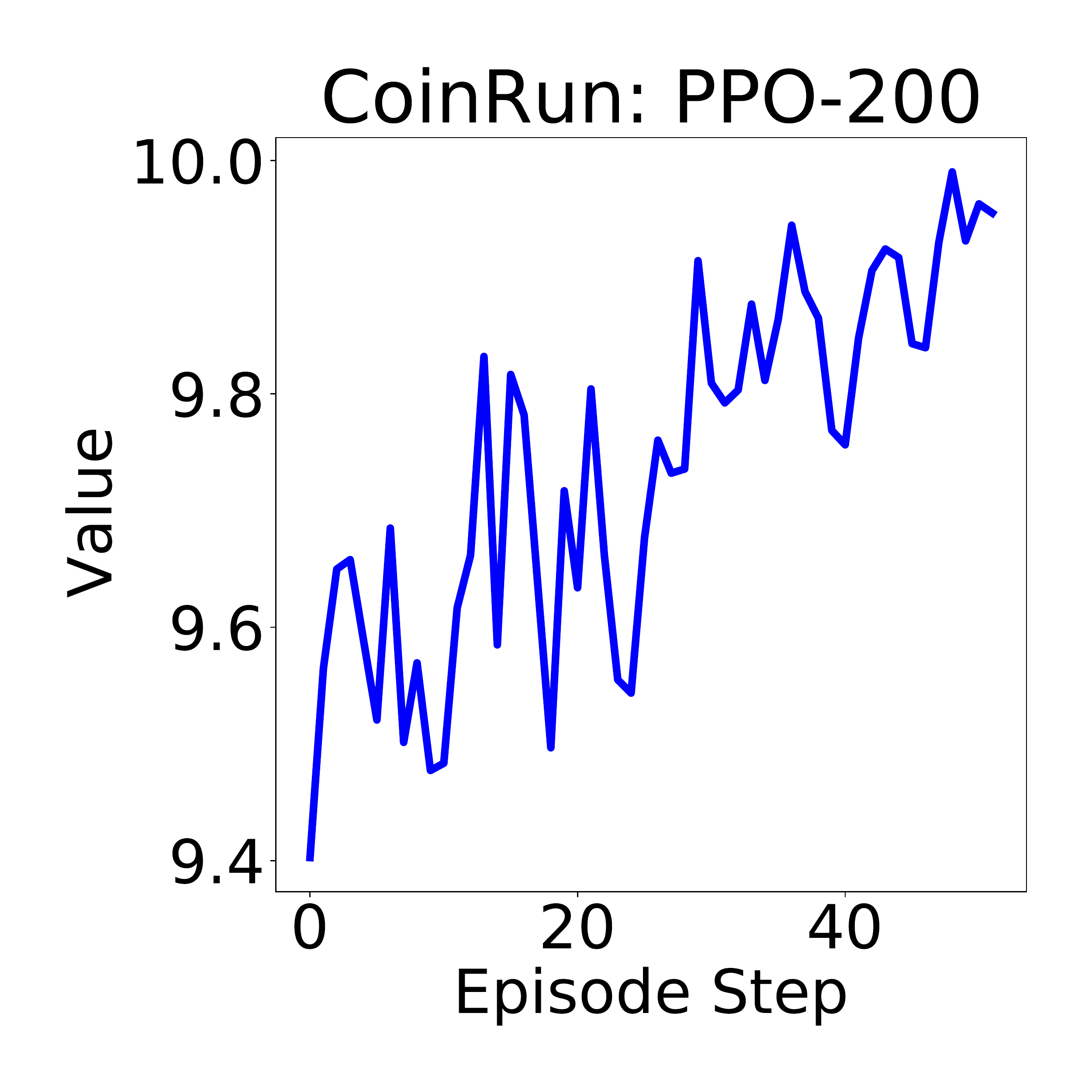}
    \includegraphics[width=0.19\textwidth]{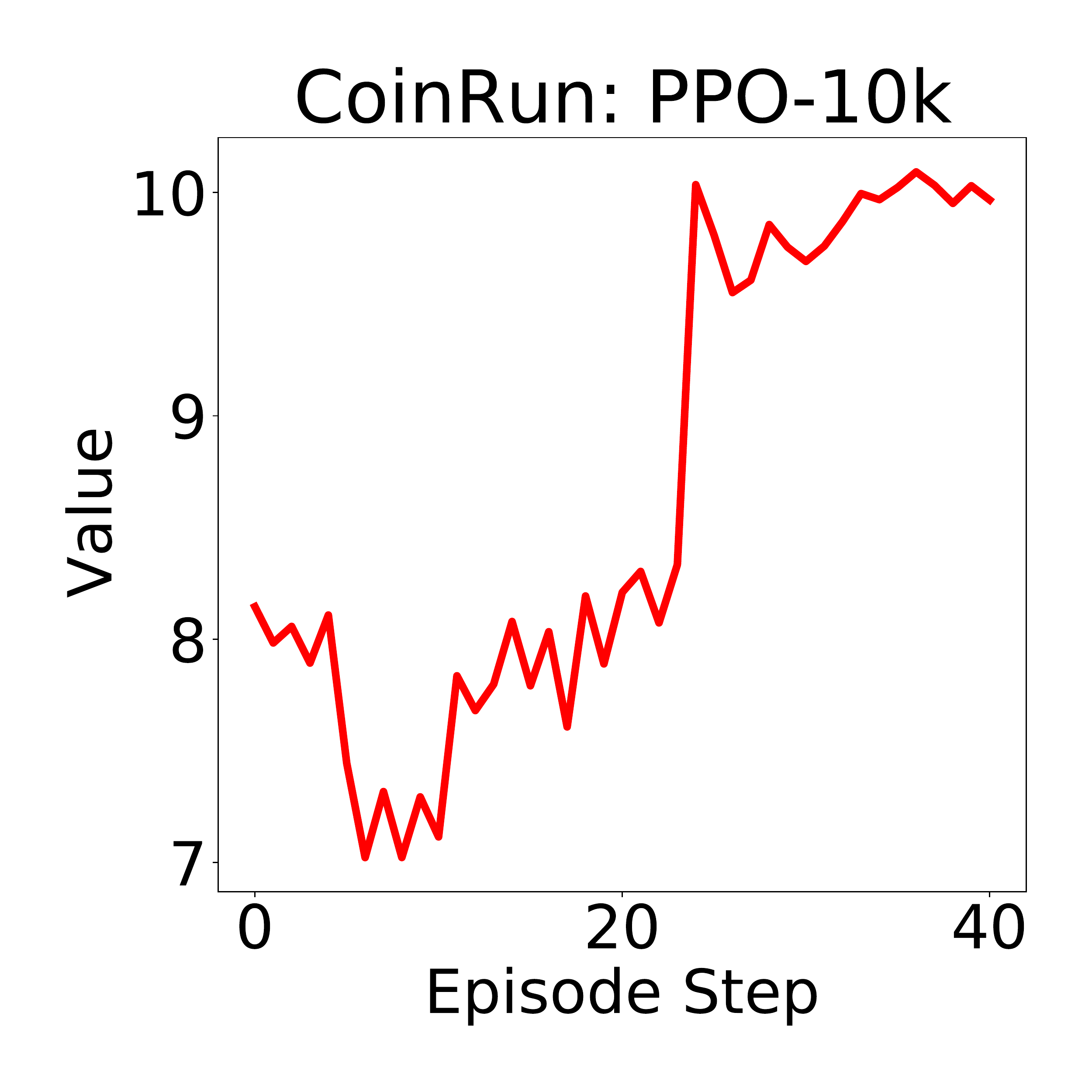}
    \includegraphics[width=0.19\textwidth]{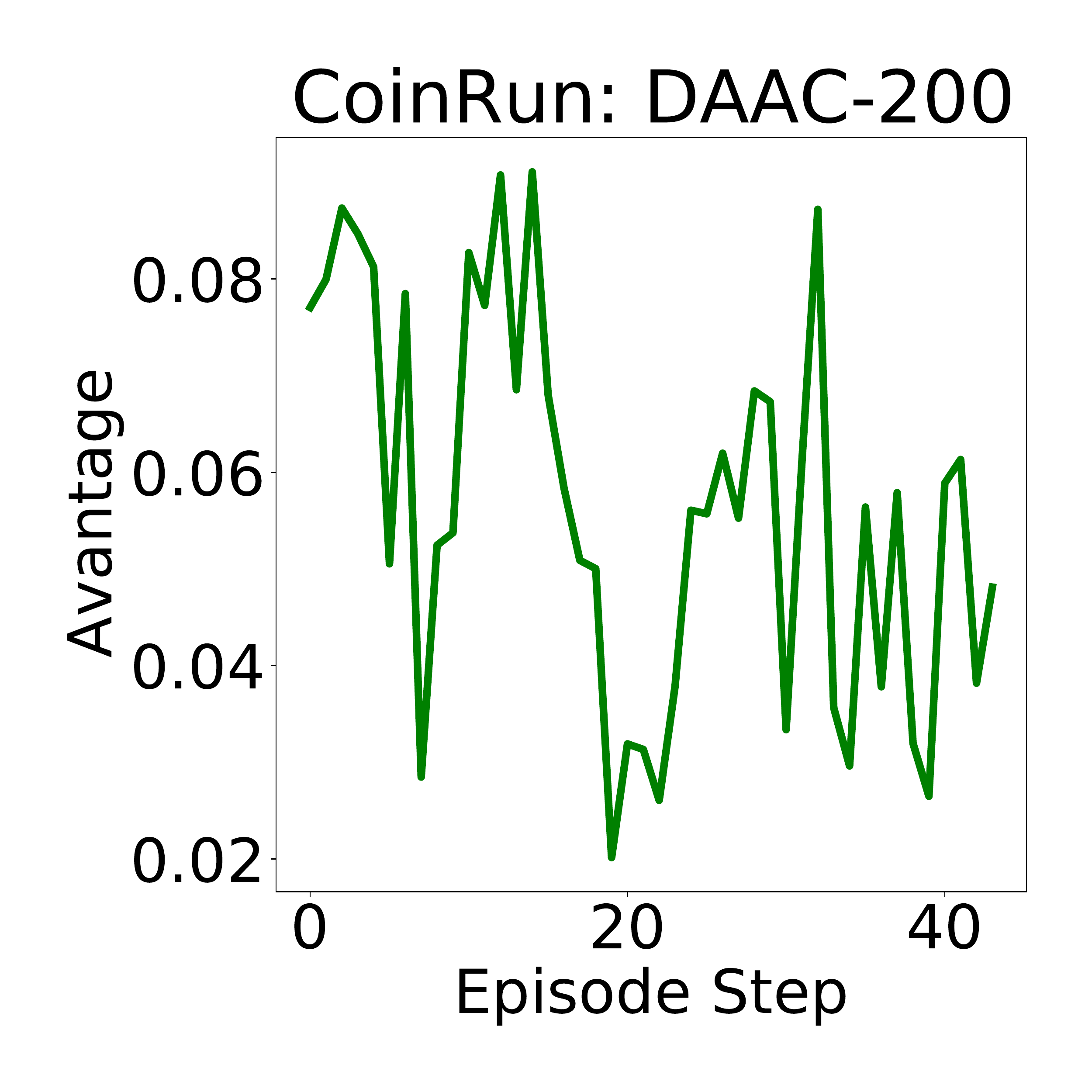}
    \includegraphics[width=0.19\textwidth]{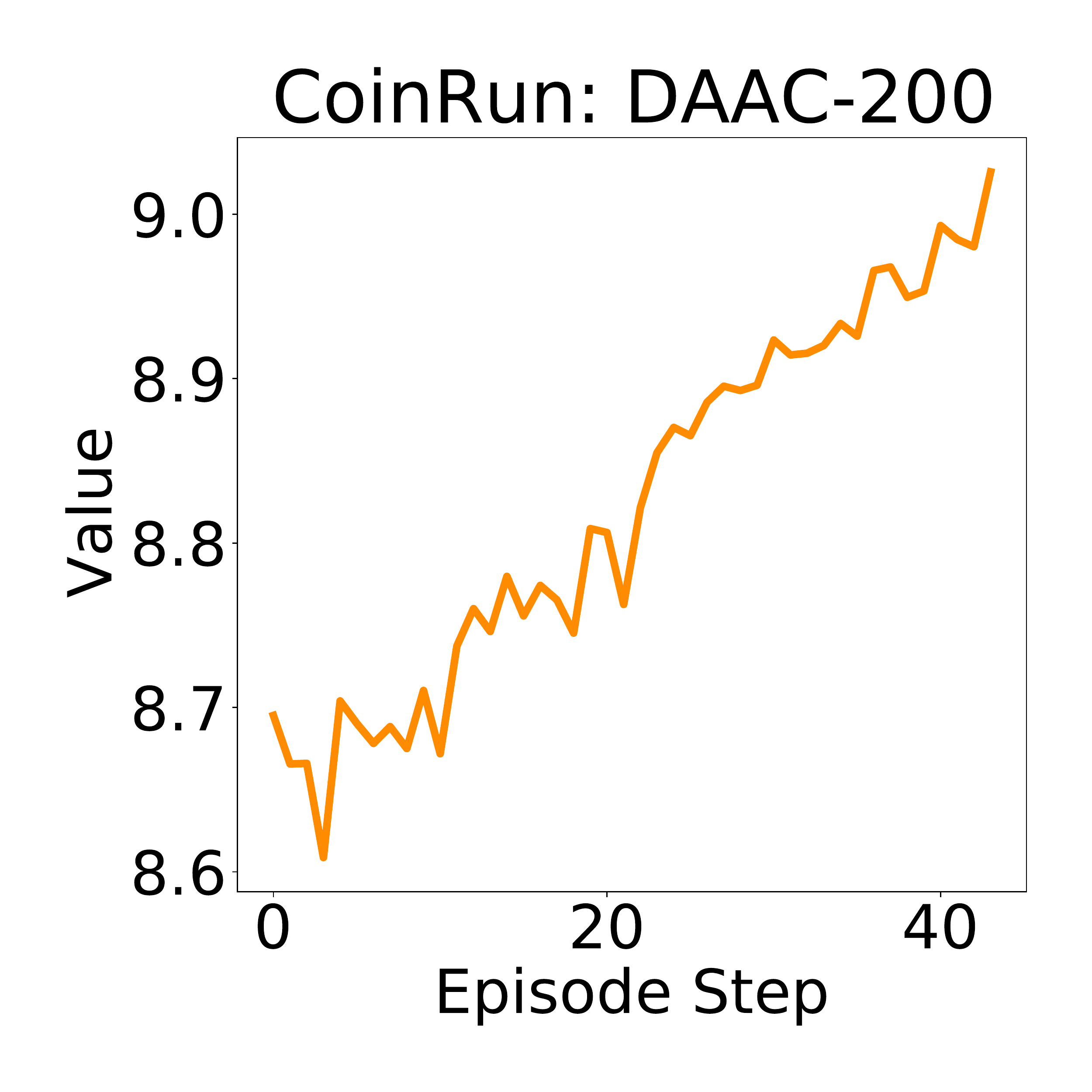}
    \includegraphics[width=0.19\textwidth]{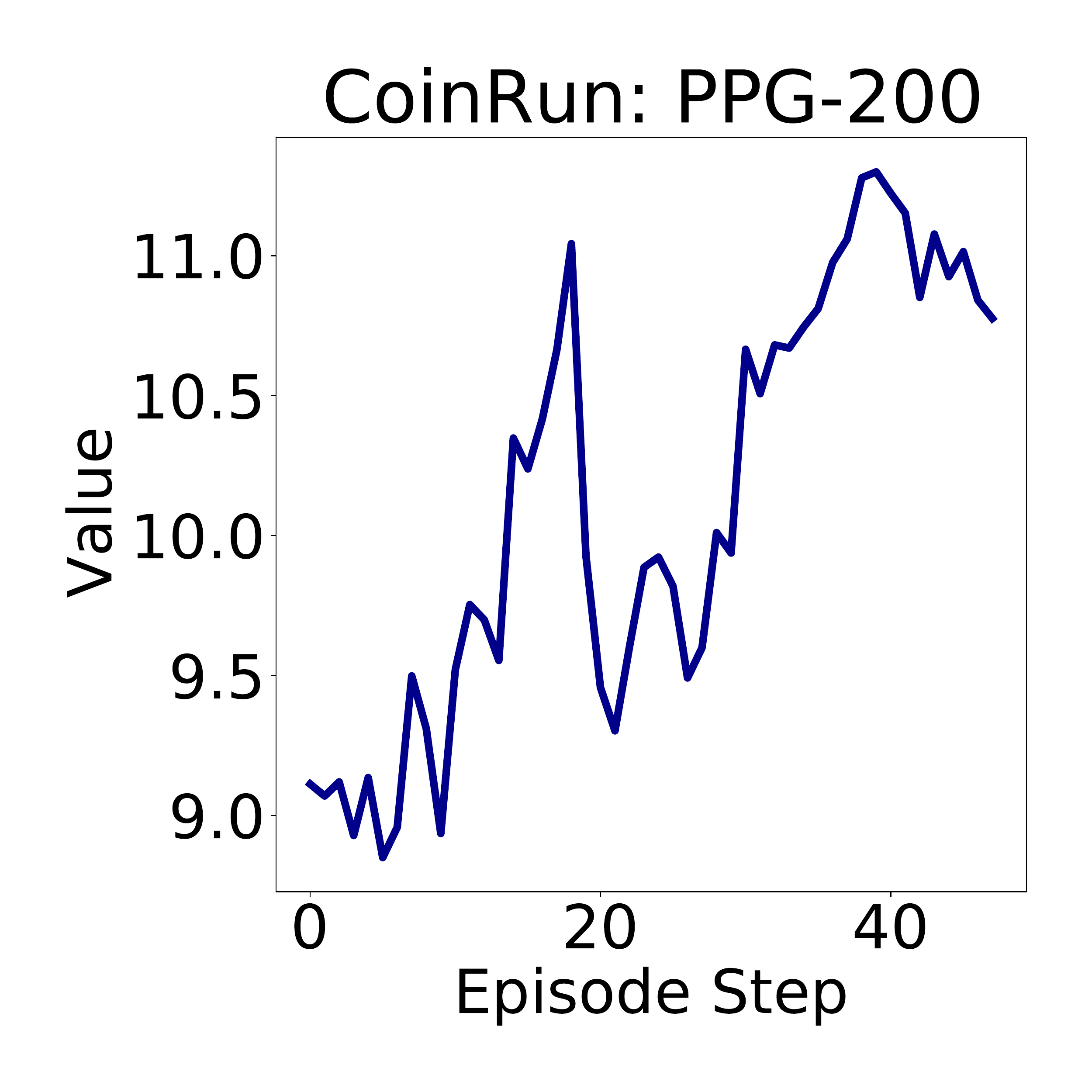}
    % \vspace{-4mm}
    \caption{\textbf{Examples illustrating the timestep-dependence of the value function for a single CoinRun level.} The dark and light blue curves show the value as a function of episode step for \ppg{} and \ppo{}, respectively, each trained on 200 levels. Note the near linear relationship, indicating overfitting to the training levels. By contrast, a \ppo{} model (red) trained on 10k levels (thus exhibiting far less overfitting) does not show this relationship. Our \gae{} model trained on 200 levels (green) also lacks this adverse dependence in the advantage prediction which is used for training the policy network, thus is able to generalize better than the PPO model trained on the same amount of data (see Fig.~\ref{fig:procgen_results}). Nevertheless, \gae{}'s value estimate still have a linear trend (orange) but, in contrast to \ppo{} and \ppg{}, this does not negatively affect the policy since we use separate networks for learning the policy and value.}
    \label{fig:step_value_gae_coinrun}
\end{figure*}

\begin{figure*}[ht!]
    \centering
    \includegraphics[width=0.19\textwidth]{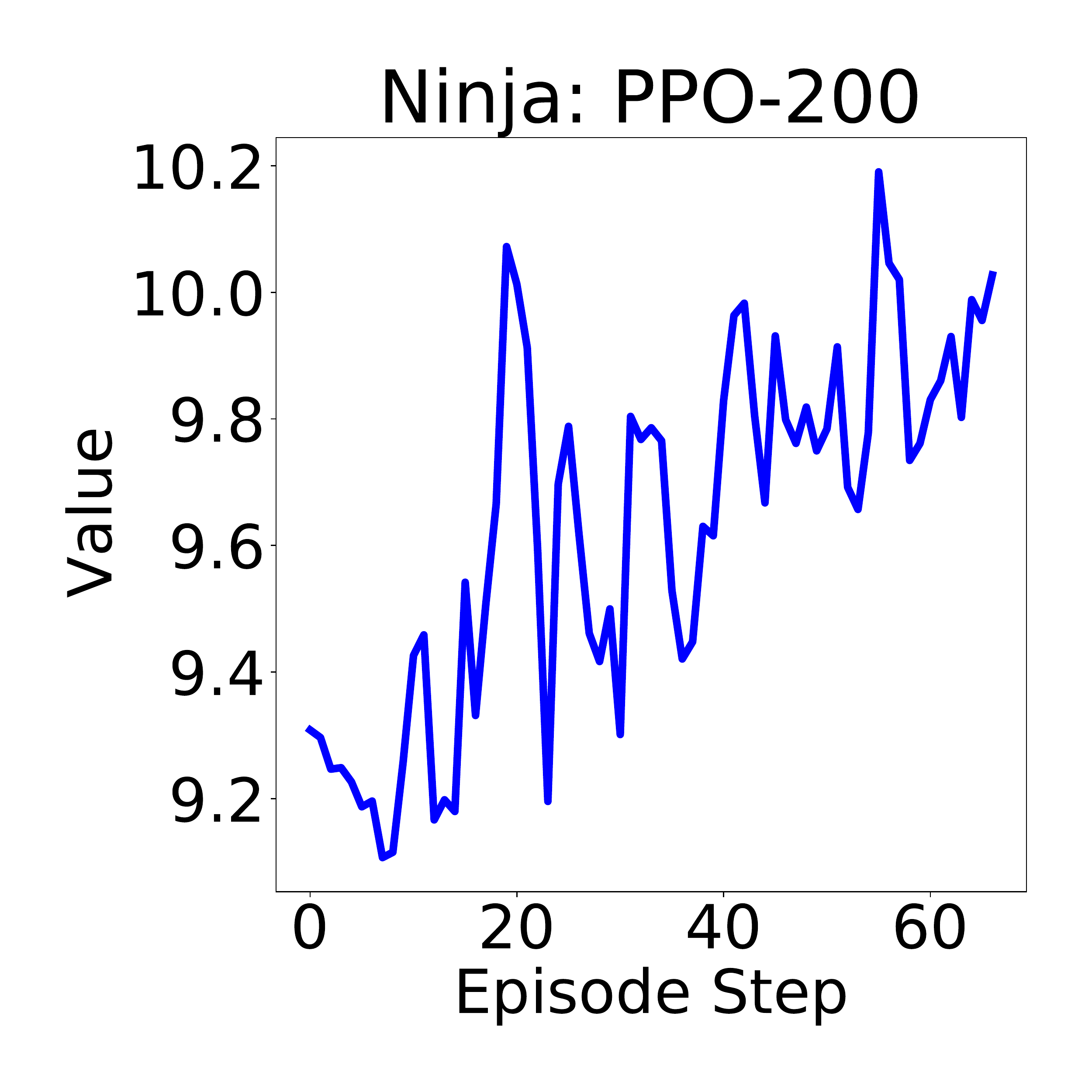}
    \includegraphics[width=0.19\textwidth]{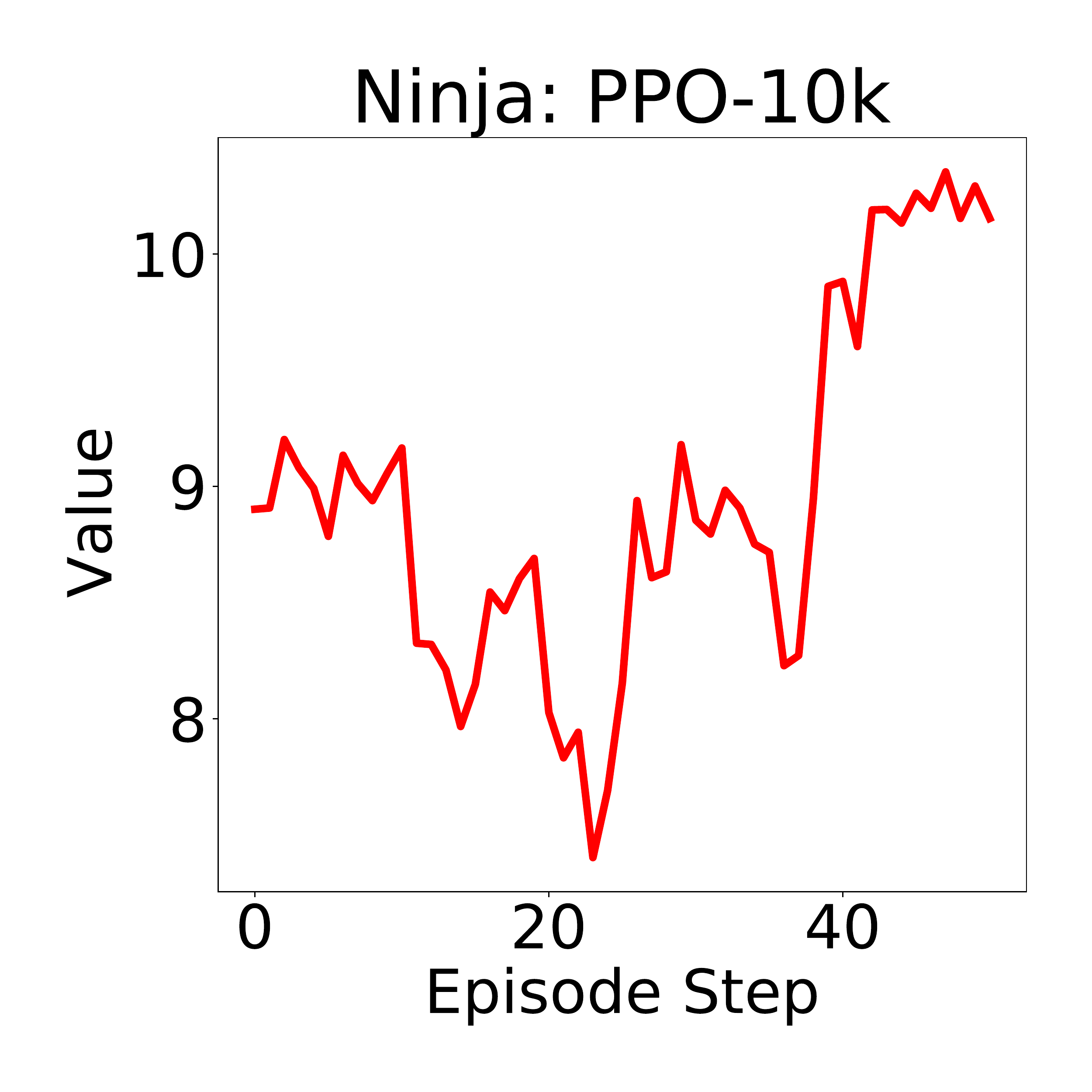}
    \includegraphics[width=0.19\textwidth]{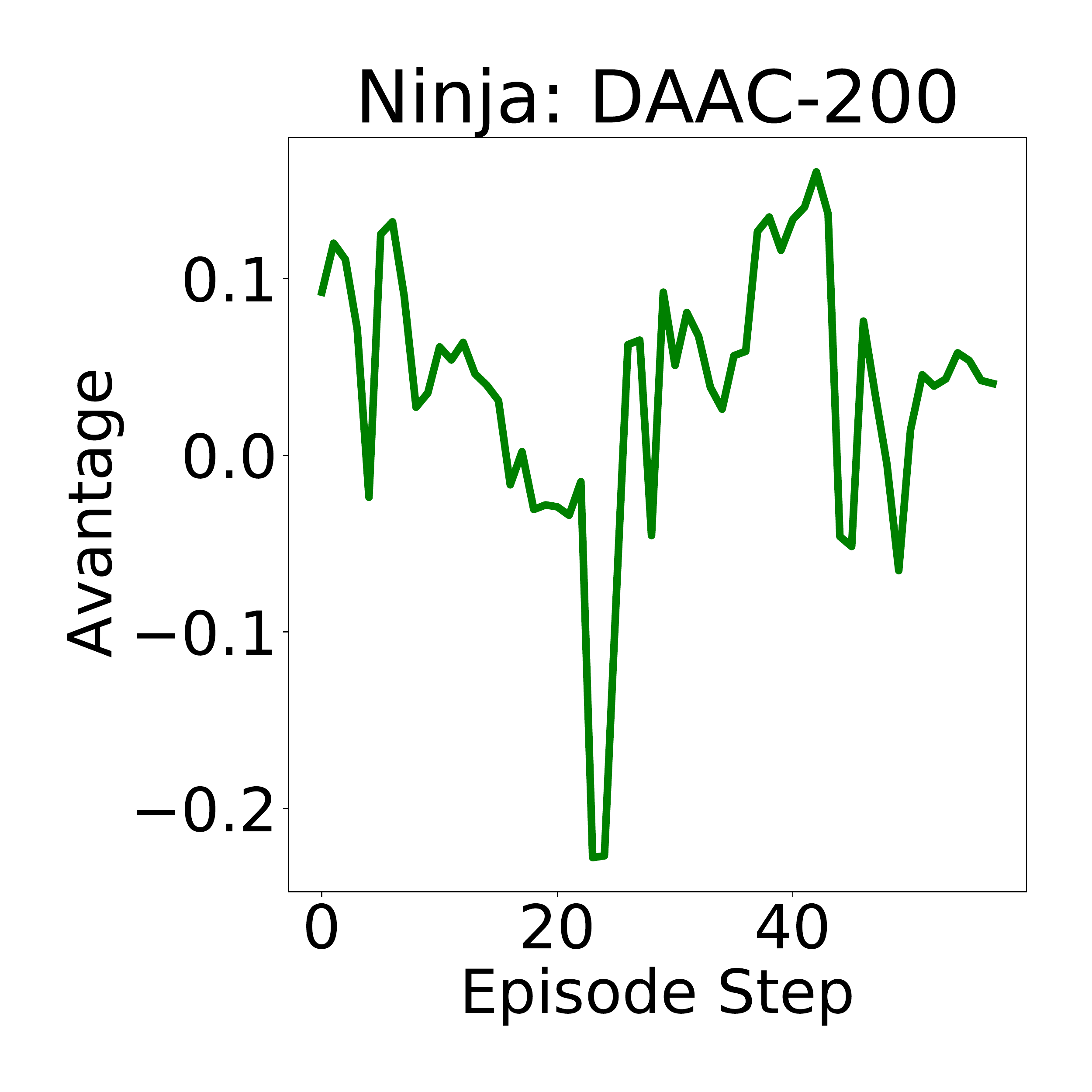}
    \includegraphics[width=0.19\textwidth]{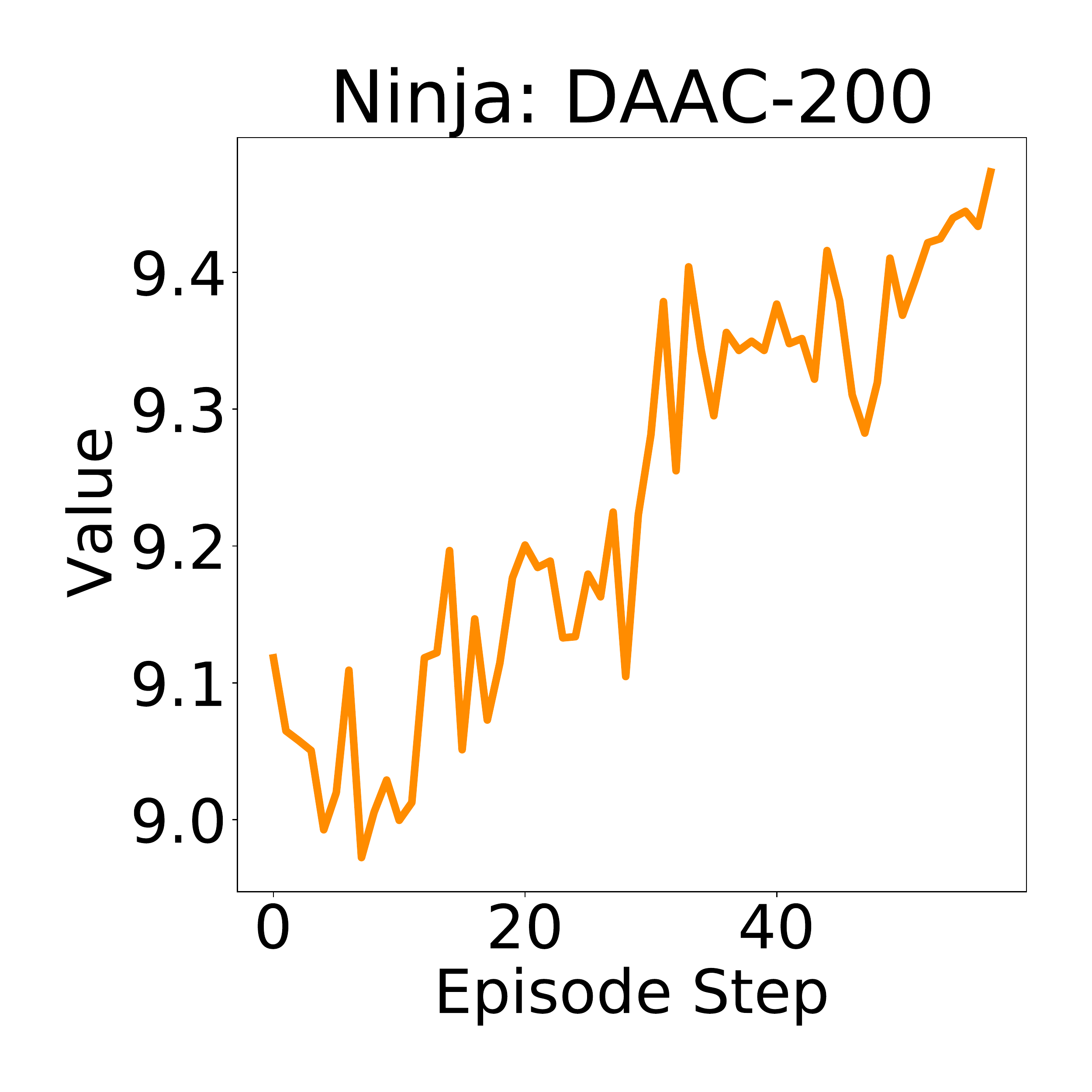}
    \includegraphics[width=0.19\textwidth]{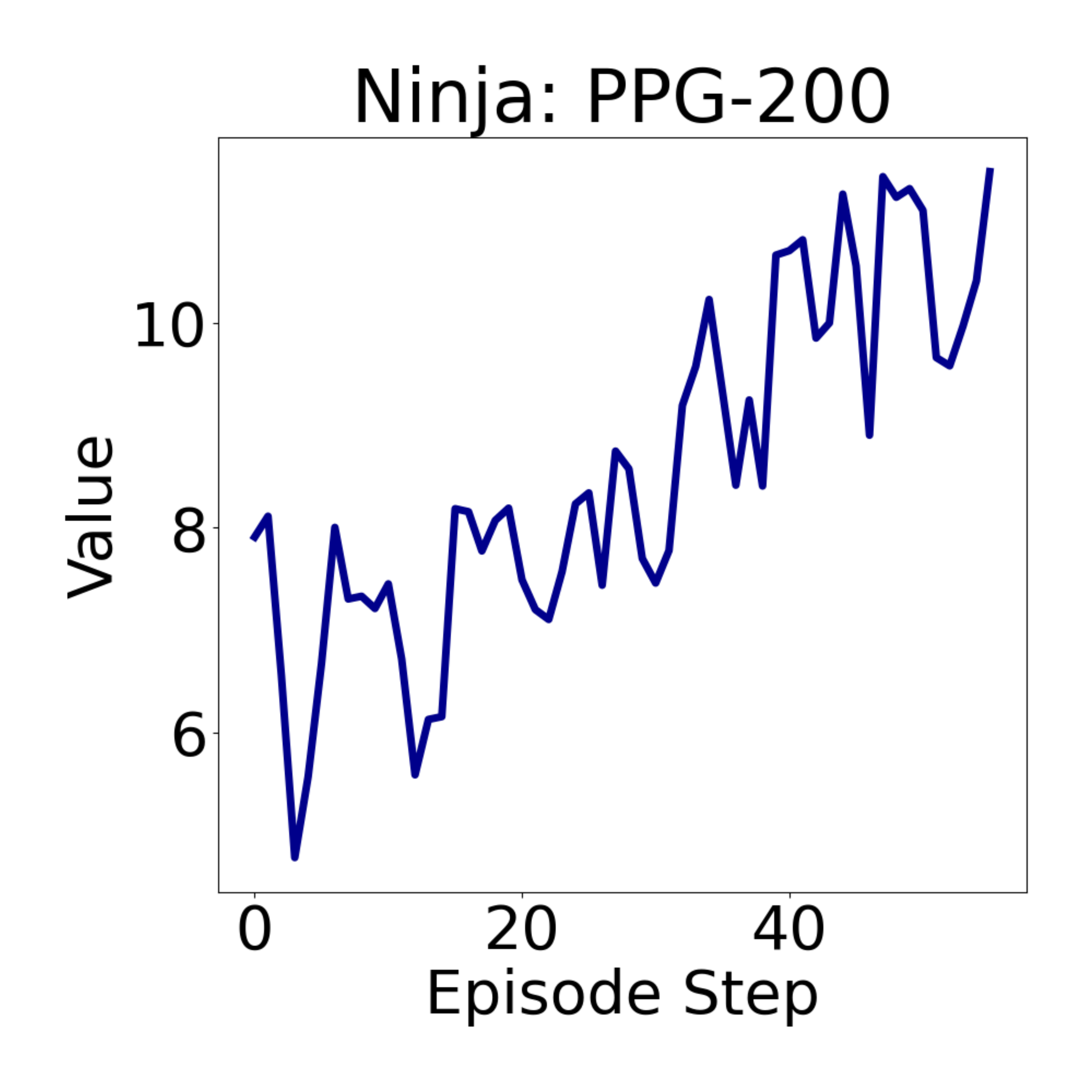}
    % \vspace{-4mm}
    \caption{\textbf{Examples illustrating the timestep-dependence of the value function for a single Ninja level.} The dark and light blue curves show the value as a function of episode step for \ppg{} and \ppo{}, respectively, each trained on 200 levels. Note the near linear relationship, indicating overfitting to the training levels. By contrast, a \ppo{} model (red) trained on 10k levels (thus exhibiting far less overfitting) does not show this relationship. Our \gae{} model trained on 200 levels (green) also lacks this adverse dependence in the advantage prediction which is used for training the policy network, thus is able to generalize better than the PPO model trained on the same amount of data (see Fig.~\ref{fig:procgen_results}). Nevertheless, \gae{}'s value estimate still have a linear trend (orange) but, in contrast to \ppo{} and \ppg{}, this does not negatively affect the policy since we use separate networks for learning the policy and value.}
    \label{fig:step_value_gae_ninja}
\end{figure*}

\begin{figure*}[ht!]
    \centering
    \includegraphics[width=0.19\textwidth]{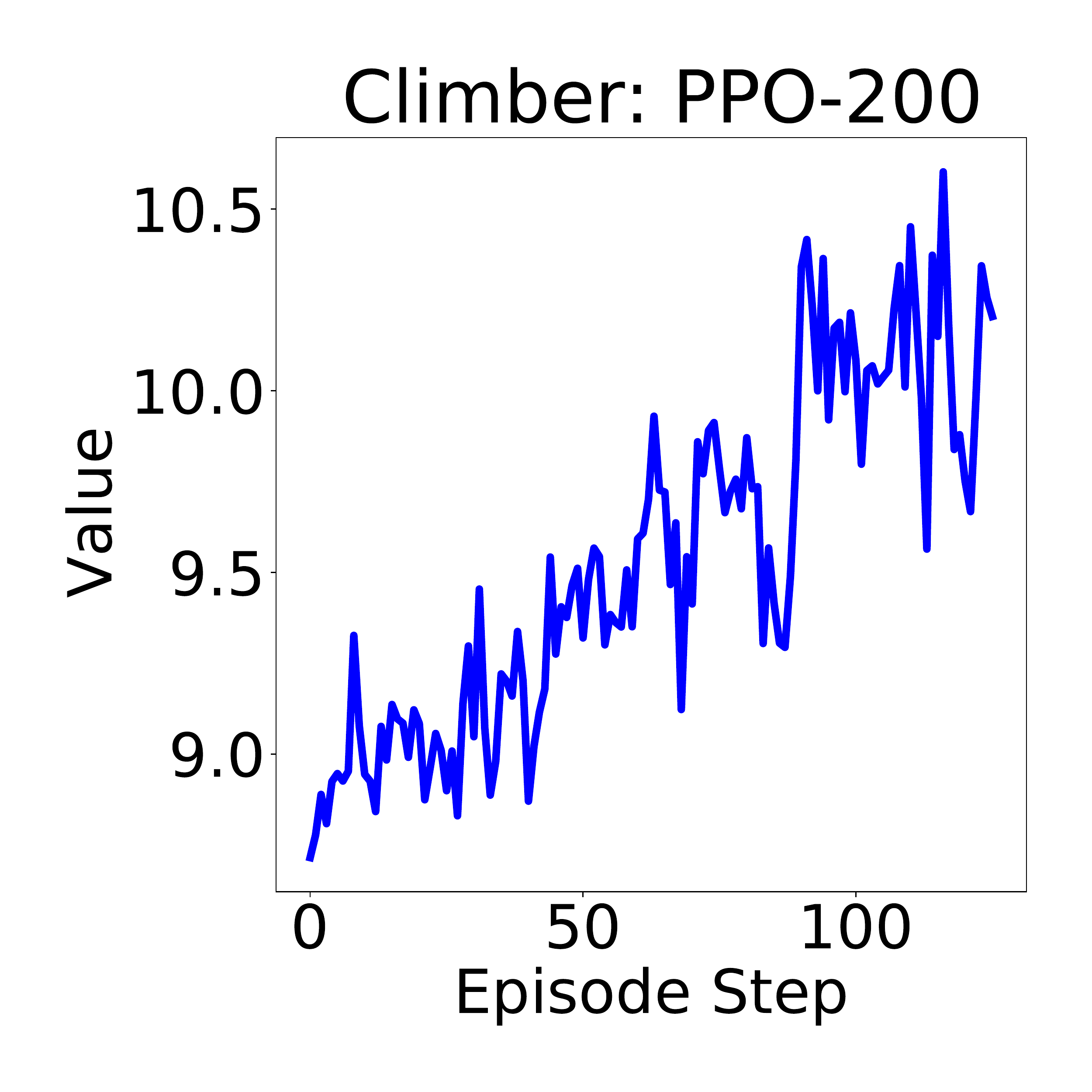}
    \includegraphics[width=0.19\textwidth]{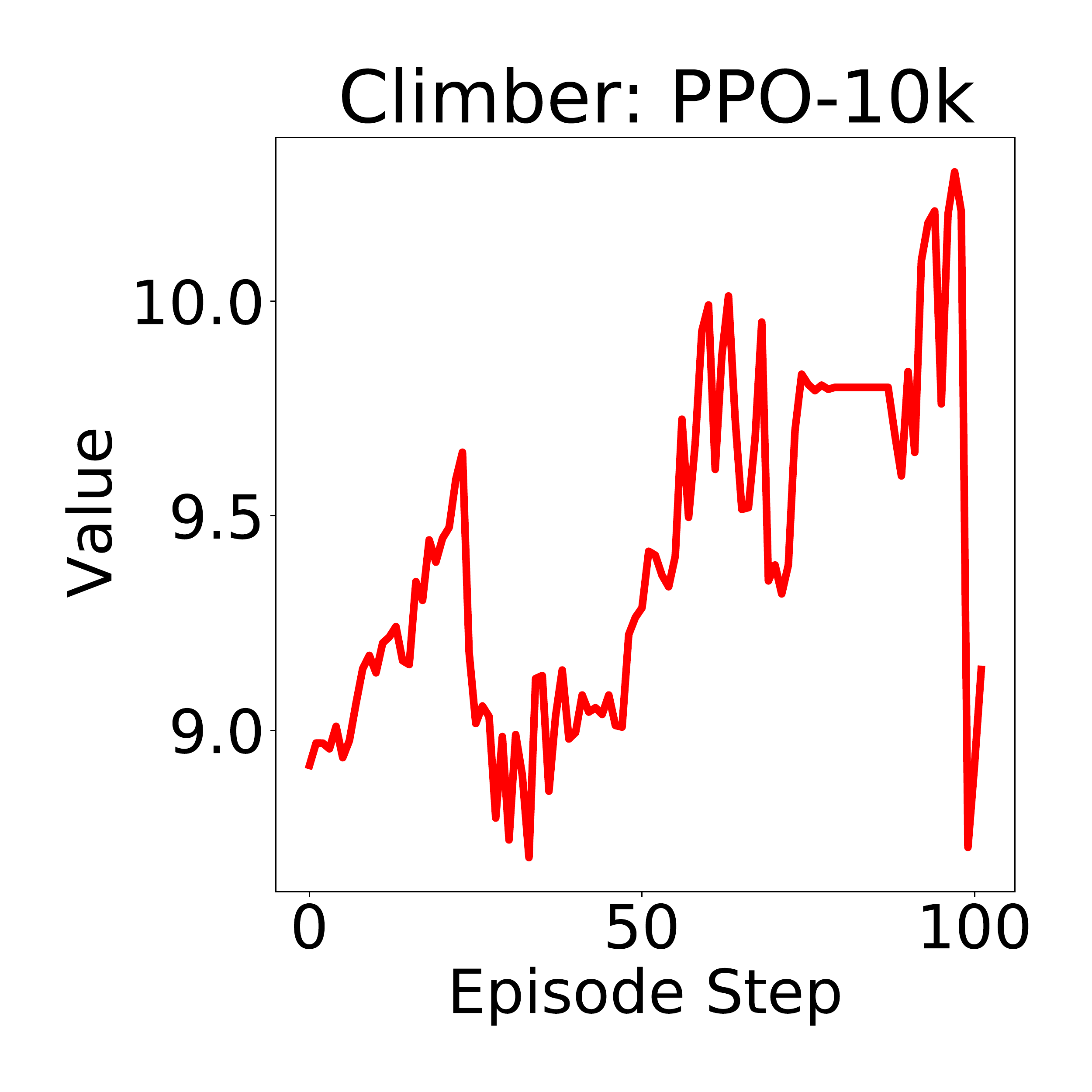}
    \includegraphics[width=0.19\textwidth]{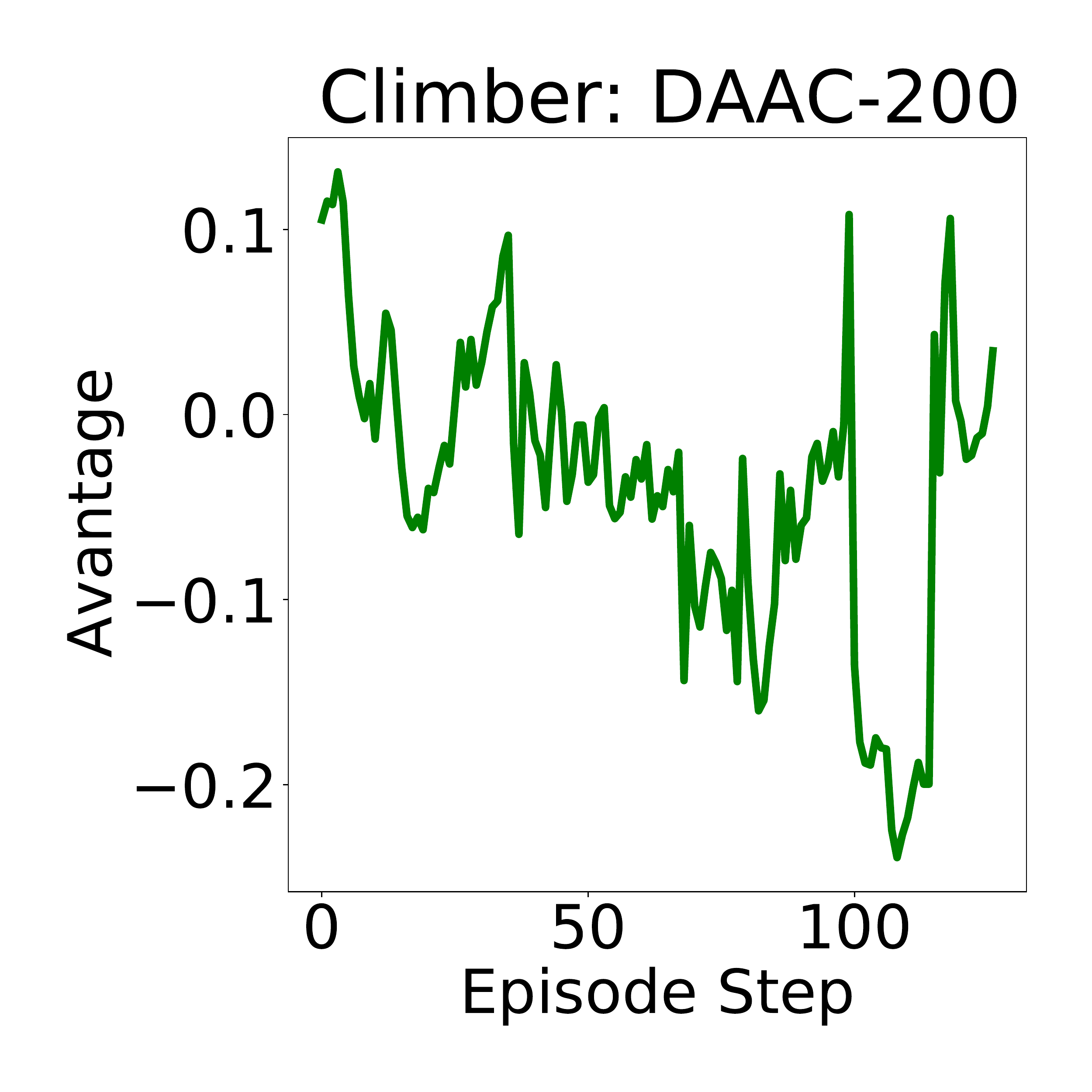}
    \includegraphics[width=0.19\textwidth]{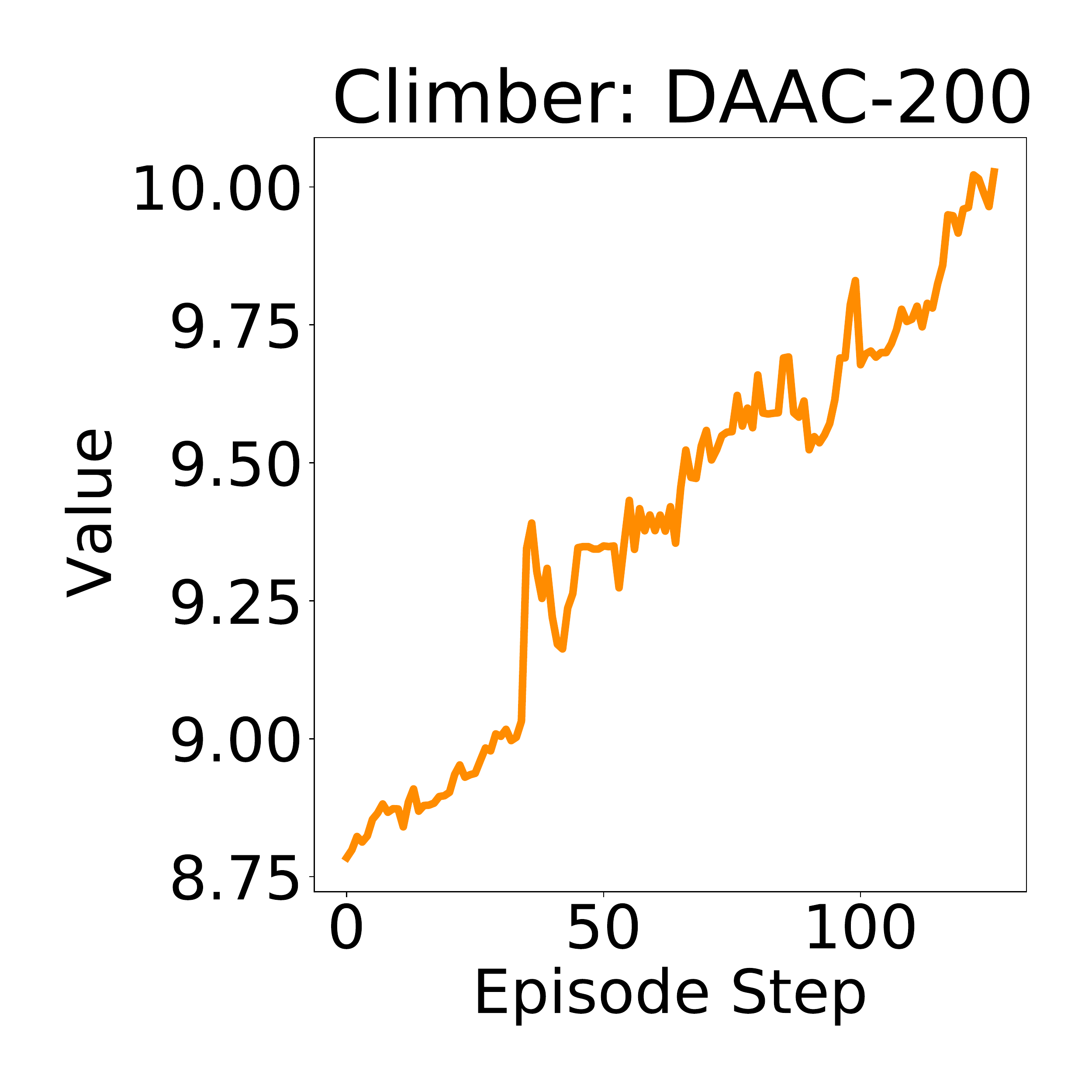}
    \includegraphics[width=0.19\textwidth]{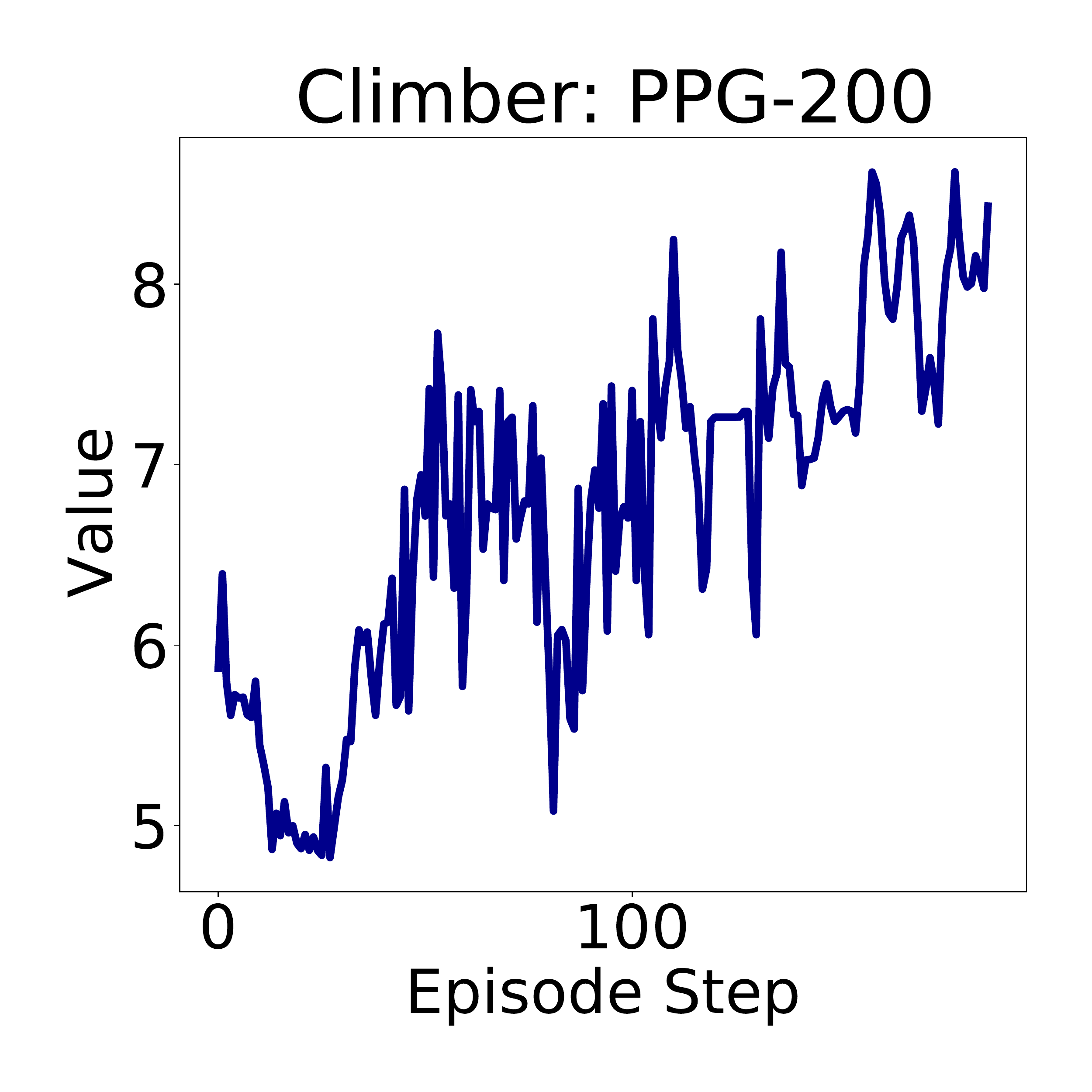}
    % \vspace{-4mm}
    \caption{\textbf{Examples illustrating the timestep-dependence of the value function for a single Climber level.} The dark and light blue curves show the value as a function of episode step for \ppg{} and \ppo{}, respectively, each trained on 200 levels. Note the near linear relationship, indicating overfitting to the training levels. By contrast, a \ppo{} model (red) trained on 10k levels (thus exhibiting far less overfitting) does not show this relationship. Our \gae{} model trained on 200 levels (green) also lacks this adverse dependence in the advantage prediction which is used for training the policy network, thus is able to generalize better than the PPO model trained on the same amount of data (see Fig.~\ref{fig:procgen_results}). Nevertheless, \gae{}'s value estimate still have a linear trend (orange) but, in contrast to \ppo{} and \ppg{}, this does not negatively affect the policy since we use separate networks for learning the policy and value.}
    \label{fig:step_value_gae_climber}
\end{figure*}

\begin{figure*}[ht!]
    \centering
    \includegraphics[width=0.19\textwidth]{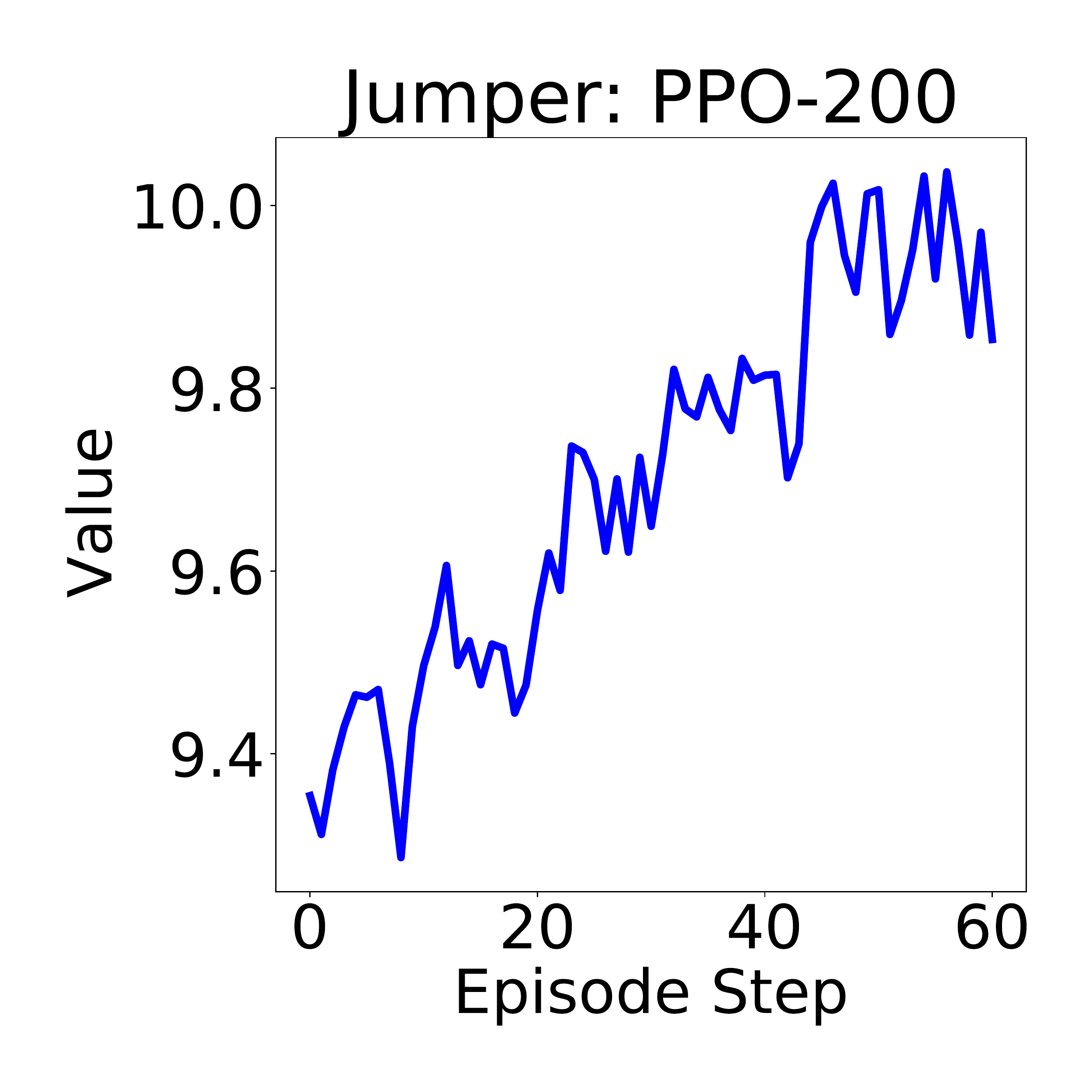}
    \includegraphics[width=0.19\textwidth]{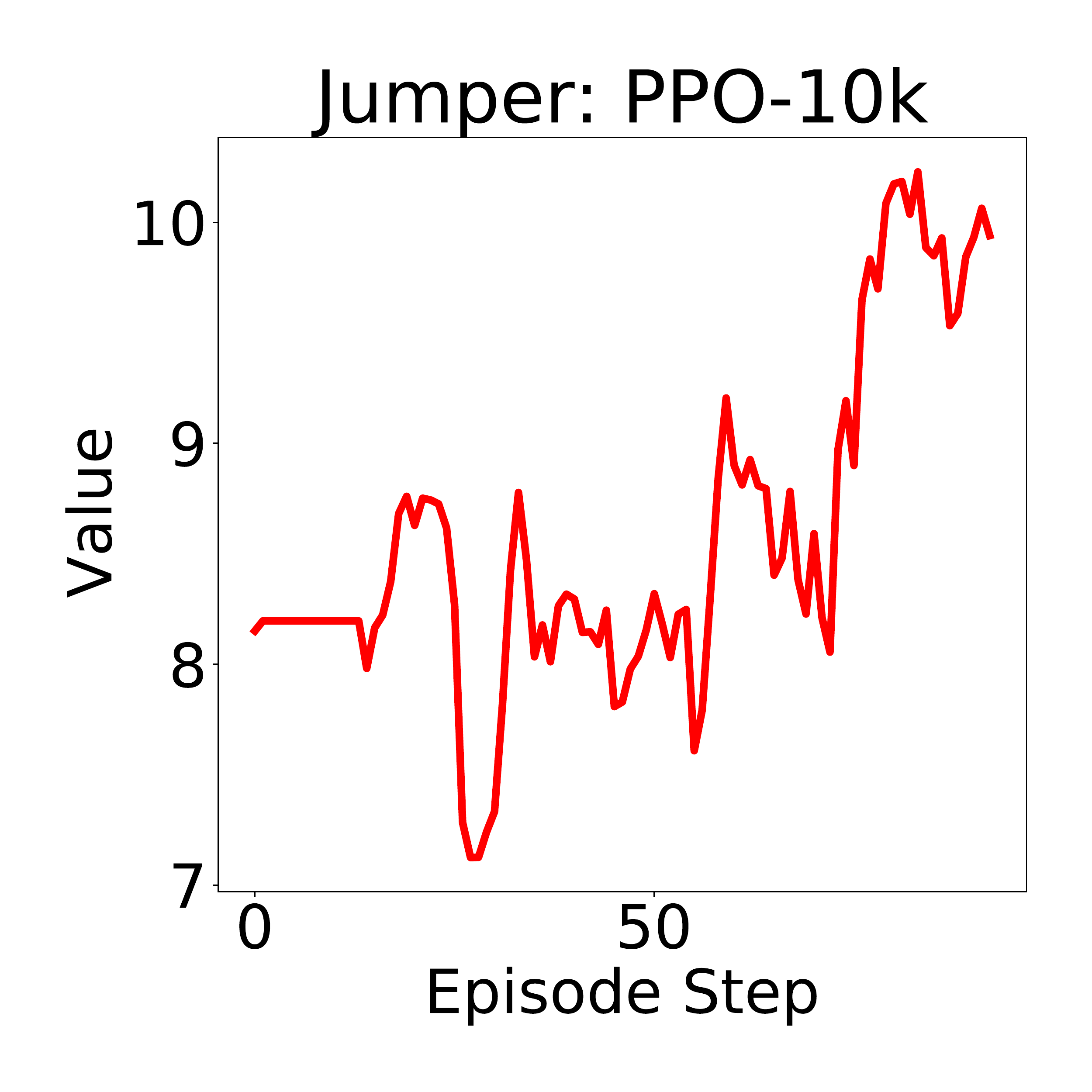}
    \includegraphics[width=0.19\textwidth]{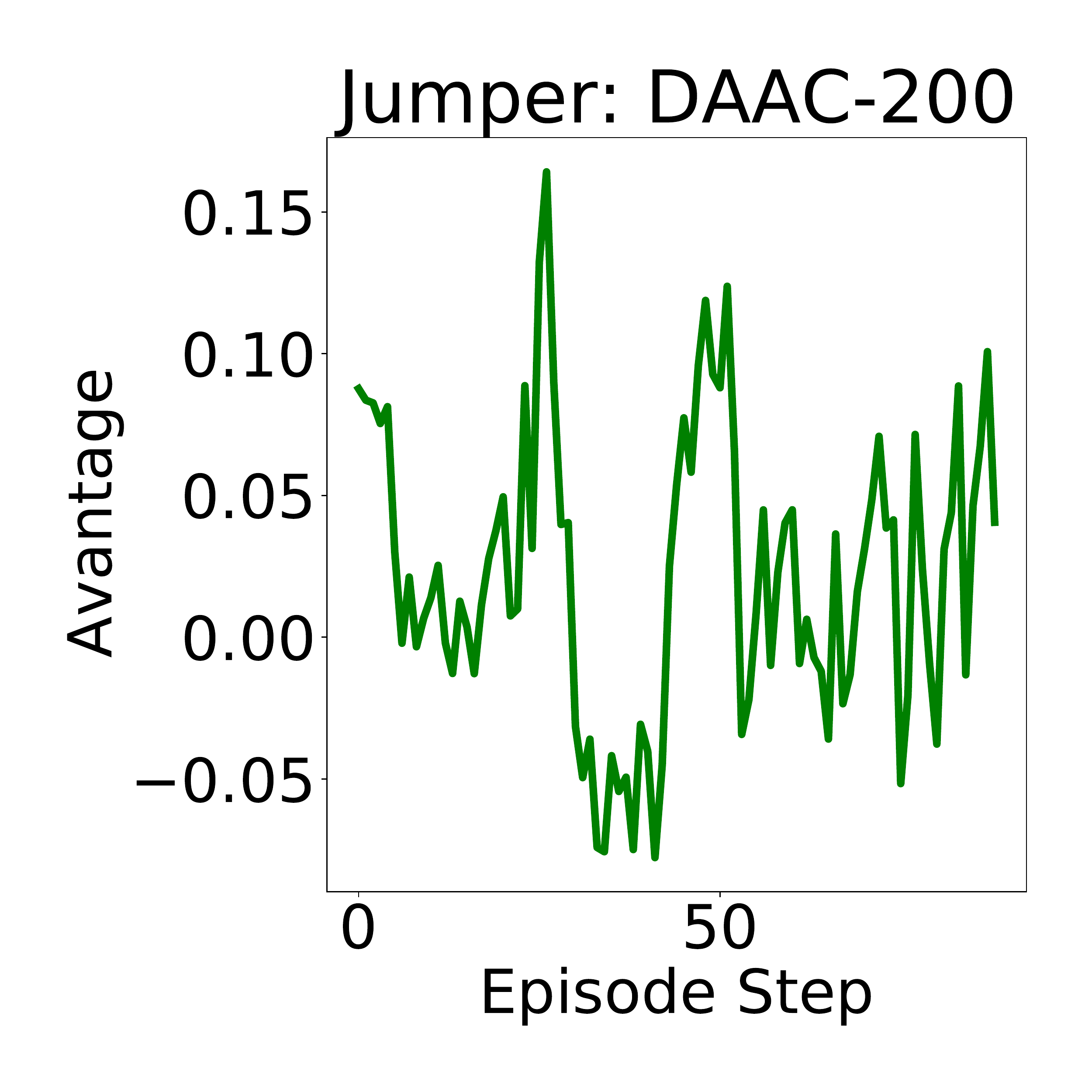}
    \includegraphics[width=0.19\textwidth]{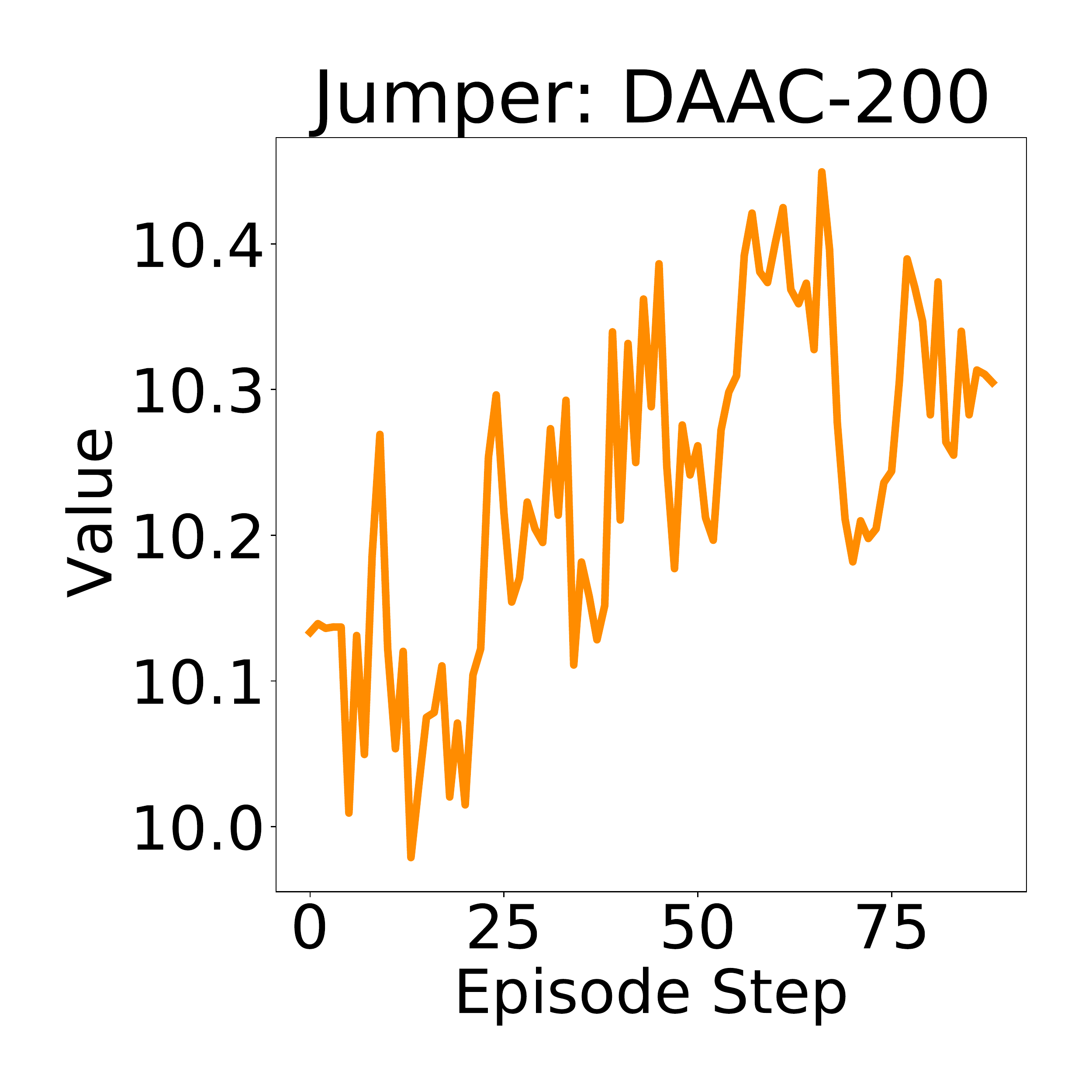}
    \includegraphics[width=0.19\textwidth]{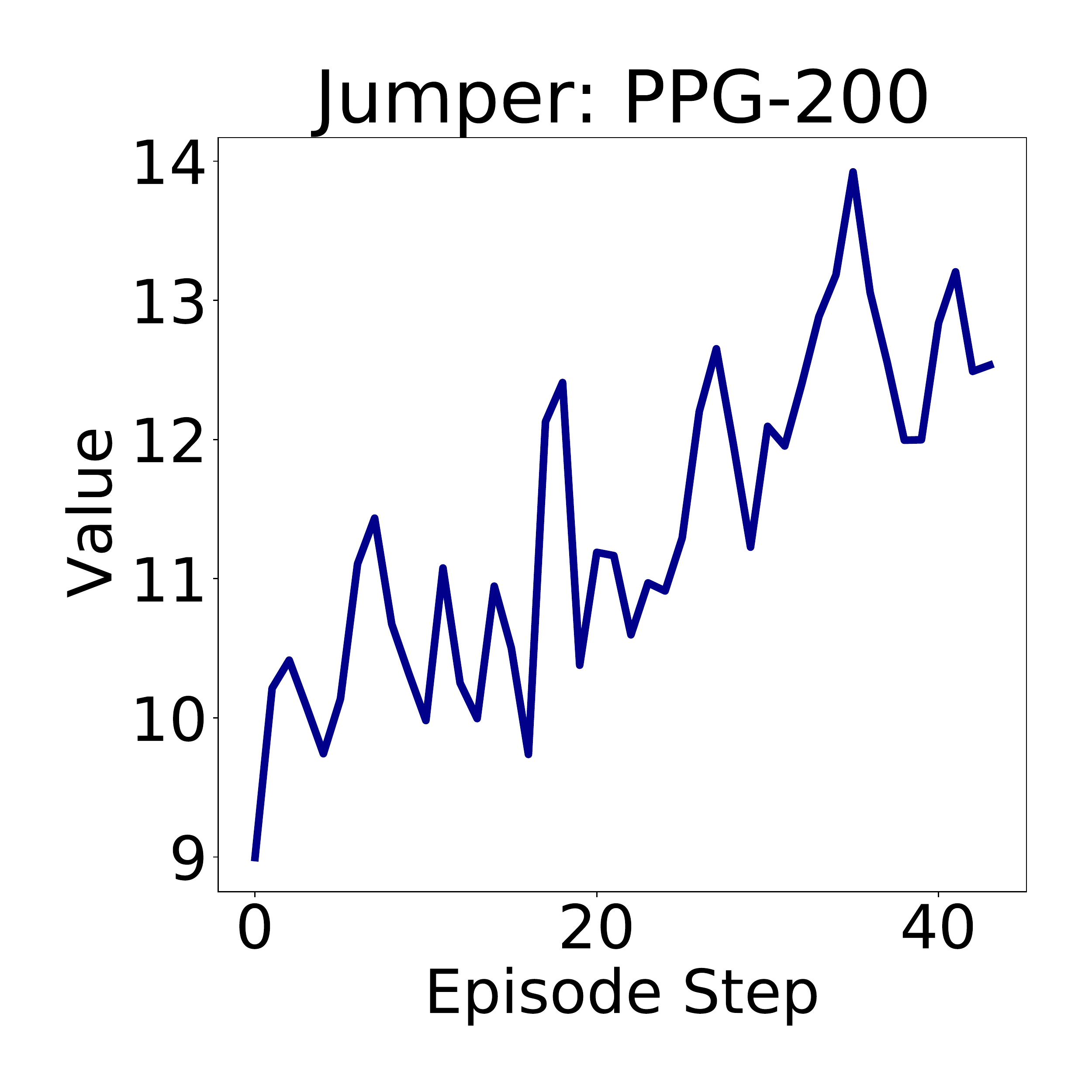}
    % \vspace{-4mm}
    \caption{\textbf{Examples illustrating the timestep-dependence of the value function for a single Jumper level.} The dark and light blue curves show the value as a function of episode step for \ppg{} and \ppo{}, respectively, each trained on 200 levels. Note the near linear relationship, indicating overfitting to the training levels. By contrast, a \ppo{} model (red) trained on 10k levels (thus exhibiting far less overfitting) does not show this relationship. Our \gae{} model trained on 200 levels (green) also lacks this adverse dependence in the advantage prediction which is used for training the policy network, thus is able to generalize better than the PPO model trained on the same amount of data (see Fig.~\ref{fig:procgen_results}). Nevertheless, \gae{}'s value estimate still have a linear trend (orange) but, in contrast to \ppo{} and \ppg{}, this does not negatively affect the policy since we use separate networks for learning the policy and value.}
    \label{fig:step_value_gae_jumper}
\end{figure*}

We also analyze the timestep-dependence of the predictions (\ie{} value or advantage) made by various models over the course of an episode. Figures~\ref{fig:step_value_gae_coinrun},~\ref{fig:step_value_gae_ninja}, ~\ref{fig:step_value_gae_climber}, and~\ref{fig:step_value_gae_jumper} show examples from CoinRun, Ninja, Climber, and Jumper, respectively. While both PPO and PPG (trained on 200 levels) learn a value function which is increasing almost linearly with the episode step, \gae{} (also trained on 200 levels) learns an advantage function which does not have a clear dependence on the episode step. Thus, \gae{} is less prone to overfitting than PPO or PPG. Similarly, a \ppo{} model trained on 10k levels with good generalization performance does not display a linear trend between value and episode step. This suggests there is a trade-off between value accuracy and generalization performance. However, by decoupling the policy and value representations, \gae{} is able to achieve both accurate value predictions and good generalization abilities.

%%%%%%%%%%%%%%%%%%%%%%%%%%%%%%%%
\clearpage
\section{Value Variance}
\label{app:val_var}

\begin{figure*}[h!]
    \centering
    \includegraphics[width=0.25\textwidth]{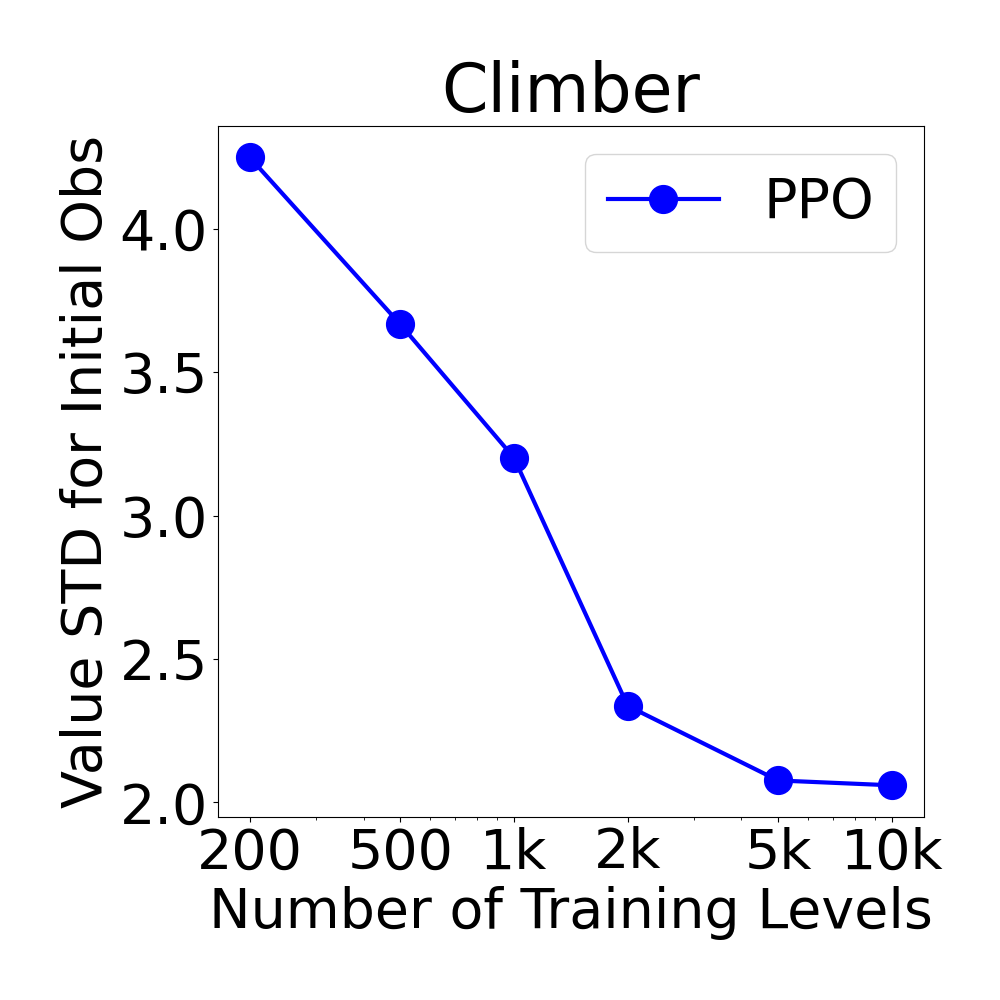}
    \includegraphics[width=0.25\textwidth]{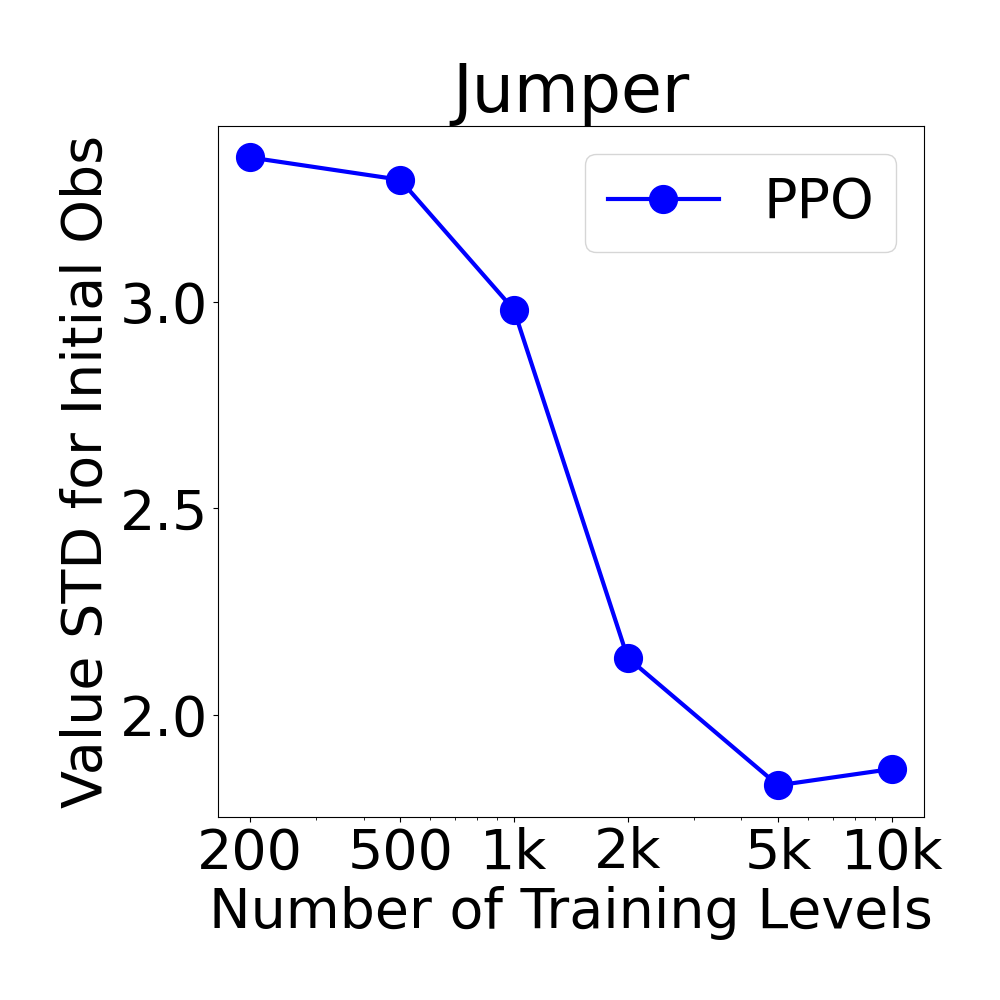}
    \includegraphics[width=0.25\textwidth]{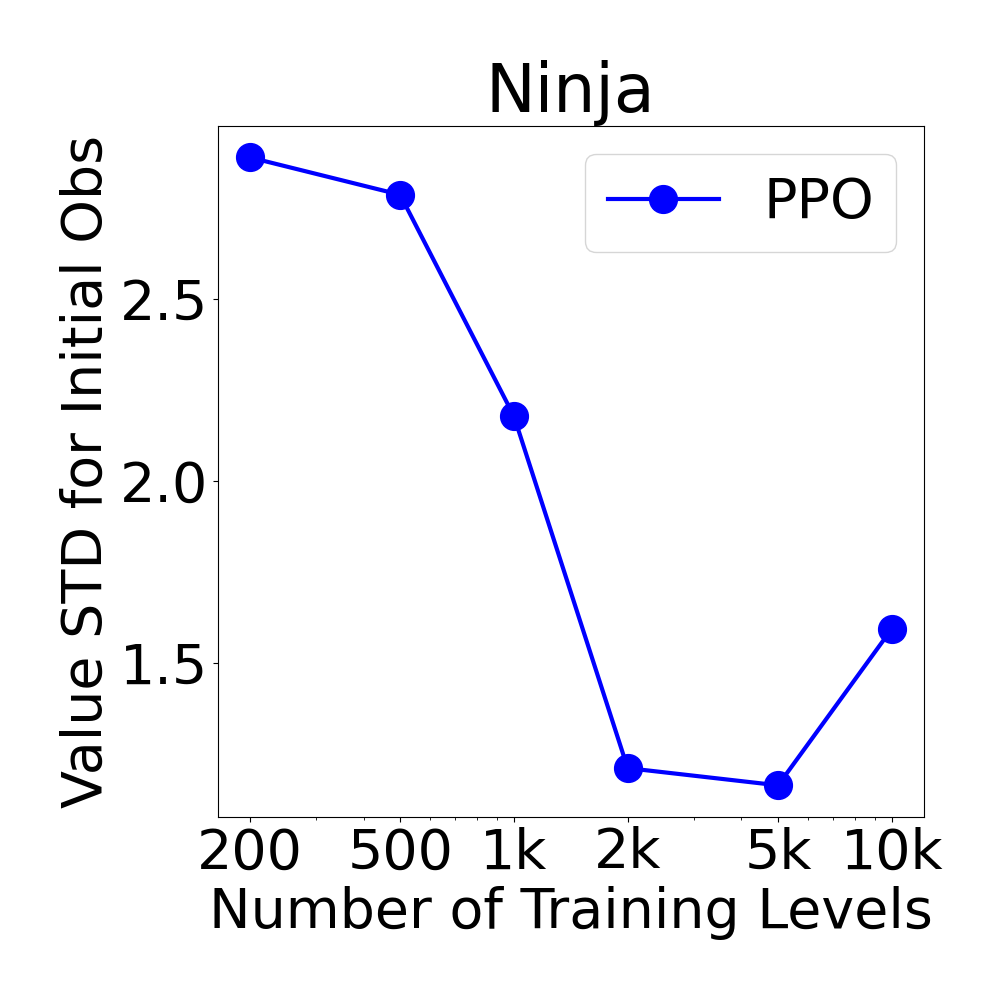}
    \caption{\textbf{Standard deviation of the predicted values for the initial observations from 200 levels, as a function of the number of training levels.} From left to right: Climber, Jumper, and Ninja. The 200 levels used to compute the standard deviation were part of the training set for all the agents. Note that, in general, the variance of the value decreases with the number of training levels. This is consistent with our claim that when sharing parameters for the policy and value, models which generalize better predict close values for observations that are semantically similar (\eg{} the initial observation) since they learn representations which are less prone to overfitting to level-specific features.}
    \label{fig:value_variance}
\end{figure*}

In this section, we look at the variance in the predicted values for the initial observation, across all training levels. In partially-observed procedurally generated environments, there should be no way of telling how difficult or long a level is (and thus how much reward is expected) from the initial observation alone since the end of the level cannot be seen. Thus, we would expect a model with strong generalization to predict similar values for the initial observation irrespective of the environment instance. If this is not the case and the model uses a common representation for the policy and value function, the policy is likely to overfit to the training environments. As Figure~\ref{fig:value_variance} shows, the variance decreases with the number of levels used for training. This is consistent with our observation that models with better generalization predict more similar values for observations that are semantically different such as the initial observation. In contrast, models trained on a low number of levels, memorize the value of the initial observation for each level, leading to poor generalization in new environments. Note that we chose to illustrate this phenomenon on three of the Procgen games (\ie{} Climber, Jumper, and Ninja) where it is more apparent due to their partial-observability and substantial level diversity (in terms of length).

%%%%%%%%%%%%%%%%%%%%%%%%%%%%%%%%
% \clearpage
\section{Robustness to Spurious Correlations}
\label{app:robustness}

In this section, we investigate how robust the learned features, policies, and predicted values or advantages, are to spurious correlations. To answer this question, we measure how much the features, policies, and predictions vary when the background of an observation changes. Note that the change in background does not change the underlying state of the environment but only its visual aspect. Hence, a change in background should not modify the agent's policy or learned representation. For this experiment, we collect a buffer of 1000 observations from 20 training levels, using a PPO agent trained on 200 levels. Then, we create 10 extra versions of each observation by changing the background. We then measure the L1-norm and L2-norm between the learned representation (\ie{} final vector before the policy's softmax layer) of the original observation and each of its other versions. We also compute the difference in predicted outputs (\ie{} values for \ppo{} and \ppg{} or advantages for \gae{} and \ordergae{}) and the Jensen-Shannon Divergence (JSD) between the policies. For all these metrics, we first take the mean for all 10 backgrounds to obtain a single point for each observation, and then we report the mean and standard deviation across all 1000 observations, resulting in an average statistic of how much these metrics change as a result of varying the background. 

\begin{figure*}[ht!]
    \centering
    \includegraphics[width=0.24\textwidth]{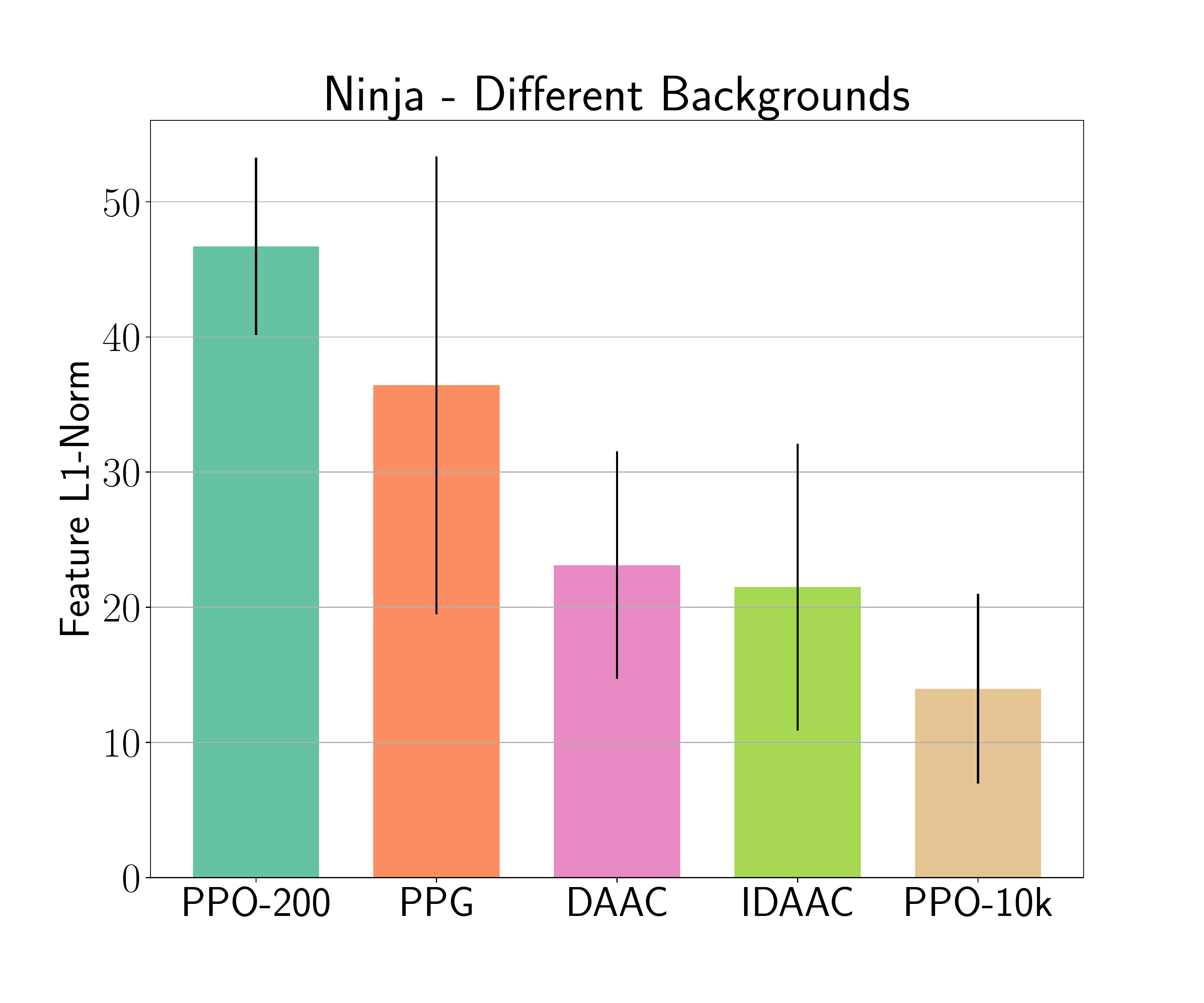}
    \includegraphics[width=0.24\textwidth]{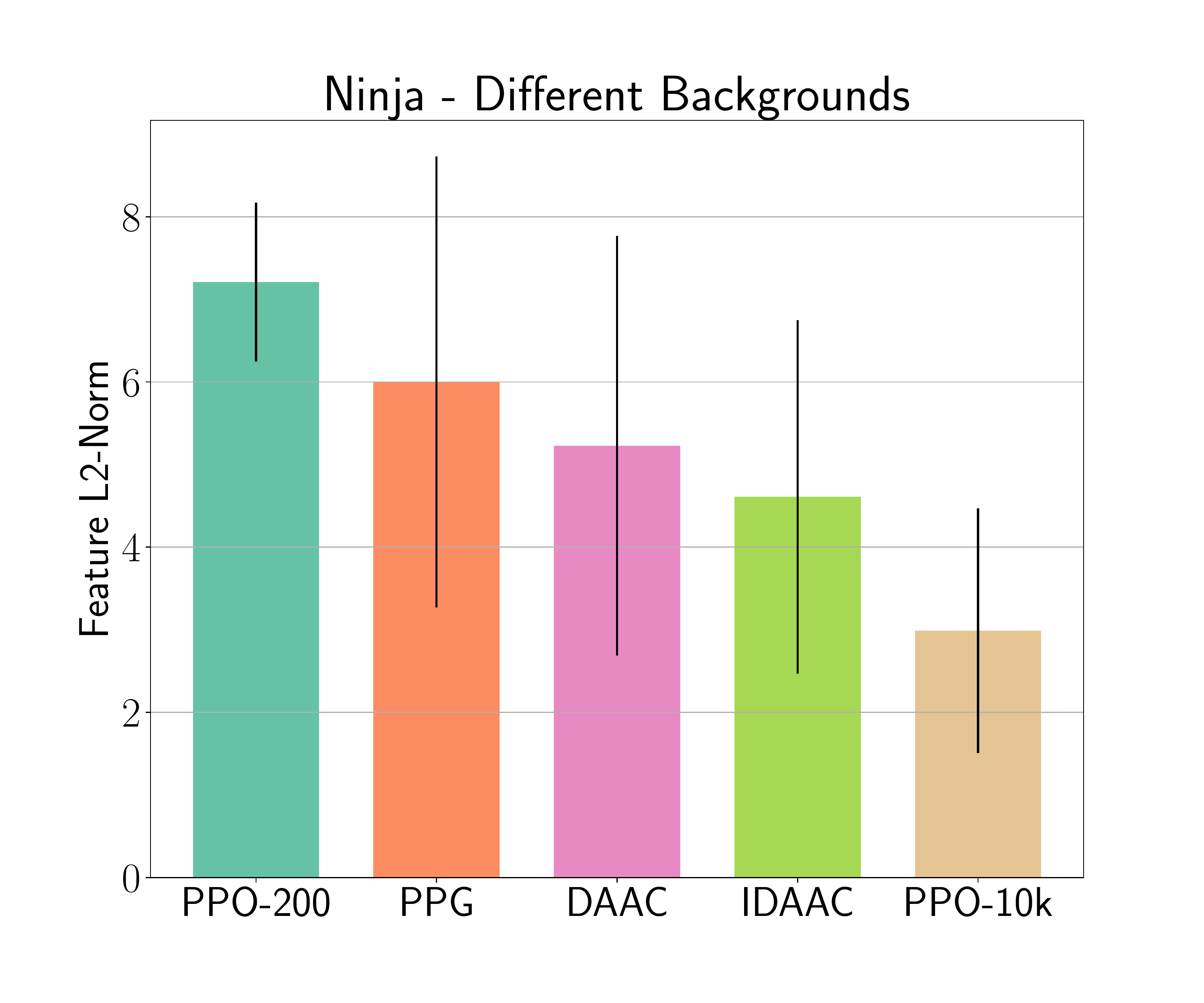}
    \includegraphics[width=0.24\textwidth]{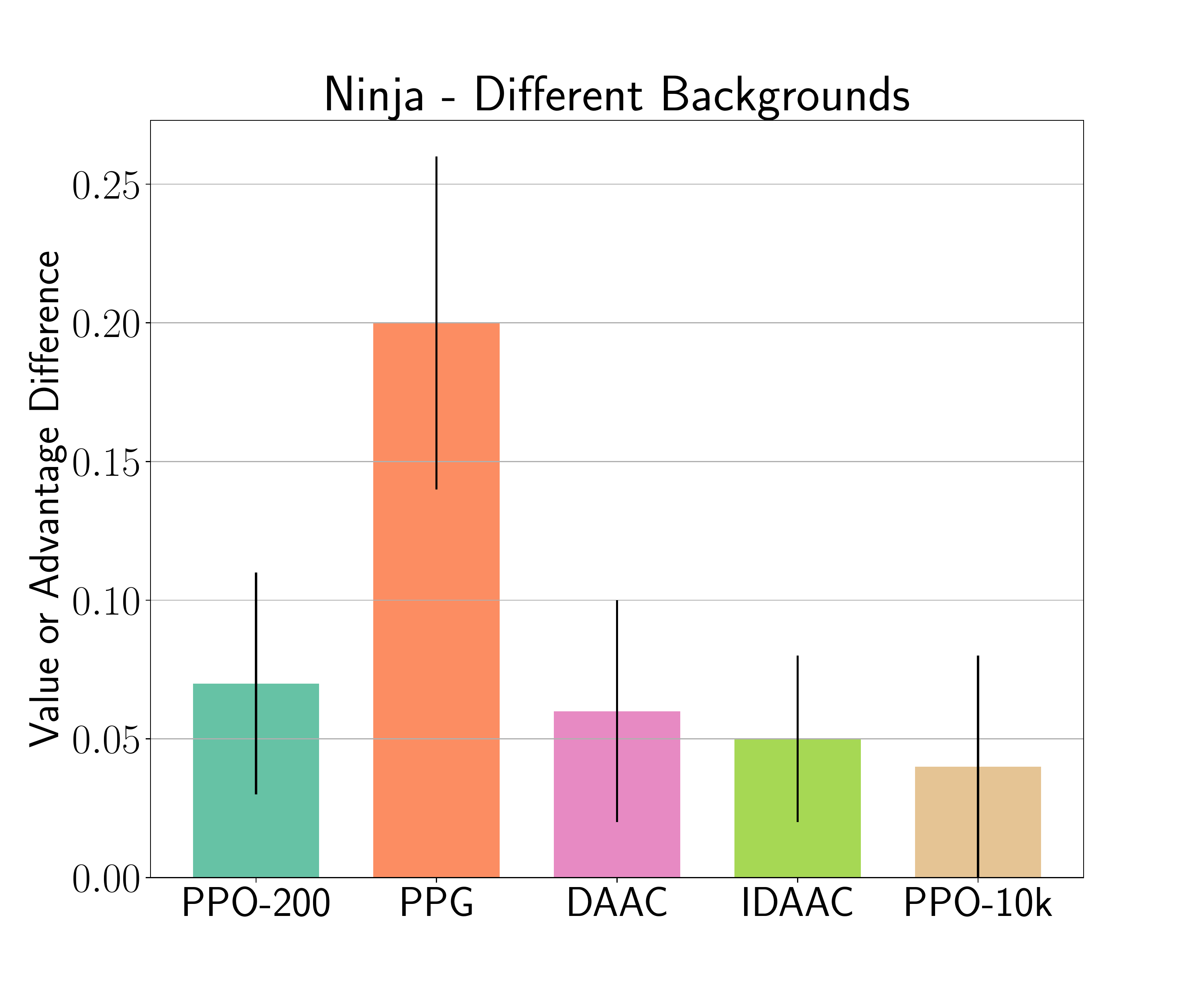}
    \includegraphics[width=0.24\textwidth]{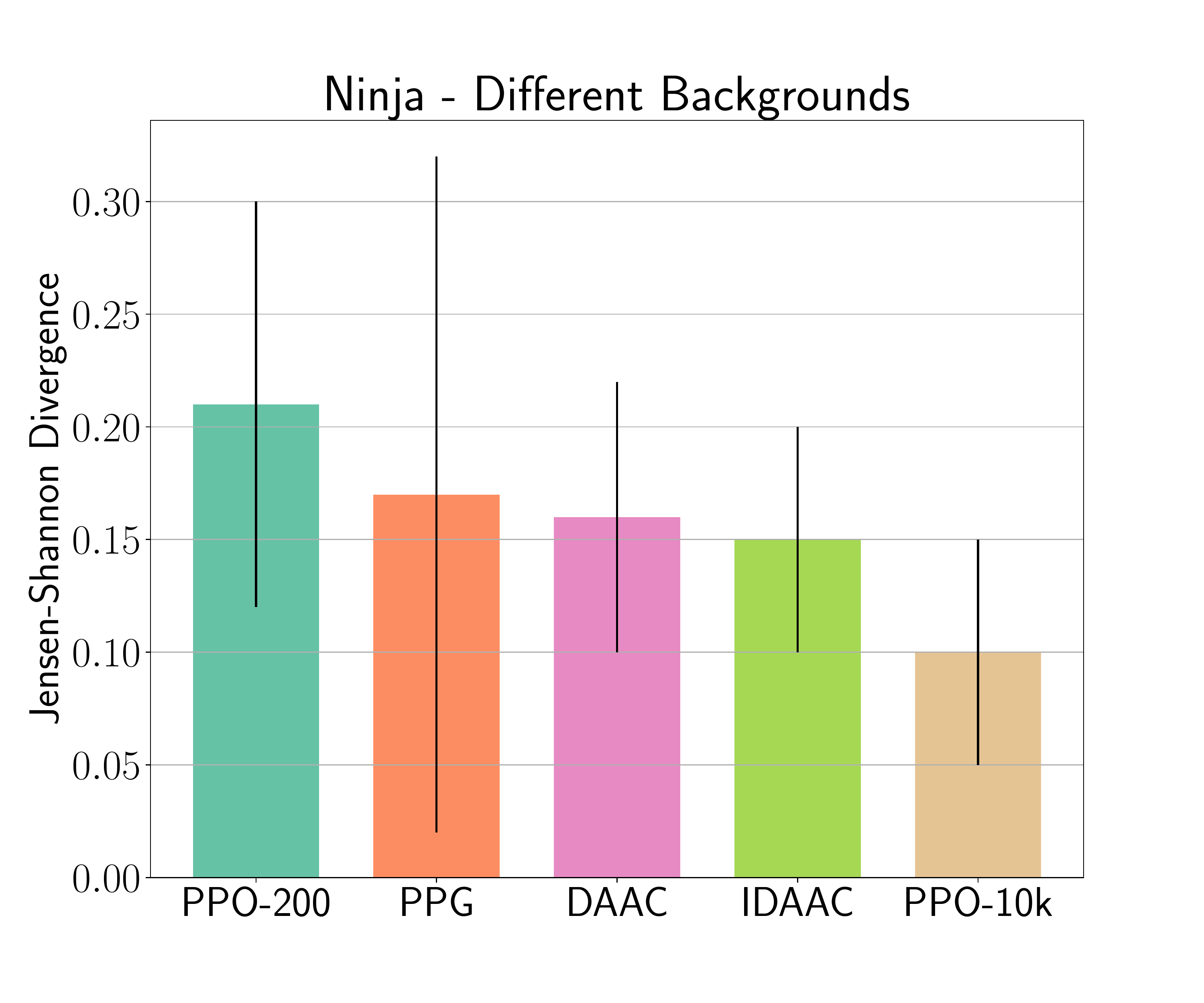}
    \caption{\textbf{Variations in the learned features, policies, and values or advantages when changing the background in Ninja.} From left to right we report the L1 and L2-norm for the features, the value or advantage difference, and the Jensen-Shannon Divergence for the policy. We compare \ppo{} trained on 200 and 10k levels with \ppg{}, \gae{}, and \ordergae{}. Our models are more robust to changes in the background (which does not affect the state). The means and standard deviations were computed over 10 different backgrounds.}
    \label{fig:difback_ninja}
\end{figure*}

\begin{figure*}[ht!]
    \centering
    \includegraphics[width=0.24\textwidth]{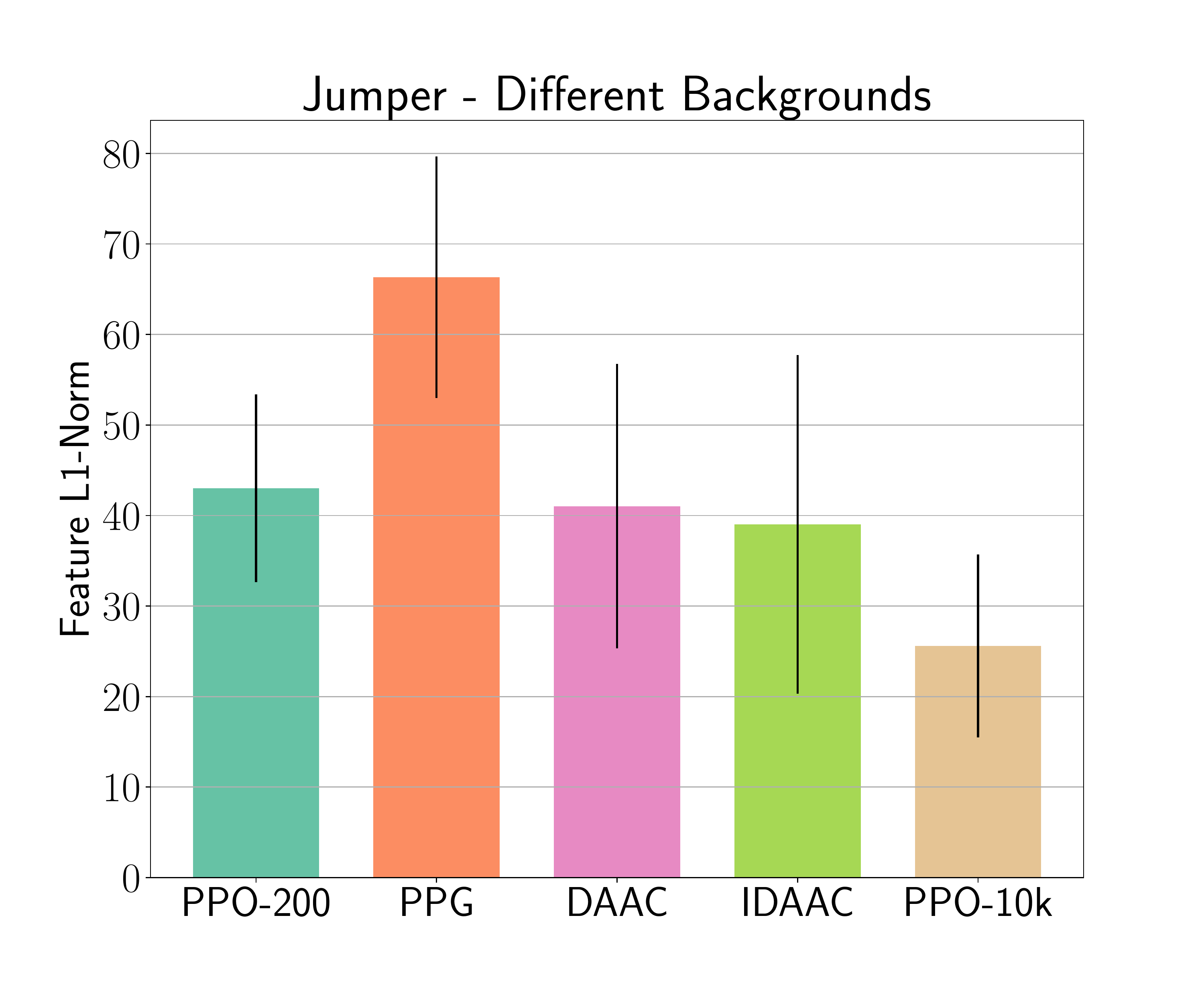}
    \includegraphics[width=0.24\textwidth]{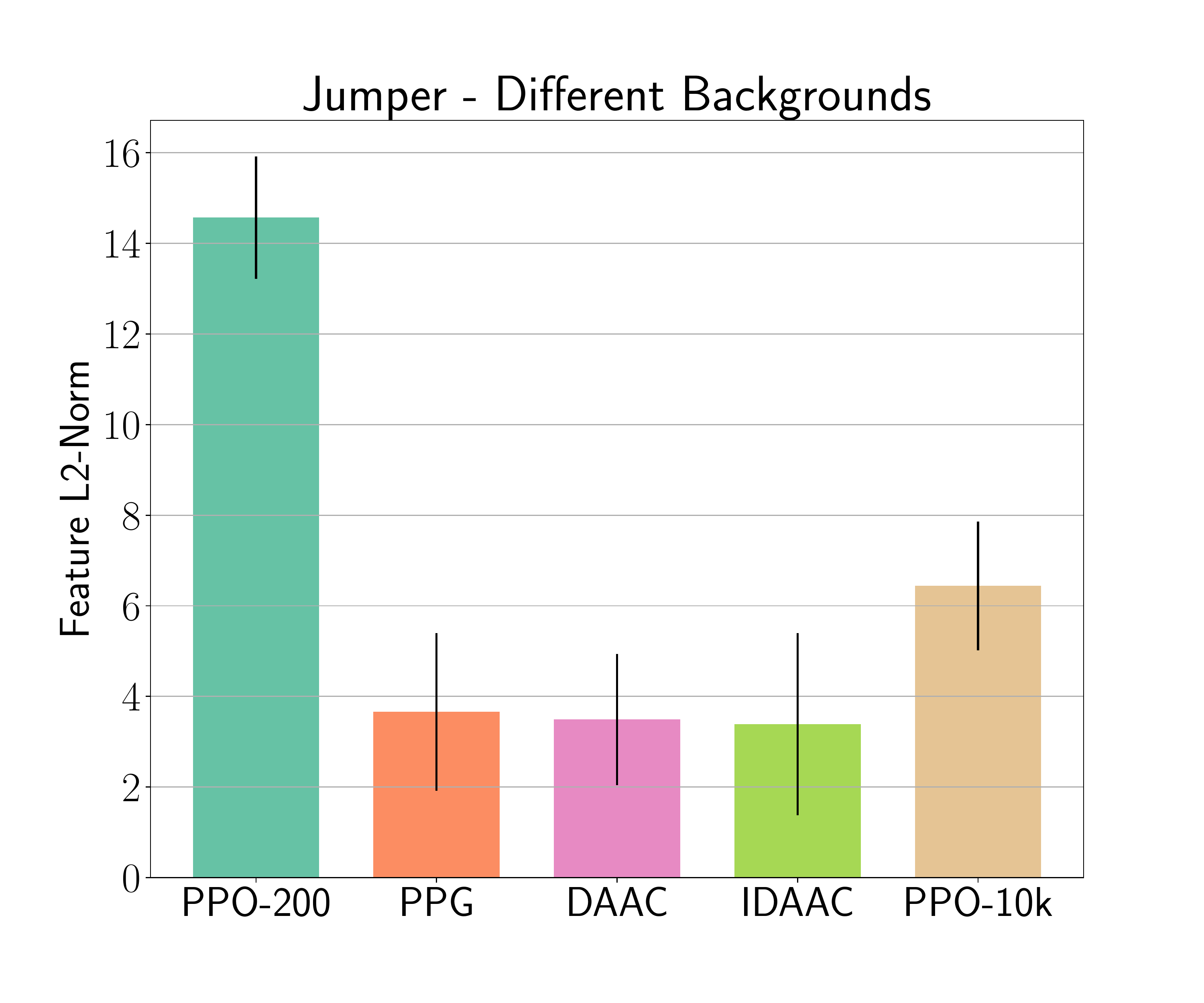}
    \includegraphics[width=0.24\textwidth]{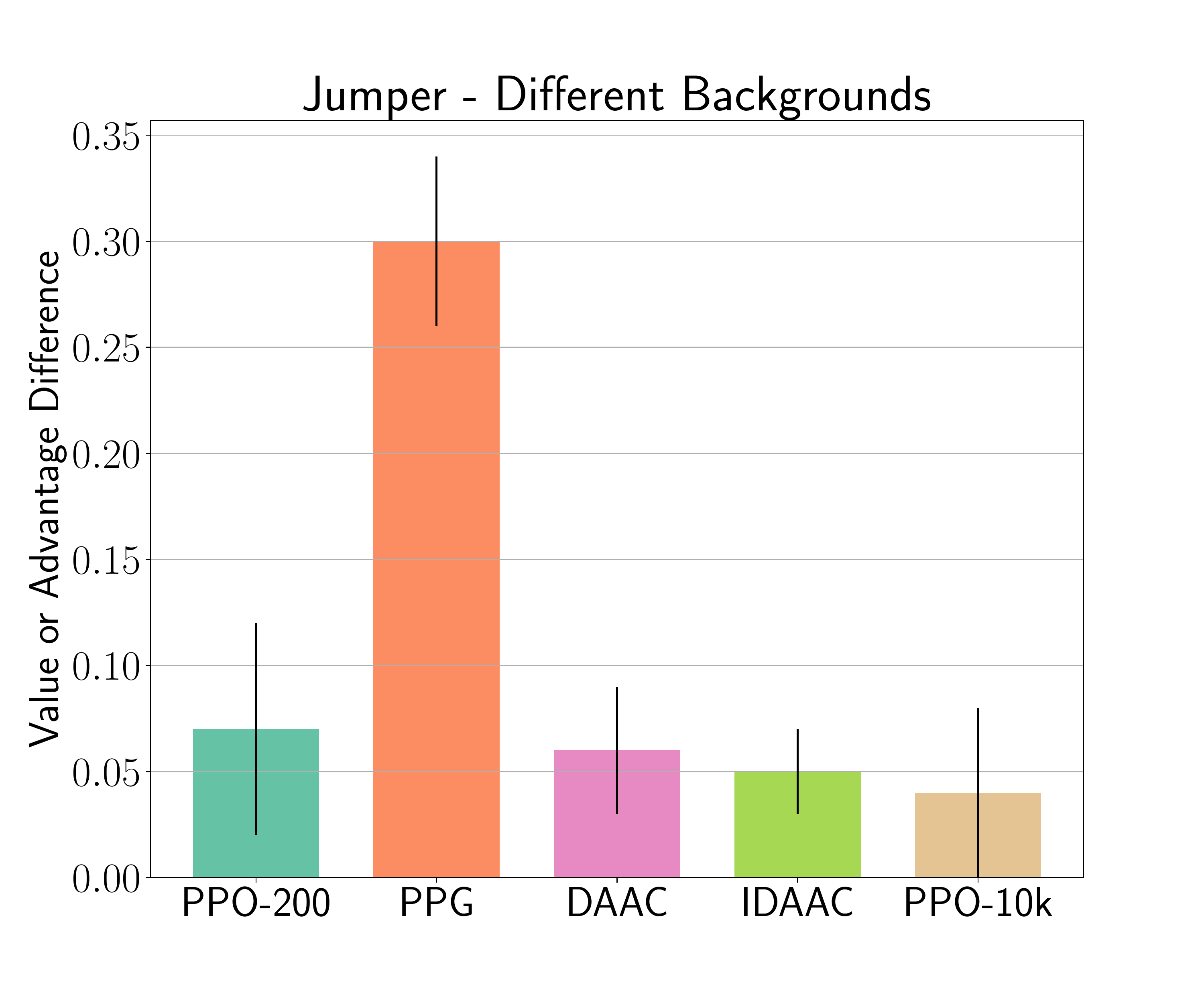}
    \includegraphics[width=0.24\textwidth]{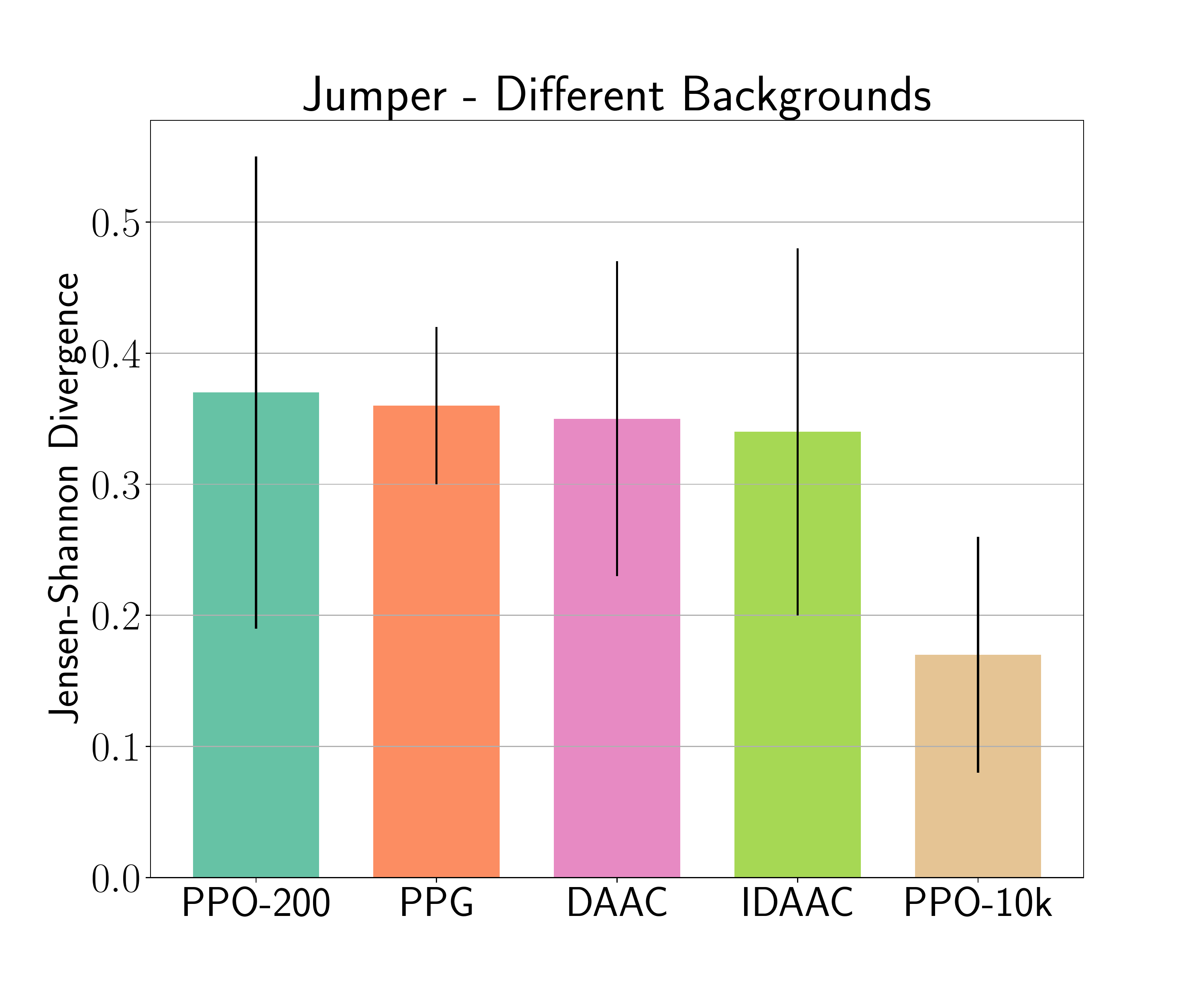}
    \caption{\textbf{Variations in the learned features, policies, and values or advantages when changing the background in Jumper.} From left to right we report the L1 and L2-norm for the features, the value or advantage difference, and the Jensen-Shannon Divergence for the policy. We compare \ppo{} trained on 200 and 10k levels with \ppg{}, \gae{}, and \ordergae{}. Our models are more robust to changes in the background (which does not affect the state). The means and standard deviations were computed over 10 different backgrounds.}
    \label{fig:difback_jumper}
\end{figure*}

\begin{figure*}[ht!]
    \centering
    \includegraphics[width=0.24\textwidth]{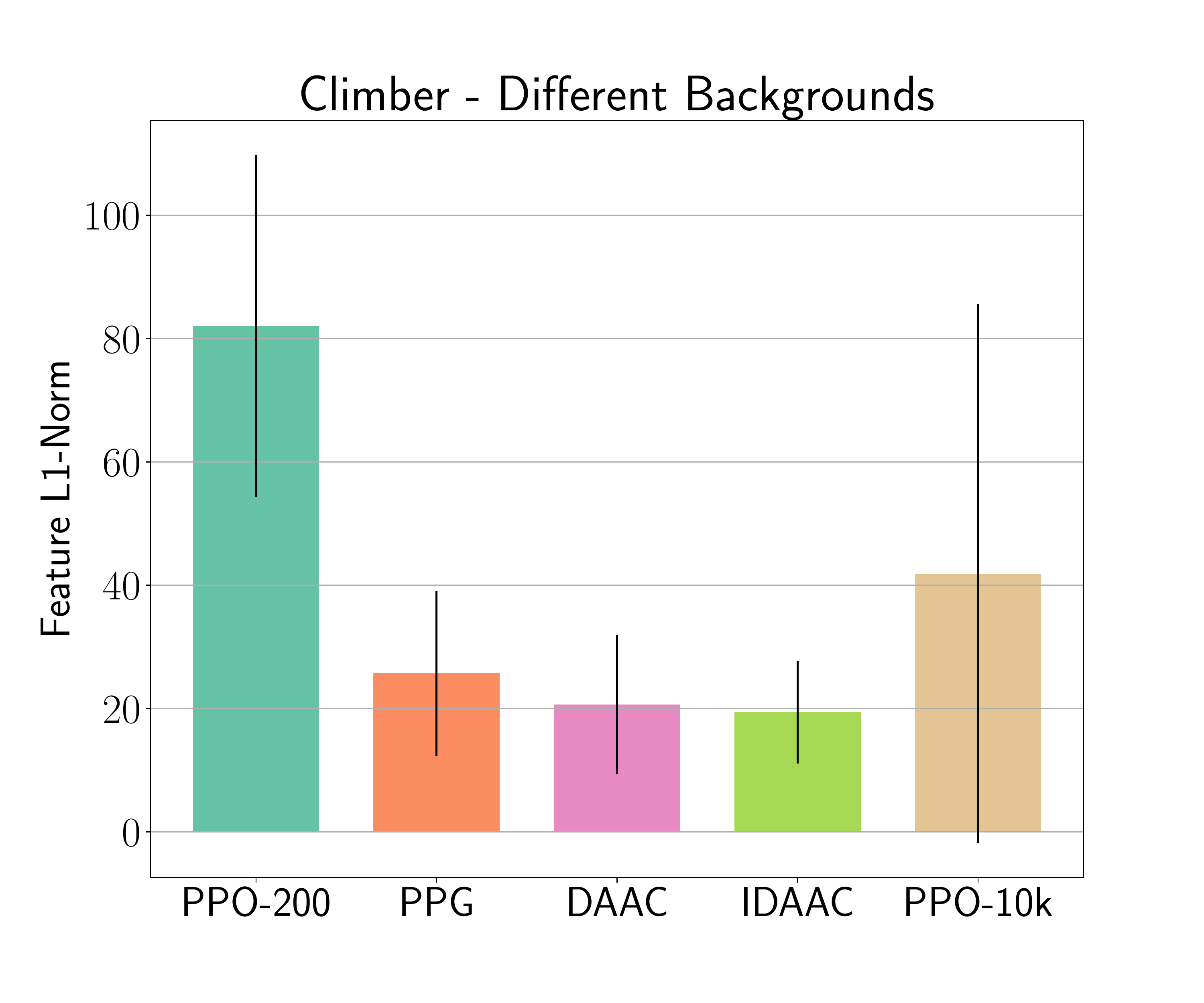}
    \includegraphics[width=0.24\textwidth]{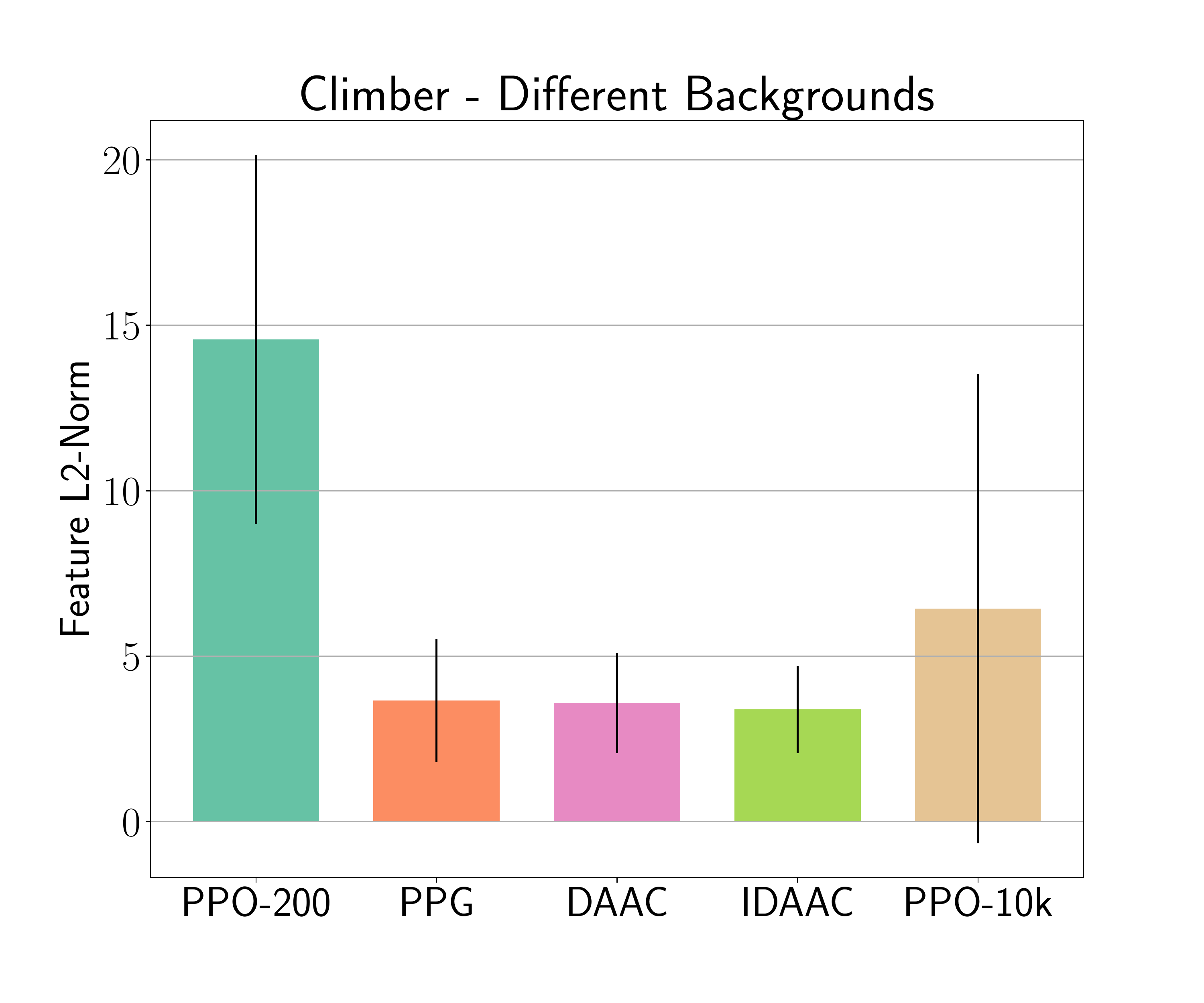}
    \includegraphics[width=0.24\textwidth]{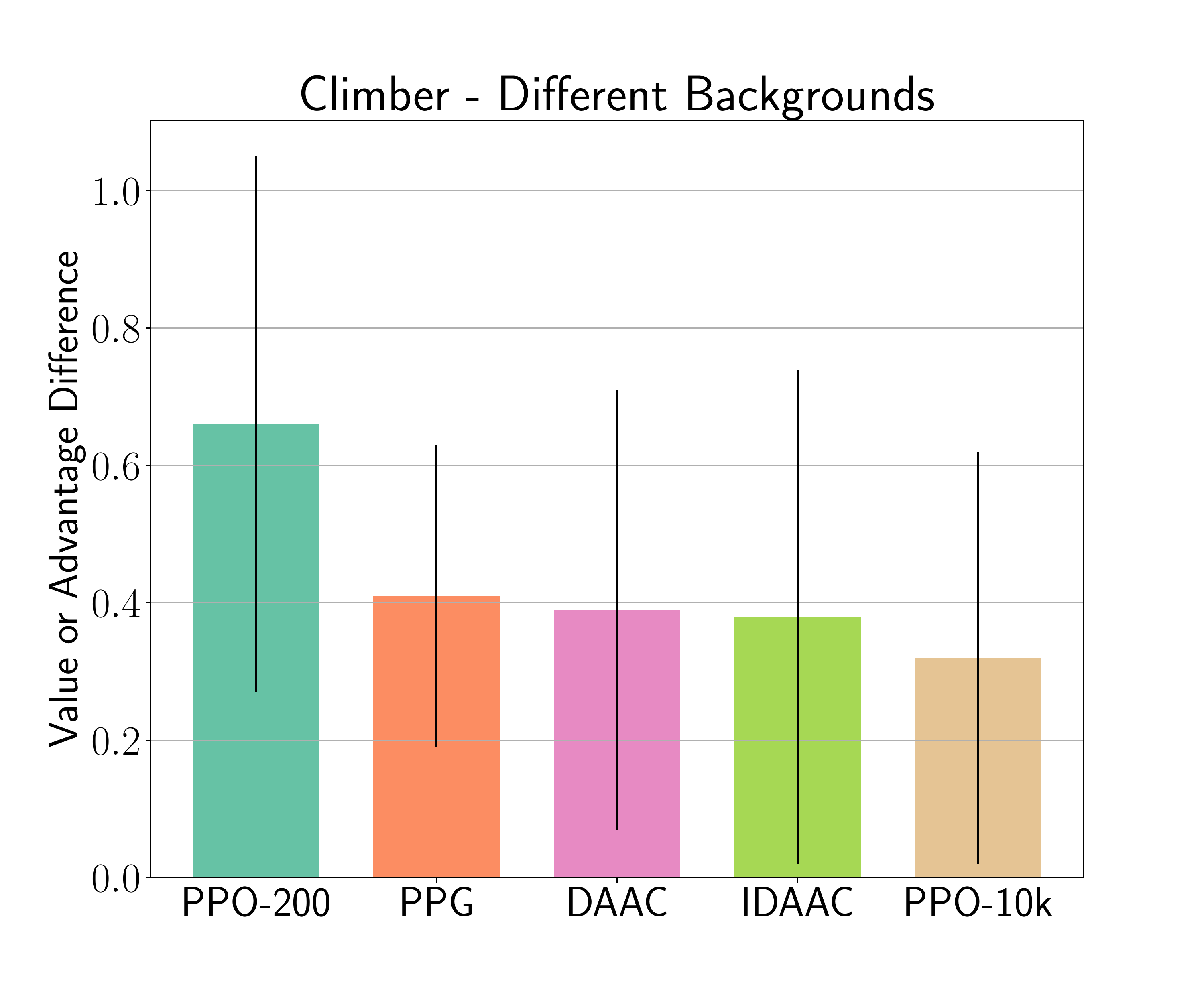}
    \includegraphics[width=0.24\textwidth]{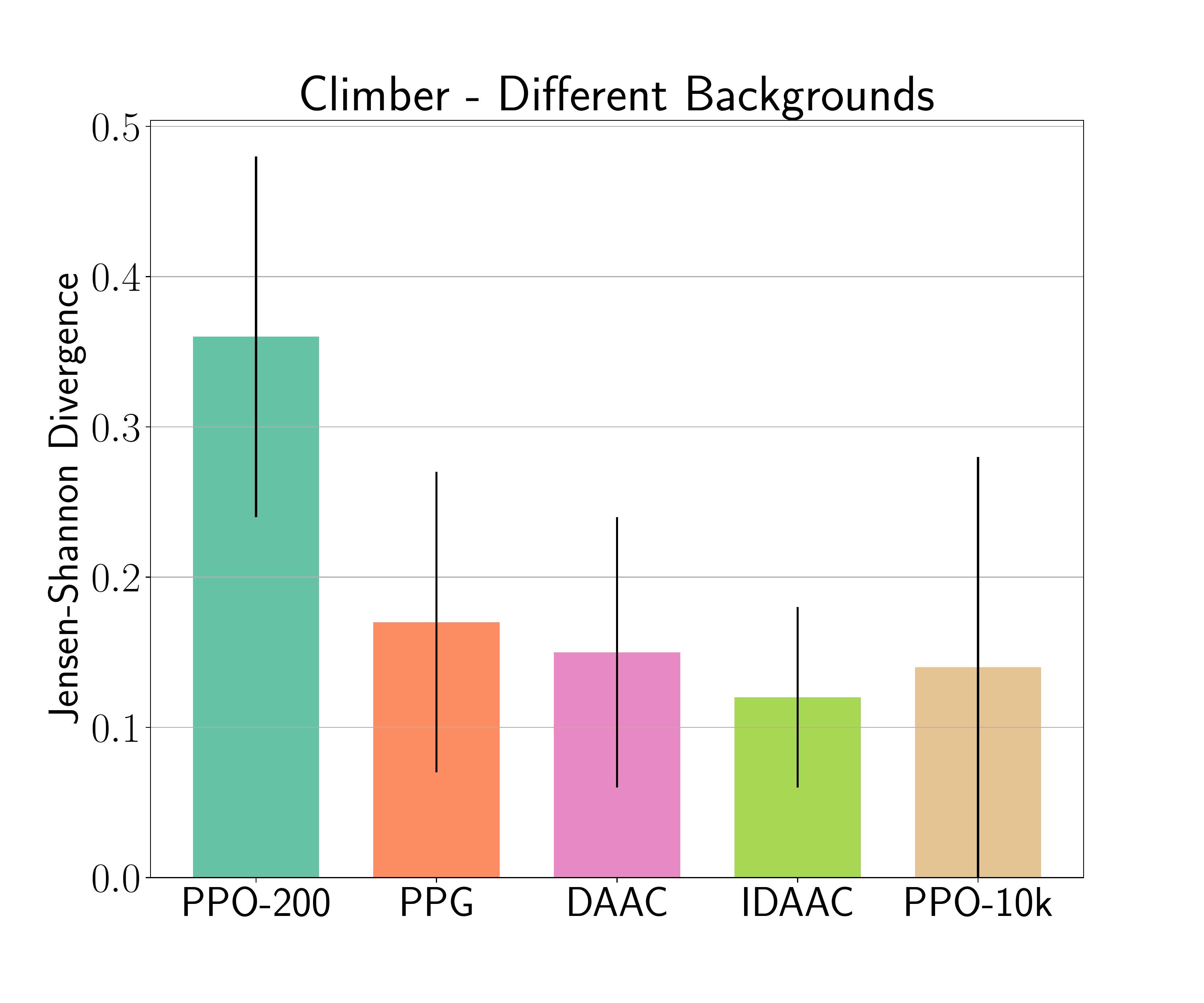}
    \caption{\textbf{Variations in the learned features, policies, and values or advantages when changing the background in Climber.} From left to right we report the L1 and L2-norm for the features, the value or advantage difference, and the Jensen-Shannon Divergence for the policy. We compare \ppo{} trained on 200 and 10k levels with \ppg{}, \gae{}, and \ordergae{}. Our models are more robust to changes in the background (which does not affect the state). The means and standard deviations were computed over 10 different backgrounds.}
    \label{fig:difback_climber}
\end{figure*}

Figures~\ref{fig:difback_ninja},~\ref{fig:difback_jumper}, and~\ref{fig:difback_climber} show the results for Ninja, Jumper, and Climber, respectively, comparing \ordergae{}, \gae{}, \ppg{}, as well as \ppo{} trained on 200 and 10k levels. In particular, the results show that both our methods learn representations which are more robust to changes in the background than PPO and PPG (assuming all methods are trained on the same number of levels \ie{} 200). Overall, the differences due to background changes in the auxiliary outputs of the policy networks for DAAC and IDAAC (\ie{} the predicted advantages) are smaller than those of PPO and PPG (\ie{} the predicted values). These results indicate that DAAC and IDAAC are more robust than PPO and PPG to visual features which are irrelevant for control. 

We do not observe a significant difference across the JSDs of the different methods. However, in the case of Procgen, the JSD isn't a perfect measure of the semantic difference between two policies because some of the actions have the same effect on the environment (\eg{} in Ninja, there are two actions that move the agent to the right) and thus are interchangeable. Two policies could have a large JSD while being semantically similar, thus rendering the policy robustness analysis inconclusive. 

In some cases, PPO-10k exhibits better robustness to different backgrounds than DAAC and IDAAC by exhibiting a lower feature norm and value or advantage difference. However, PPO-10k is a PPO model trained on 10000 levels of a game, while DAAC and IDAAC are trained only on 200 levels. For most Procgen games, training on 10k levels is enough to generalize to the test distribution so PPO-10k is expected to generalize better than methods trained on 200 levels. In this paper, we are interested in generalizing to unseen levels from a small number of training levels, so PPO-10k is used as an upper-bound rather than a baseline since a direct comparison wouldn't be fair. Hence, it is not surprising that some of these robustness metrics are better for PPO-10k than DAAC and IDAAC. 

% \todo{} add more discussion on how these metrics depend on the game. some of them might need fewer levels to solve in which case PPO-10k does better and some of these might be more prone to step overfitting so IDAAC helps more.

\end{document}